\documentclass[10pt]{article} 

\usepackage[accepted]{tmlr}

\usepackage[utf8]{inputenc} 
\usepackage[T1]{fontenc}    
\usepackage{hyperref}       
\usepackage{url}            
\usepackage{booktabs}       
\usepackage{amsfonts}       
\usepackage{nicefrac}       
\usepackage{microtype}      
\usepackage{xcolor}         
\usepackage{graphicx}
\usepackage{amsmath}
\usepackage{subcaption}
\usepackage[ruled,vlined]{algorithm2e}
\usepackage{wrapfig}
\newcommand{\mtx}[1]{\ensuremath{\mathbf{#1}}}
\newcommand{\vtr}[1]{\ensuremath{\mathbf{#1}}}

 \newcommand{\abs}[1]{\left\lvert#1\right\rvert}

\newcommand{\argmin}{\mathop{\mathrm{argmin}}}

\title{Learning Identity-Preserving Transformations on Data Manifolds}

\author{%
  Marissa Connor\thanks{equal contribution.}\;, \; Kion Fallah\footnotemark[1]\;, \;Christopher J. Rozell\\
  School of Electrical and Computer Engineering\\
  Georgia Institute of Technology\\
  Atlanta, GA 30332 \\
  \texttt{(marissa.connor, kion, crozell)@gatech.edu} \\
}

\def\month{03}  
\def\year{2023} 
\def\openreview{\url{https://openreview.net/forum?id=gyhiZYrk5y}} 

\begin{document}

\maketitle

\begin{abstract}
Many machine learning techniques incorporate identity-preserving transformations into their models to generalize their performance to previously unseen data. These transformations are typically selected from a set of functions that are known to maintain the identity of an input when applied (e.g., rotation, translation, flipping, and scaling). However, there are many natural variations that cannot be labeled for supervision or defined through examination of the data. As suggested by the manifold hypothesis, many of these natural variations live on or near a low-dimensional, nonlinear manifold. Several techniques represent manifold variations through a set of learned Lie group operators that define directions of motion on the manifold. However, these approaches are limited because they require transformation labels when training their models and they lack a method for determining which regions of the manifold are appropriate for applying each specific operator. We address these limitations by introducing a learning strategy that does not require transformation labels and developing a method that learns the local regions where each operator is likely to be used while preserving the identity of inputs. Experiments on MNIST and Fashion MNIST highlight our model's ability to learn identity-preserving transformations on multi-class datasets. Additionally, we train on CelebA to showcase our model's ability to learn semantically meaningful transformations on complex datasets in an unsupervised manner.

\end{abstract}

\section{Introduction}

A goal of many machine learning models is to accurately identify objects as they undergo natural transformations -- a task that humans are adept at. According to the manifold hypothesis, natural variations in high-dimensional data lie on or near a low-dimensional, nonlinear manifold~\citep{fefferman2016testing}. Additionally, the manifolds representing different classes are separated by low density regions~\citep{rifai2011manifold}. Natural physical laws govern the possible transformations that objects can undergo and many of the identity-preserving transformations (e.g., changes in viewpoint, color, and lighting) are shared among classes of data. The sharing of transformations between classes enables increased efficiency in defining data variations -- a model can represent a limited set of transformations that can describe a majority of variations in many classes. Several machine learning models incorporate specific identity-preserving transformations that are shared among a large number of classes to generalize the performance of their model to unseen data. These include equivariant models that incorporate transformations like translation and rotation into intermediate network layers ~\citep{cohen2016group, cohen2018spherical} and data augmentation techniques that apply known identity-preserving variations to data while training~\citep{cubuk2019autoaugment,ho2019population,lim2019fast}. However, in many datasets there may be natural transformations shared among classes that are not easily prespecified from intuition, making it critical that we develop a model that can learn both 1) a representation for these transformations without explicit transformation labels and 2) the context of when each transformation is likely to be relevant.

Manifold learning strategies estimate the low-dimensional manifold structure of data and examine how points are related through natural transformations. Most common techniques learn manifold embeddings of data points without a method for generating transformed points on the manifold~\citep{tenenbaum2000global,roweis2000nonlinear,belkin2003laplacian, maaten_visualizing_2008}. Other manifold learning techniques learn to transform points on the manifold either along linear manifold tangent planes~\citep{dollar2007learning,bengio2005non,park2015bayesian,rifai2011contractive,kumar2017semi} or through nonlinear Lie group operators that transverse the manifold~\citep{rao1999learning,miao2007learning,culpepper2009learning,sohl2010unsupervised,cohen2014learning,hauberg2016dreaming,connor2020representing,connor2021variational}. Lie group operators represent infinitesimal transformations which can be applied to data through an exponential mapping to transform points along a manifold, and a manifold can be globally defined by a set of operators that each move in different directions along it~\citep{hoffman1966lie, dodwell1983lie}. A Lie group operator model is well-suited for representing natural data variations because the operators can be learned from the data, applied to data points to transform them beyond their local neighborhoods, and used to estimate geodesic paths. 

While the Lie group operator models have many benefits, previous approaches demonstrate two shortcomings. First, to learn Lie group operators that represent a data manifold, pairs of training points are selected which lie within a neighborhood of one another. The training objective encourages efficient paths between these nearby points, and the choice of training point pairs strongly influences the types of manifold transformations that are learned.  Recent papers incorporating Lie group operators into machine learning models have either used predefined operators that represent known transformations groups (e.g., the 3D rotational group $SO(3)$~\citep{falorsi2019reparameterizing}), required transformation labels for selecting point pairs when training~\citep{connor2020representing}, or randomly selected pairs of points from the same class~\citep{connor2021variational}. To learn an effective model with datasets having no labeled transformation structure, we require a strategy for selecting point pairs that automatically identifies points that are related through the transformations the model aims to learn. Second, existing Lie group operator models have lacked a method for determining which regions of the manifold are appropriate for each operator, meaning existing approaches assume that every operator is equally likely to be used at every point on the manifold. This is a flawed assumption because, while many transformations are shared between classes, there are also data variations that are unique to specific data classes. Additionally, in a dataset with several manifolds (each representing a different class), there is a limited extent to which a transformation can be applied without moving a point onto another manifold. 

The main contributions of this paper are the development of methods to address, in the context of Lie group operator models, the two critical shortcomings of generative manifold models noted above. Specifically, motivated by finding perceptually similar training samples without transformation labels, we first introduce a point pair selection strategy to learn a manifold representation of natural variations shared across multiple data classes without requiring transformations labels. Second, we develop a method that uses a pretrained classifier (measuring identity-preservation of transformed samples) to learn the local regions where each operator is likely to be used while preserving the identity of transformed samples. This approach enables us to analyze the local structure of the data manifold in the context of the learned operators and to describe the invariances of the classifier. We demonstrate the efficacy of these strategies in the context of the Manifold Autoencoder (MAE) model \citep{connor2020representing} to learn semantically meaningful transformations on MNIST~\citep{lecun1998gradient}, Fashion MNIST~\citep{xiao2017online}, and CelebA~\citep{liu2015faceattributes}\footnote{Code available at: \href{https://github.com/Sensory-Information-Processing-Lab/manifold-autoencoder-extended}{\texttt{https://github.com/Sensory-Information-Processing-Lab/manifold-autoencoder-extended}}.}.

\section{Background}

\paragraph{Manifold Learning}
Traditional manifold learning models estimate the low-dimensional structure of high-dimensional data by utilizing the property that local neighborhoods on the manifold are approximately linear. Some techniques represent the manifold through a low-dimensional embedding of the data points~\citep{tenenbaum2000global,roweis2000nonlinear,belkin2003laplacian, maaten_visualizing_2008}. Others learn functions that map high-dimensional data points to linear tangent planes that represent directions of manifold motion in the local neighborhood around the points~\citep{dollar2007learning,bengio2005non,park2015bayesian}. While traditional manifold learning approaches that are applied directly to data points are useful for understanding low-dimensional data structure, in many cases, the input data space is an inefficient representation of the data. For example, data in the pixel space suffers from the curse of dimensionality and cannot be smoothly interpolated while maintaining identity~\citep{bengio2005curse}. Many approaches have used neural networks to learn a low-dimensional latent space in which manifold models can be incorporated. The contractive autoencoder (CAE) estimates manifold tangents by minimizing the Jacobian of the encoder network, encouraging invariance of latent vectors to image space perturbations. Tangent directions can then be estimated through top singular vectors of the encoder Jacobian~\citep{rifai2011contractive,rifai2011higher,rifai2011manifold,kumar2017semi}. Several methods estimate interpolated geodesic paths in the latent space of a trained variational autoencoder (VAE) model~\citep{arvanitidis2017latent,chen2017metrics,shao2018riemannian,arvanitidis2019fast} and show that geodesic paths result in more natural interpolations. Some extend this approach to learn VAEs with priors that are estimated using the Riemannian metrics computed in the latent space~\citep{arvanitidis2021prior,kalatzis2020variational}.

\paragraph{Lie Group Operators}

A Lie group is a group of continuous transformations which also defines a manifold by representing infinitesimal transformations that can be applied to input data~\citep{hoffman1966lie,dodwell1983lie}.  Several methods incorporate Lie groups into neural networks to represent data transformations that are identity-preserving within the model~\citep{cohen2014learning, cohen2018spherical}. A prevalent strategy is to learn a dictionary of Lie group operators that are mapped to a specific group element through the matrix exponential $\mathrm{expm}(\cdot)$ ~\citep{rao1999learning,miao2007learning,culpepper2009learning,sohl2010unsupervised,cohen2014learning,connor2020representing,connor2021variational}. In these models, each operator $\mtx{\Psi}_m$, called a \textit{transport operator}, describes a single direction along the manifold and is parameterized by a single coefficient $c_m$. Given an initial data point $\vtr{z}$, the transport operators define a generative model where transformations can be derived from sampling sparse coefficients $c_m \sim \text{Laplace}\left(0,\zeta\right)$:

\begin{equation}\label{eq:TOgen}
\widehat{\vtr{z}} = \mathrm{expm}\left(\sum_{m=1}^M{\mtx{\Psi}_m c_m}\right)\vtr{z} + n,
\end{equation}
where $n \sim \mathcal{N}(0, \sigma^2_n\mtx{I})$. For analysis on the effect of this sparsity prior on the coefficients, see Appendix~\ref{app:sparsity_analysis}.

The manifold autoencoder (MAE) incorporates the transport operator model into the latent space of an autoencoder to learn a dictionary of operators that represent the global, nonlinear manifold structure in the latent space~\citep{connor2020representing}. This model has been shown to learn operators effectively with transformation supervision, and it will provide the context for demonstrating the effectiveness of the methods developed in this paper. Specifically, we demonstrate a method for learning natural data variations in the MAE latent space using a new strategy for selecting point pairs without transformation labels, as well as a method for learning the regions of the manifold where each operator is likely to be used.

\section{Methods}
The MAE learns a low-dimensional latent representation of the data by defining an encoder function $f : \mathcal{X} \rightarrow \mathcal{Z}$ that maps high-dimensional data points $\vtr{x} \in \mathbb{R}^D$ to low-dimensional latent vectors $\vtr{z} \in \mathbb{R}^d$ and a decoder function $g : \mathcal{Z} \rightarrow \mathcal{X}$ that maps the latent vectors back into the data space~\citep{connor2020representing}. Transport operators $\mtx{\Psi}$ are incorporated into the latent space to learn manifold-based transformations. Before learning the transport operators, the autoencoder is pretrained to extract a latent representation of the data using the traditional autoencoder reconstruction objective.

After pretraining, the autoencoder weights are fixed and the operators are trained with the following objective, which encourages the learning of transport operators that generate efficient paths between the latent vectors $\vtr{z}_0$ and $\vtr{z}_1$ (coinciding with $f(\vtr{x}_0)$ and $f(\vtr{x}_1)$) that are nearby on the manifold:
\begin{equation} \label{eq:objFun}
\mathcal{L}_{\Psi} = \frac{1}{2}\left\|\vtr{z}_1 - \mathrm{expm}\left(\sum_{m=1}^M{\mtx{\Psi}_mc_m} \right) \vtr{z}_0\right\|_2^2 + \frac{\gamma}{2}\sum_m\|\mtx{\Psi}_m\|_F^2 +\zeta\|\vtr{c}\|_1,
\end{equation}
where $\gamma, \zeta > 0$.

Objective~(\ref{eq:objFun}) is minimized via an alternating minimization scheme, as in the linear sparse coding model \cite{olshausen1997sparse}. Specifically, at each training iteration, points pairs $\vtr{x}_0$ and $\vtr{x}_1$ are selected in the input space and encoded into the latent space $\vtr{z}_0$ and $\vtr{z}_1$.
Then, coefficients $\vtr{c}$ are inferred between the encoded latent vectors (by minimizing (\ref{eq:objFun}) with respect to $\vtr{c}$) to estimate the best path between $\vtr{z}_0$ and $\vtr{z}_1$.  After inference, the coefficients are fixed and a gradient step is taken to minimize (\ref{eq:objFun}) with respect to transport operator weights. Once learned, these operators represent different types of motion that traverse the manifold, and they can be combined to generate natural paths on the manifold.

After fitting transport operators to the latent space of a fixed autoencoder network, there is a final fine-tuning training phase that updates the network weights and transport operators simultaneously using a joint objective that combines the autoencoder reconstruction loss with $\mathcal{L}_{\Psi}$. This fine-tuning step has been shown to dramatically improve the quality of learned transformations \cite{connor2020representing} and addresses the potential mismatch between the data manifold and the learned latent structure by adapting the latent structure to fit the transport operators learned between the selected training point pairs. We empirically find that breaking up training into these three phases increases the stability with which the transport operators can be learned. Given the context of this MAE model, in the following sections we describe our contributions.

\subsection{Unsupervised Transformation Learning}\label{sec:pointPair}
There are many possible strategies for selecting training point pairs for a Lie group operator model, and the choice of a strategy can dictate the quality of transformations that are learned. For example, point pairs may be selected as random samples from the dataset or from within the same class. These points, however, are likely to be outside of local manifold neighborhoods and may not provide representations of the natural, perceptually smooth variations in the dataset. Alternatively, point pairs can be selected as those with the lowest Euclidean distance in the pixel or autoencoder latent space. Both of these approaches are limited, however, since nearest neighbors in high dimensions has been noted as ineffective \cite{aggarwal2001} and since unsupervised autoencoder features may not capture semantic transformations of interest through Euclidean distance. In some applications, there may be additional information about the dataset that can aid in point pair selection, such as rotation angle, semantic label, or time index in a temporal sequence. However, for most complex datasets, information about the transformations of interest is not available. This necessitates a strategy for selecting training point pairs which can learn natural transformations in complex datasets without requiring additional transformation supervision.

We generalize transport operator training to incorporate an unsupervised learning strategy that can be applied to a wide array of datasets. In particular, we select point pairs using a perceptual loss metric~\citep{johnson2016perceptual}. The perceptual loss measures the distance between feature representations of input images that are extracted from a classifier pretrained on a large dataset, like ImageNet. It has been shown that features in the penultimate layer of such pretrained classifiers serve as an effective surrogate for visual similarity~\citep{johnson2016perceptual,yosinski2015understanding}. Given an initial point $\mathbf{x}_0$, we define point pairs in a feature embedding $r(\mathbf{x})$ as:
\begin{equation}
    \label{eq:nn}
    \vtr{x}_1 =  \argmin_{\vtr{x} \in \mathcal{D}} \Vert r(\vtr{x}_0) - r(\vtr{x}) \Vert_2^2,
\end{equation}
where, in practice, we randomly select $\mathbf{x}_1$ from the $N$ nearest points to $\mathbf{x}_0$ in the classifier feature space.

With the proposed approach, we are capable of training transport operators without labels to select point pairs. Furthermore, we find that the perceptual loss results in transport operators that correspond to smooth, semantically meaningful transformations. We compare the effect of training with the proposed approach versus other point pair strategies in Appendix~\ref{app:pp_supervision}.

\subsection{Learning Local Transport Operator Statistics to Encourage Identity Preservation}\label{sec:coeffEnc}%

A trained transport operator model provides a dictionary of operators that can be applied globally across the entire data space. This model has flexibility to define local manifold characteristics through the coefficients $\vtr{c}$ which specify the combination of operators to apply to a given point. The standard prior on the transport operator coefficients is a factorial Laplace distribution parameterized by a single scale parameter $\zeta$, which encourages transformations to be defined by a sparse set of operators:

\begin{equation}
    p_\zeta(\vtr{c})=\prod_{m=1}^M \frac{1}{2 \zeta}\exp\left(- \frac{ \abs{c_m}}{\zeta}\right). \label{eq:priorGTfact}
\end{equation}

By setting a fixed prior on coefficients across the entire manifold, there is an implicit assumption that the manifold structure is the same over the whole dataset and every operator is equally likely to be applied at every point. However, this is a flawed assumption because not all transformations in one class of data are present in all other classes and even within classes there may be regions with different manifold structure. Not capturing the local statistics of transport operator usage  could result in transformed points that depart from the data manifold and change their identity.

To address this limitation, we introduce a coefficient encoder network which maps latent vectors to scale parameters that specify the Laplace distribution on the coefficients at those points in the latent space. The goal of this network is to estimate local coefficient distributions that maximize the identity-preservation of points that are transformed with operators characterized by randomly sampled coefficients:

\begin{equation}
    \label{eq:gen_aug}
    \widehat{\vtr{z}} = T_{\Psi}(\vtr{c})\vtr{z} + \epsilon, \quad \epsilon \sim \mathcal{N}(0, \sigma_{\epsilon}^2 I),
\end{equation}
{where $T_\Psi(\vtr{c}) = \mathrm{expm}\left(\sum_{m=1}^M{\Psi_m c_m}\right) \in \mathbb{R}^{d \times d}$,  is the transformation matrix corresponding to the transport operators (parameterized by coefficients $\vtr{c} \in \mathbb{R}^M$).

Given labeled observations $(\vtr{z}, y)$ and a pretrained classifier $r(\cdot)$, we aim to learn a network $q_\phi (\vtr{c} \mid \vtr{z})$ that outputs distribution parameters from which $\vtr{c}$ can be sampled and applied to produce augmented samples $\widehat{\vtr{z}} \in\mathbb{R}^d$ without changing the classifier output $r(\widehat{\vtr{z}}) = y$. }

To maximize the likelihood of obtaining the labeled classifier output for augmented samples, we adapt the concept of consistency regularization from semi-supervised learning. Consistency regularization is applied when training a classifier in a semi-supservised setting in order to ensure that known identity-preserving augmentations cause only small changes in the classifier output~\citep{bachman2014learning}. In our context, we use a pretrained classifier and a dictionary of transport operators and we want to find a distribution on the coefficients at individual input points that results in consistent classification outputs when $T_\Psi(\vtr{c})$ is applied to inputs. The specific objective can be chosen from a variety of loss functions that encourage similarity in classifier probability outputs. We specifically minimize the KL-divegence between the label $y$ and the label of the transformed output $r(\widehat{\vtr{z}})$. 

Unfortunately, the consistency regularization objective can be trivially minimized by setting $\vtr{c} = 0$, resulting in an identity transformation and the same classifier output. Therefore, we want to encourage the largest coefficient values possible while maintaining the identity of our initial data point. This motivates the addition of a KL-divergence regularizer that encourages the distribution $q_\phi(\vtr{c} | \vtr{z})$ to be similar to a specified Laplace prior with a fixed scale $\zeta$ like in (\ref{eq:priorGTfact}). Our final objective for training the coefficient encoder network is:

\begin{equation}\label{eq:cEncObj}
E = D_{KL}(y \Vert r(\widehat{\vtr{z}})) + D_{KL}\left( q_\phi(\mathbf{c} \mid \mathbf{z}) \Vert p_\zeta(\mathbf{c})\right).
\end{equation}

 A more detailed derivation of this training objective can be found in Appendix~\ref{app:coeffEnc}.
 
 The addition of this coefficient encoder is a principled way to build identity-preservation directly into our model and to identify local manifold characteristics throughout the latent space. Since this model is trained using outputs from a pretrained classifier, the resulting encoded coefficient scale weights can be informative about classifier invariances. Specifically, we can quantify which operators are associated with large encoded coefficient scale weights in different parts of the latent space, indicating that the classifier is invariant to those transformations. Additionally, we can identify points or regions of space that have small encoded coefficient scale weights, indicating they are near class boundaries and can undergo only small transformations without changing identity.

\subsection{Decreasing Computational Complexity of Coefficient Inference}\label{sec:coeffInfer}

 To compute the gradient for the matrix exponential during training and inference, previous works have used a linear approximation \citep{rao1999learning}, learned the operators in a factored form \citep{sohl2010unsupervised}, or used analytic gradients that are not favorable for parallel computations \citep{culpepper2009learning, connor2020representing, connor2021variational}. The main computational bottleneck for learning transport operators is the coefficient inference step, where objective~(\ref{eq:objFun}) is minimized with respect to $\vtr{c}$. The $\ell_1$ norm term is non-smooth, leading to slower convergence when using subgradients found via automatic differentiation \citep{boyd_convex_2004}.  
 
 In this work, we employ a forward-backward splitting scheme \citep{beck_fast_2009} that applies automatic differentiation for the matrix exponential term, with numerical approximations that are more favorable for parallel hardware \cite{math7121174}, and proximal gradients for the $\ell_1$ norm \citep{10.1561/2400000003}. We show in Appendix~\ref{app:inference_details} (with further details on the coefficient inference algorithm) that this leads to significant speedup during training, and allows for scaling of transport operators to complex datasets that were not possible before.

\section{Analysis}

For our experiments, we examine the ability of our proposed approaches to enable the MAE model to learn natural transformations from complex datasets where  the underlying identity-preserving transformations are not easily identifiable. We also highlight the benefits of incorporating the coefficient encoder network that captures the transport operator local usage statistics by encoding coefficient scale values that best maintain the identity of latent vectors. 

We work with three datasets: MNIST~\citep{lecun1998gradient}, Fashion MNIST~\citep{xiao2017online}, and CelebA~\citep{liu2015faceattributes}. We select MNIST and Fashion MNIST because they contain several classes that share natural transformations but they do not have transformation labels. We select CelebA to highlight our ability learn natural transformations in a larger, more complex dataset. As a classic dataset used in papers that aim to disentangle dataset features \citep{higgins2017beta, chen2016infogan, hu2018disentangling, lin2019exploring}, CelebA contains semantically meaningful natural transformations that may be amenable to qualitative labeling. 

In all experiments, we followed the general training procedure put forth previously~\citep{connor2020representing} by separating the network training into three phases: the autoencoder training phase, the transport operator training phase, and the fine-tuning phase. We select training point pairs that are nearest neighbors in the feature space of the final, pre-logit layer of a ResNet-18 \citep{he2015deep} classifier pretrained on ImageNet~\citep{russakovsky2015imagenet}. After completely training the MAE, we fix the autoencoder network weights and transport operator weights and train the coefficient encoder network with the objective derived in Section~\ref{sec:coeffEnc}. Additional details on the datasets, network architectures, and training procedure are available in the Appendix.

We compare against the contractive autoencoder (CAE) \citep{rifai2011contractive} and $\beta$-VAE \citep{higgins2017beta}, two other autoencoder based methods that incorporate data structure into the latent space. The CAE represents another technique that learns a manifold representation in a neural network latent space. In their model, the manifold is represented by estimated manifold tangent planes at latent point locations. The $\beta$-VAE learns to disentangle factors of variation along latent dimensions through an increase in the weighted penalty on the KL-divergence term in the VAE objective. We choose these methods as they provide approaches to learn natural dataset variations in the latent space without transformation labels.

\subsection{Learning Natural Data Variations}\label{sec:celeba_exp}

\begin{figure}[t]

\centering
\begin{subfigure}[b]{0.8\textwidth}
  \centering
	{\includegraphics[width=0.98\textwidth]{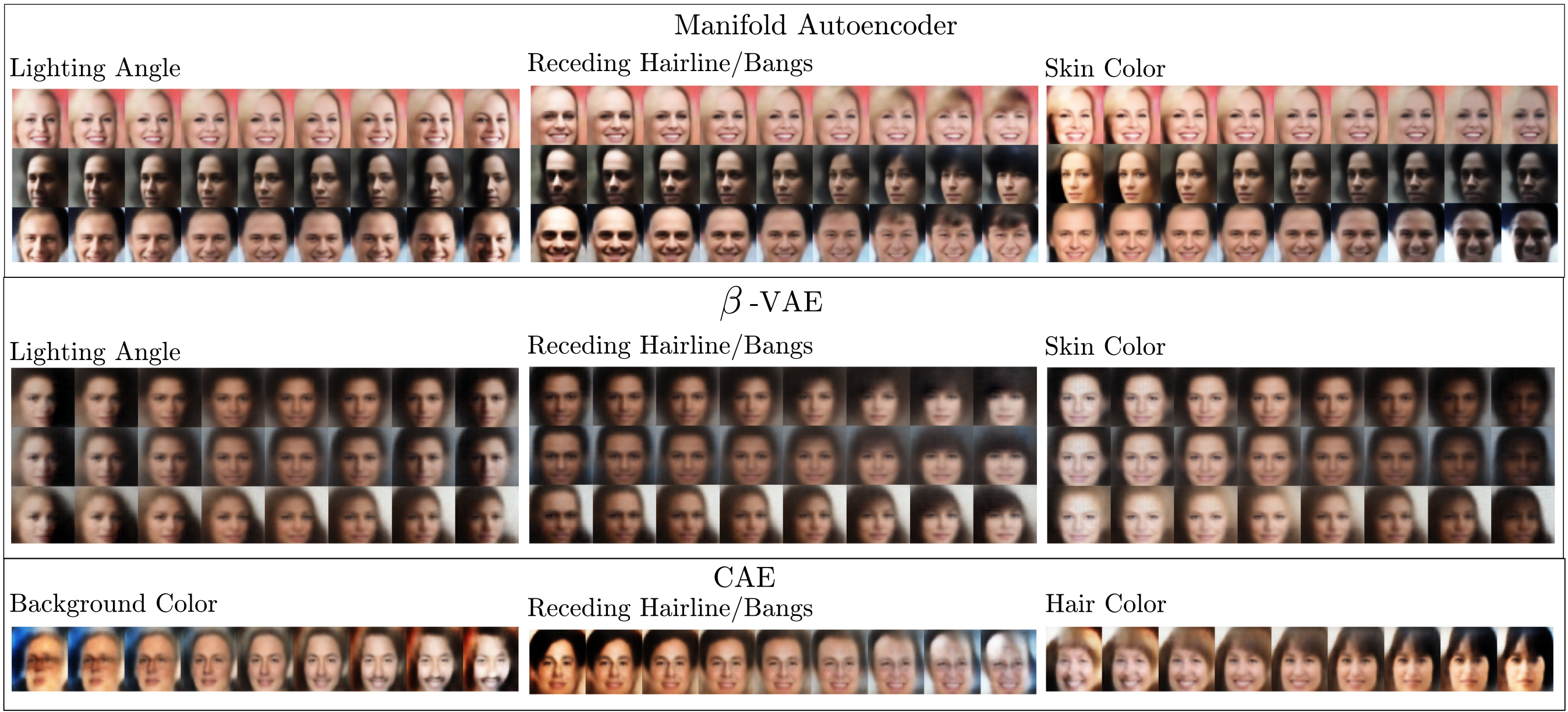}}
  \caption{}
	\label{subfig:celebA_qualTrans}
\end{subfigure}
\begin{subfigure}[b]{0.8\textwidth}
  \centering
	{\includegraphics[width=0.98\textwidth]{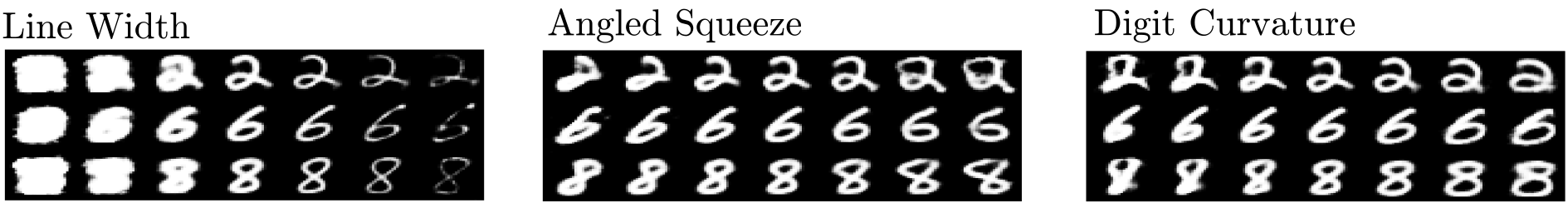}}%
  \caption{}
	\label{subfig:mnist_qualTrans}
\end{subfigure}
\begin{subfigure}[b]{0.8\textwidth}
  \centering
	{\includegraphics[width=0.98\textwidth]{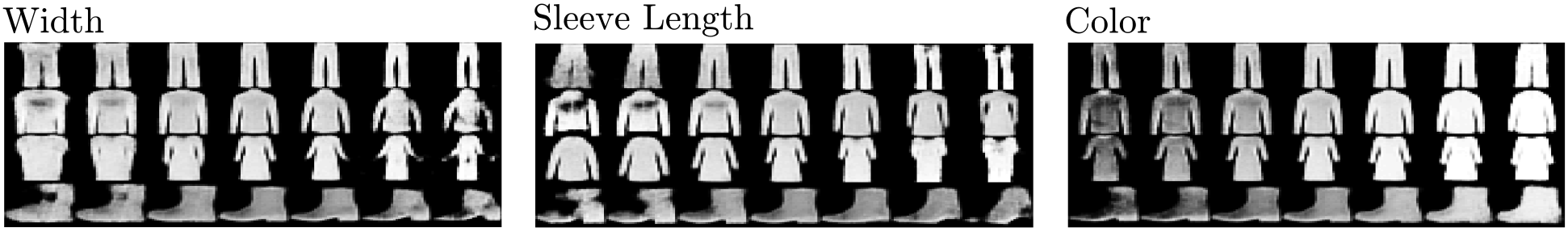}}%
  \caption{}
	\label{subfig:fmnist_qualTrans}
\end{subfigure}

  \caption{\label{fig:qualTransform} Paths generated by applying a subset of learned transport operators on three datasets. In each figure, images in the middle column of the image block are the reconstructed inputs and images to the right and left are images decoded from transformed latent vectors in positive and negative directions, respectively. (a) Comparing the transformations generated by three learned transport operators to transformations generated by $\beta$-VAE and CAE. The transport operators learn semantically meaningful transformations similar to the disentangled $\beta$-VAE representation, while maintaining a higher resolution in image outputs.  (b-c) Transport operators learned using the perceptual point pair selection strategy generate natural transformations on both MNIST and Fashion MNIST.}
	
\end{figure}

First we show how well our proposed methods enable the MAE model to learn natural data variations in the dataset when selecting point pairs using the perceptual point pair selection strategy and we highlight the usefulness of a nonlinear manifold model for generating latent space paths. Fig.~\ref{fig:qualTransform} shows the paths generated by transport operators trained in this model. The image in the middle column in each block of images is the reconstructed version of the input image $\mathbf{x}_0$. The images to the left and right of the middle show the reconstructed images generated by an individual learned operator applied to the encoded latent vector $\mathbf{z}_0$: $\mathbf{z}_c = \mathrm{expm}(\mtx{\Psi}_m c)\mathbf{z}_0$, $c = -N_c,...,N_c$. We see an individual operator generates a similar transformation across multiple inputs, and in many cases, the transformations induced by the transport operators are semantically meaningful. We also show that the perceptual point pair selection strategy is effective over a range of datasets.

\begin{figure}[t]

\centering
\begin{subfigure}[b]{0.49\textwidth}
  \centering
	{\includegraphics[width=0.98\textwidth]{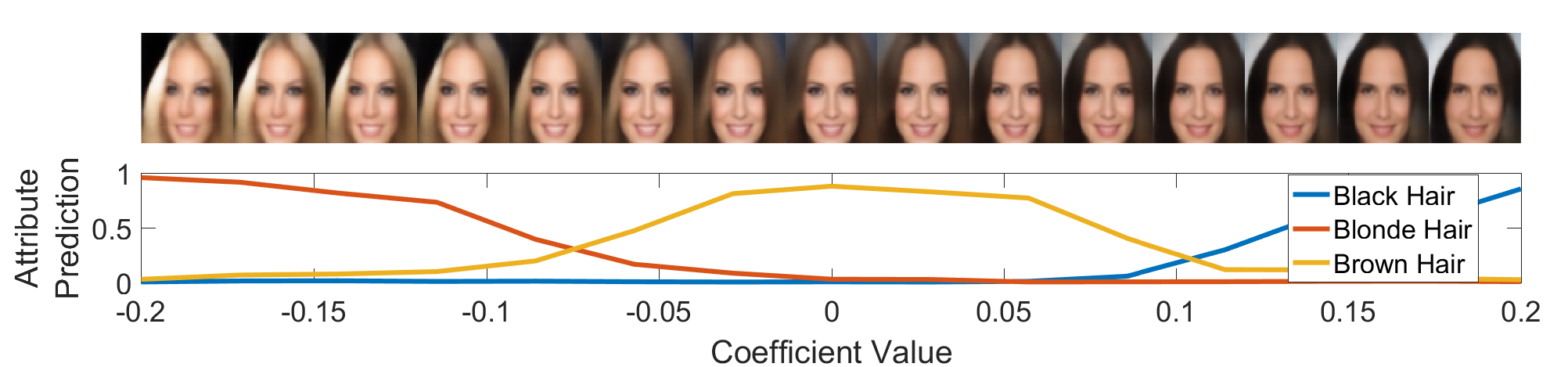}}
  \caption{}
	\label{subfig:attHair}
\end{subfigure}
\begin{subfigure}[b]{0.49\textwidth}
  \centering
	{\includegraphics[width=0.99\textwidth]{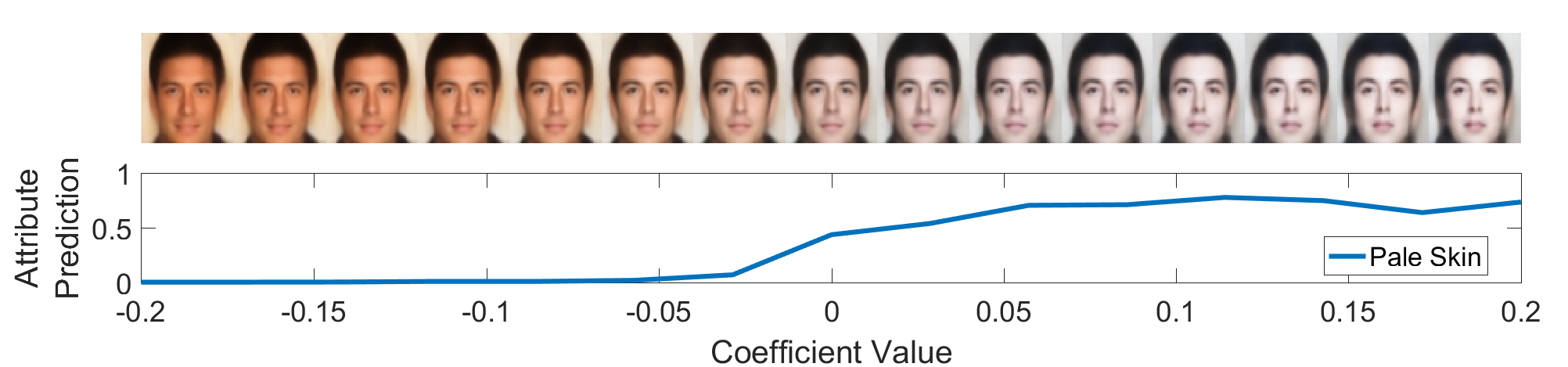}}%
  \caption{}
	\label{subfig:attPale}
\end{subfigure}
\begin{subfigure}[b]{0.49\textwidth}
  \centering
	{\includegraphics[width=0.99\textwidth]{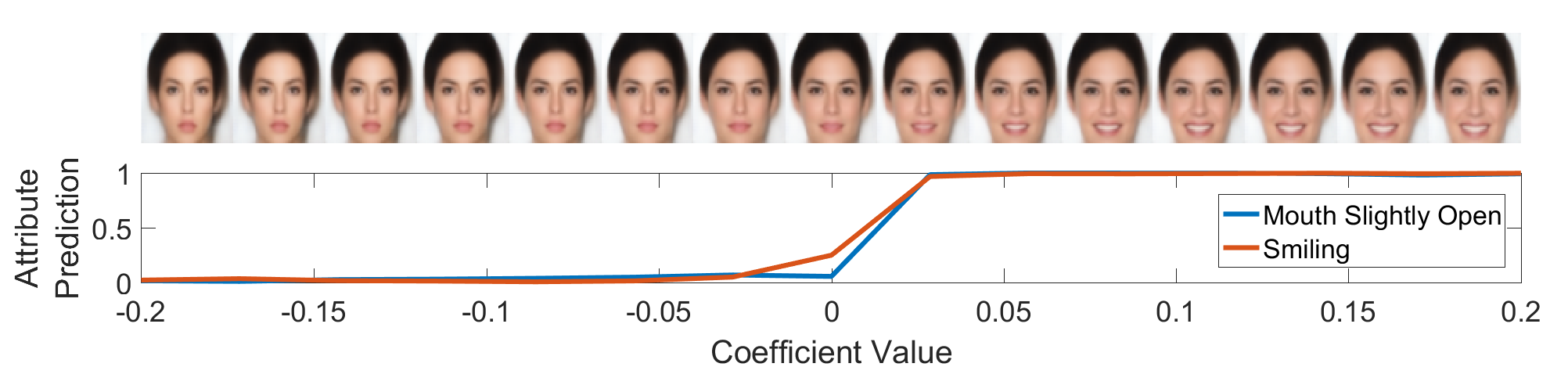}}
  \caption{}
	\label{subfig:attSmile}
\end{subfigure}
\begin{subfigure}[b]{0.49\textwidth}
  \centering
	{\includegraphics[width=0.99\textwidth]{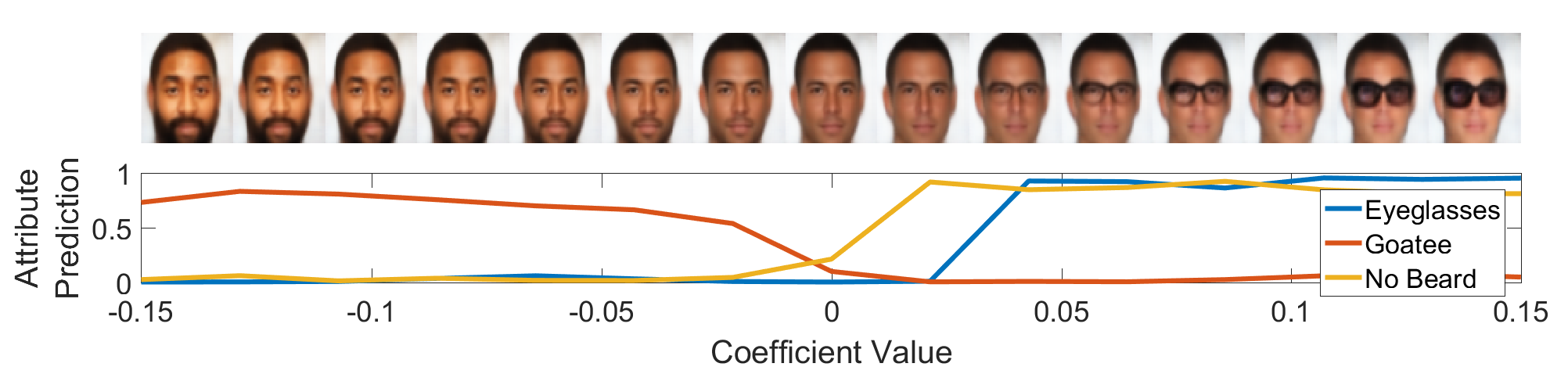}}%
  \caption{}
	\label{subfig:attBeard}
\end{subfigure}

  \caption{\label{fig:attributes} Paths generated by the application of two learned transport operators with the associated attribute classifier probability outputs for the transformed images. Our model learns operators that correspond to meaningful CelebA dataset attributes like (a) hair color (b) pale skin (c) smiling (d) sunglasses/beard}
	
\end{figure}

In Fig.~\ref{fig:qualTransform}, we compare the transport operator paths to paths generated by the CAE and the $\beta$-VAE.  { We compare with the CAE as it also incorporates manifold structure into the latent space, although it uses a tangent approximation that limits the size of reasonable transformations. On the other hand, we choose the $\beta$-VAE as a ubiquitous technique for disentangling factors of variation across Euclidean coordinates. Similarly, the MAE learns disentangled factors of variation via the non-linear paths spanned by the transport operators.} The CAE-generated paths represent the directions of motion on the tangent planes estimated at individual points. The $\beta$-VAE paths are generated by varying the value of one latent dimension, while the others remain fixed. While our method and the $\beta$-VAE learn several qualitatively similar transformations, our method is capable of doing so without significantly sacrificing reconstruction performance. In the Appendix, we include more examples of the transformations generated by each learned transport operator across each dataset.

While many of the operators can be assigned a semantic label through qualitative visual inspection, the CelebA dataset has attribute labels for the images which enable a quantitative analysis of the connection between learned transport operators and dataset attributes. To classify attributes, we fine-tune a ResNet-18 pretrained on ImageNet with 16 classification heads for attributes including smile, beard, hair color, and pale skin \citep{mao2020deep}. With this classifier, we are able to quantify the degree with which specific transport operators correspond to dataset attributes. Fig.~\ref{fig:attributes} shows the classification outputs of the attribute classifier for specific transport operators. Our model learns operators that vary hair color, skin paleness, and several others attributes, as shown in Appendix~\ref{app:celeba_extra}.

\begin{figure}[t]

\centering
\begin{subfigure}[b]{0.98\textwidth}
  \centering
	{\includegraphics[width=0.98\textwidth]{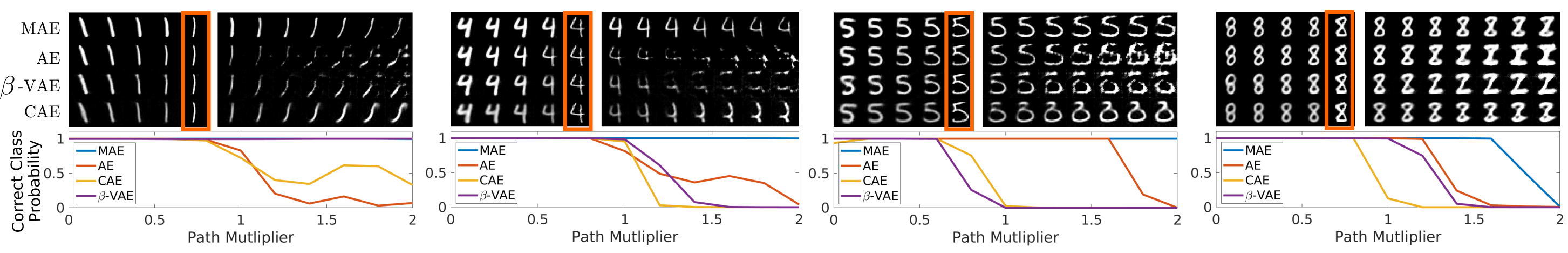}}
  \caption{}
	\label{subfig:interpMNIST}
\end{subfigure}
\begin{subfigure}[b]{0.98\textwidth}
  \centering
	{\includegraphics[width=0.98\textwidth]{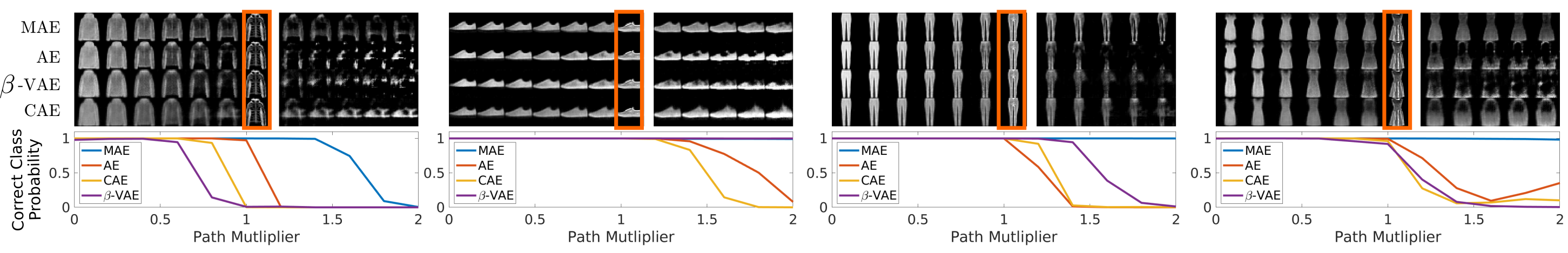}}%
  \caption{}
	\label{subfig:interpFMNIST}
\end{subfigure}

  \caption{\label{fig:interpExtrap} Identity preservation of transformed paths as quantified by a pretrained classifier output. In the figures on the top, the first block of images corresponds to the interpolated path with selected final point $x_1$ surrounded in an orange box. The second block of images corresponds to the extrapolated paths. Below the images are plots of the probability of the class label associated with the inputs $\vtr{z}_0$ and $\vtr{z}_1$. A path multiplier between 0 and 1 indicates interpolation and path multipliers beyond 1 indicate extrapolation. (a) MNIST (b) Fashion MNIST}
\vspace{-0.3in}
\end{figure}

One benefit of Lie group operator models over other manifold-based models is that the learned transport operators provide a natural way for estimating nonlinear manifold paths between points in the latent space. To highlight this benefit, in Fig.~\ref{fig:interpExtrap} we compare the MAE-based interpolated and extrapolated paths to those estimated using CAE and $\beta$-VAE, as well as linear paths in our original autoencoder before fine-tuning it to fit the manifold structure defined by the learned transport operators. With each method, we estimate a path between the latent vectors associated with two points $\vtr{z}_0$ and $\vtr{z}_1$, interpolate between the two points, and extrapolate beyond $\vtr{z}_1$ to extend the path. For MAE, this path is estimated by inferring the set of transport operator coefficients $\vtr{c}^*$ between the latent points and then generating the path: $\vtr{z}_t = \mathrm{expm}\left(\sum_{m=1}^M{\Psi_m c^*_m t}\right) \vtr{z}_0$. When the path multiplier $t$ is between 0 and 1 that indicates interpolation and path multipliers beyond 1 indicate extrapolation. To estimate paths in AE, CAE and $\beta$-VAE, we compute the vector difference between $\vtr{z}_0$ and $\vtr{z}_1$ and interpolate and extrapolated on that vector direction. In these figures, the first block of images corresponds to the interpolated path with the selected final point $\vtr{x}_1$ surrounded in an orange box. The second block of images corresponds to the extrapolated paths.

\begin{figure}[t]

\centering
\begin{subfigure}[b]{0.3\textwidth}
  \centering
	{\includegraphics[width=0.98\textwidth]{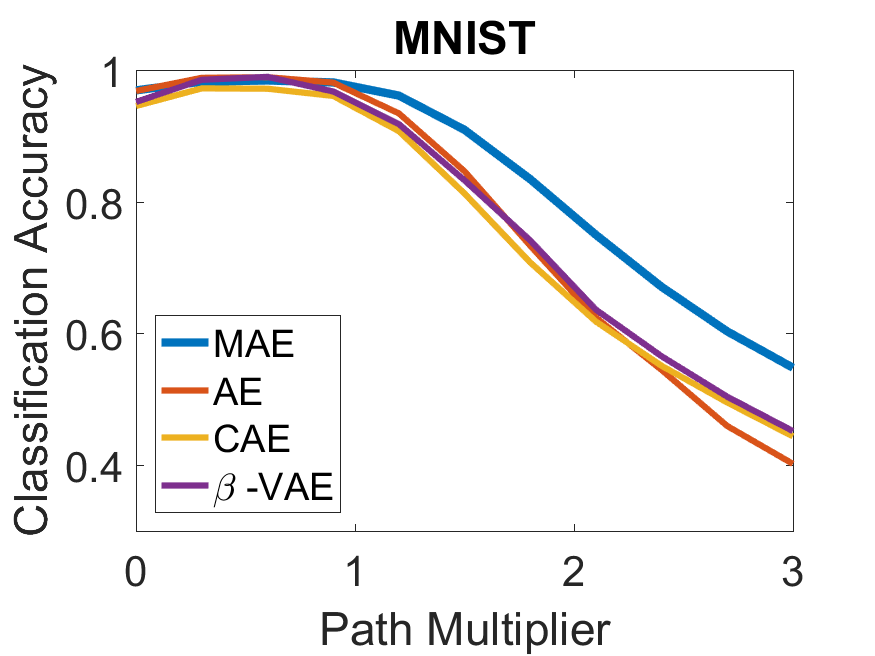}}
  \caption{}
	\label{subfig:extrapClass_mnist}
\end{subfigure}
\begin{subfigure}[b]{0.3\textwidth}
  \centering
	{\includegraphics[width=0.98\textwidth]{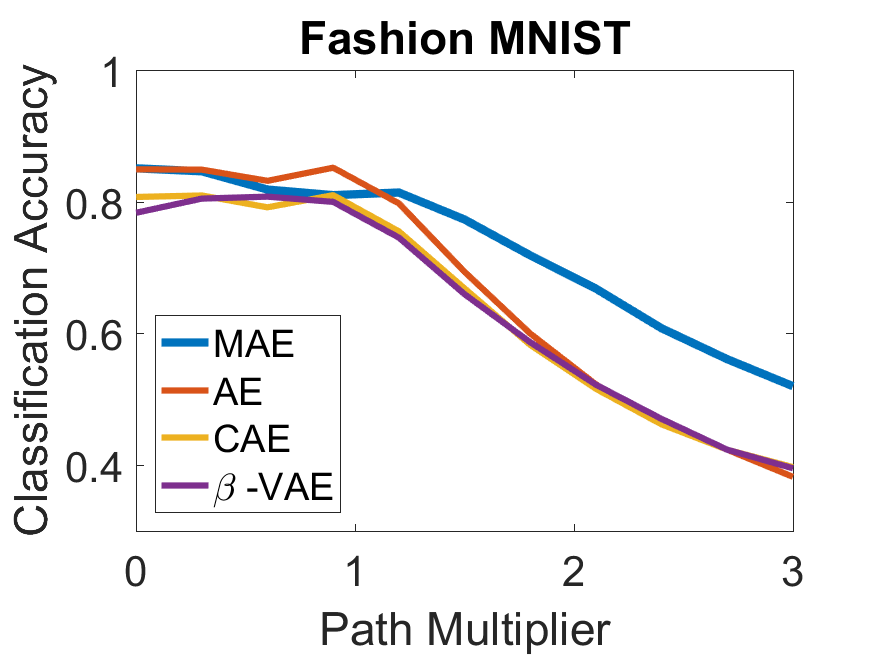}}%
  \caption{}
	\label{subfig:extrapClass_fmnist}
\end{subfigure}

  \caption{\label{fig:extrapClass} The average accuracy of the classifier output for images at each point along the interpolation/extrapolation sequence. When the path multiplier is between 0 and 1 that indicates interpolation and path multipliers beyond 1 indicate extrapolation. (a) MNIST (b) Fashion MNIST}
\vspace{-0.3in}
\end{figure}

For quantifying the identity preservation of each transformation, we input each generated image into a pretrained classifier and plot the probability of the class label associated with the inputs. All four methods perform interpolation effectively but our trained model estimates the extrapolated paths more accurately. Fig.~\ref{fig:extrapClass} shows how that accuracy varies during extrapolation sequences for 4000 samples. The MAE is better at generating extrapolated outputs that maintain class identity. The lower classification accuracy in fashion MNIST is due to both a more challenging dataset and the lower resolution of autoencoder image outputs when compared to input images.

\subsection{Learning Local Manifold Structure}

\begin{figure}[t]

\centering
\begin{subfigure}[b]{0.48\textwidth}
  \centering
	{\includegraphics[width=0.98\textwidth]{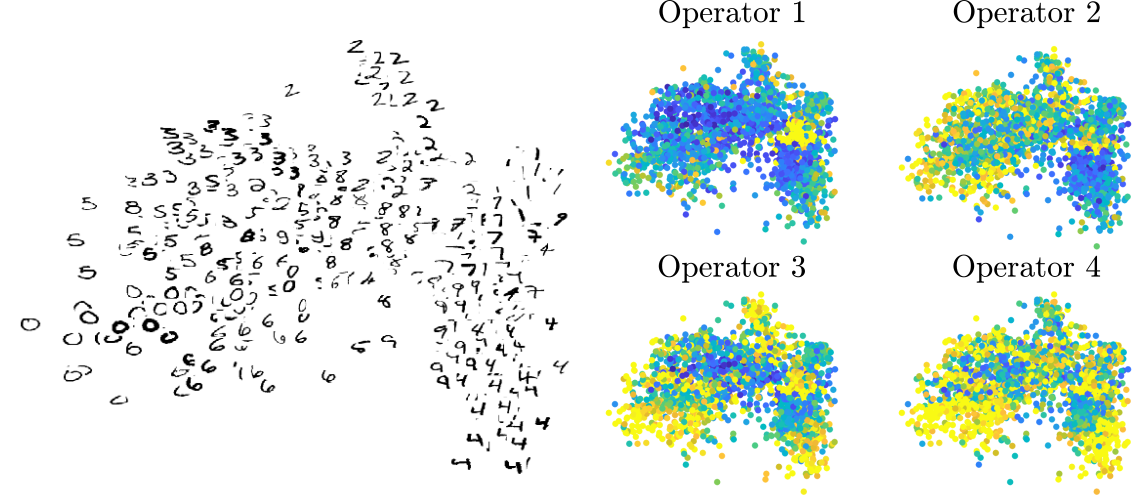}}
  \caption{}
	\label{subfig:embed_mnist}
\end{subfigure}
\begin{subfigure}[b]{0.48\textwidth}
  \centering
	{\includegraphics[width=0.98\textwidth]{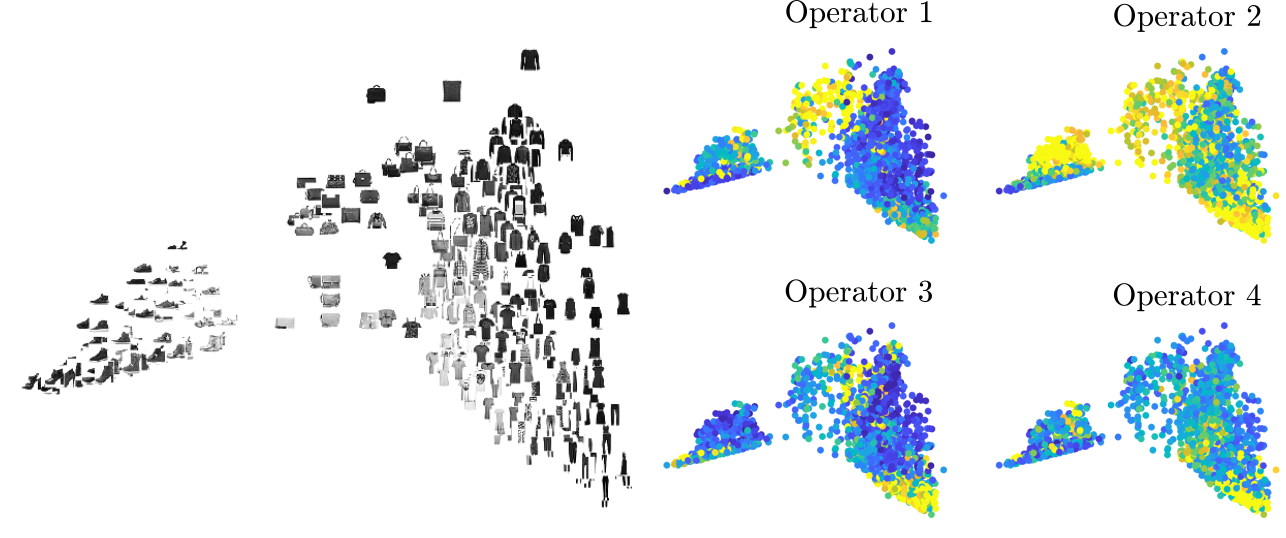}}
  \caption{}
	\label{subfig:embed_fmnist}
\end{subfigure}

\begin{subfigure}[b]{0.48\textwidth}
 \centering
	{\includegraphics[width=0.98\textwidth]{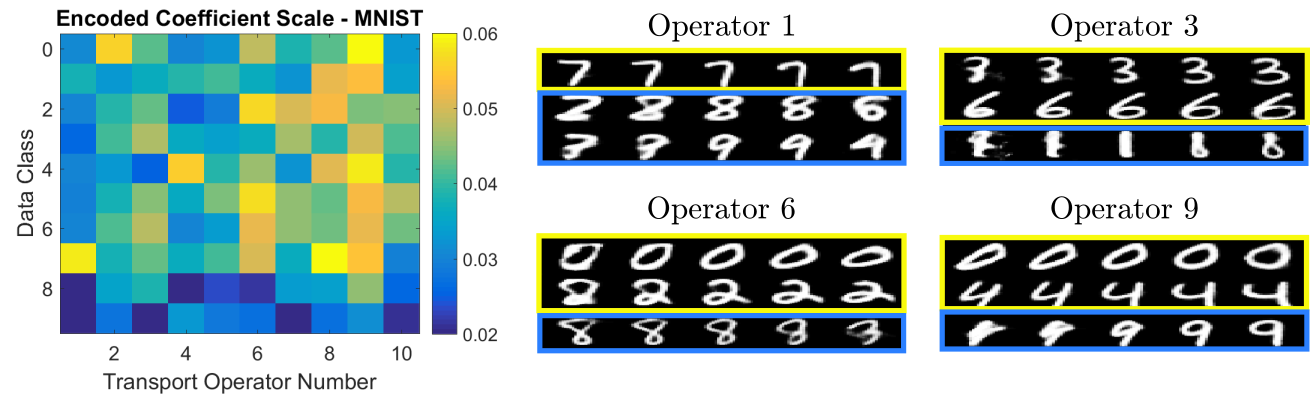}}
 \caption{}
	\label{subfig:spread_mnist}
\end{subfigure}
\begin{subfigure}[b]{0.48\textwidth}
 \centering
	{\includegraphics[width=0.98\textwidth]{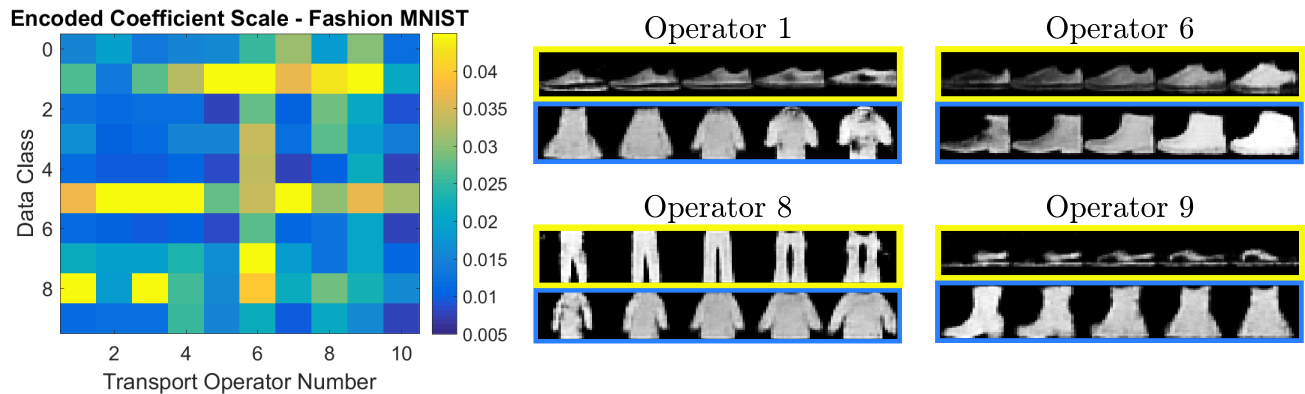}}
 \caption{}
	\label{subfig:spread_fmnist}
\end{subfigure}

  \caption{\label{fig:embedCombo} Visualizations of the encoded coefficient scale weights for (a, c) MNIST and (b, d) Fashion MNIST. Figure best viewed zoomed in. (a, b): Isomap embedding of the latent vectors with input images overlaid. The scatter plots on the right show the same Isomap embedding colored by the encoded coefficient scale weights for several operators (yellow indicates large scale weights and blue small scale weights). We see operators whose use is localized in regions of the manifold space. (c, d): The average coefficient scale weights for each class on each transport operator. High scale weights for a given operator (yellow) indicate it can be applied to a given class without easily changing identity. The images on the right show examples of the operators applied to classes with high encoded scale weights (in the top yellow boxes) and classes with low encoded scale weights (in the bottom blue boxes). The examples with low coefficient scale weights change classes more easily than other examples.}
	
\end{figure}

After the MAE is trained, we have a dictionary of transport operators that describe manifold transformations and a network with a latent space that is adapted to the manifold. We then train the coefficient encoder from Section~\ref{sec:coeffEnc} to estimate the coefficient scale weights as a function of points in the latent space. To visualize how the use of the transport operators varies over the latent space, we generate an Isomap embedding~\citep{tenenbaum2000global} of latent vectors and color each point by the encoded scale parameter for coefficients associated with each of the transport operators. Fig.~\ref{subfig:embed_mnist} and~\ref{subfig:embed_fmnist}  shows these embeddings for MNIST and Fashion MNIST, respectively. Each operator has regions of the latent space where their use is concentrated.

\begin{figure}[t]

	{\includegraphics[width=0.98\textwidth]{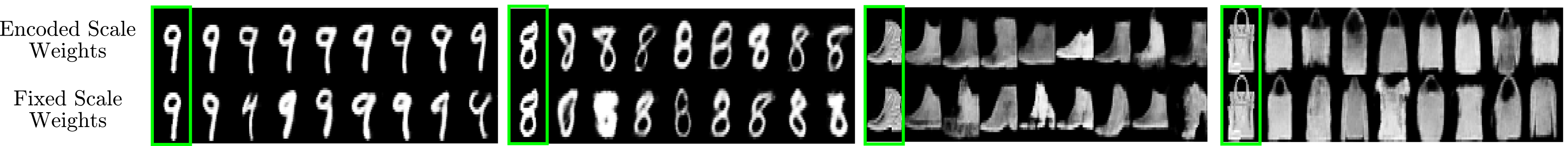}}

  \caption{\label{fig:samp} Examples of samples generated by transport operators using coefficients sampled with encoded scale weights (top row) and with a fixed scale weight (bottom row). Images in the green box are the input images and the remaining images in each row are transformed outputs.  }
	
\end{figure}

By training the coefficient encoder to maximize the similarity between the classification outputs for an input sample and for a transformed version of that sample, the network aids in identifying which transport operators can be applied to inputs in regions of the data space without changing the identity of the input. This helps significantly with data augmentation where the goal is to create new samples with in-class variations. To highlight this benefit, in Fig.~\ref{fig:samp} we show samples augmented by applying transport operators with randomly sampled coefficients to an input latent vector. In each block of images, the leftmost image (in a green box) is the input image and the images to the right are decoded augmentations. The top row shows samples augmented with transport operators controlled by coefficients sampled from Laplace distributions with encoded coefficient scale weights. The bottom row  shows samples augmented with transport operators controlled by coefficients sampled from Laplace distributions with a fixed scale parameter. {The augmentations in these figures differ from previous ones where augmentations were generated using coefficients interpolated within a fixed range, rather than sampled.} While both strategies generate some realistic variations of the data, using the encoded scale weights improves identity-preservation of the transformed output. The augmentations with the encoded scale weights are better at maintaining the identity of the sampled points.

Because the coefficient encoder uses a pretrained classifier to determine the coefficient distributions in the latent space, the resulting encoded scale weights can inform us about the types of manifold transformations the classifier is invariant to. Fig.~\ref{subfig:spread_mnist} and~\ref{subfig:spread_fmnist} shows the average coefficient scale weights for each class (rows) and each transport operator (columns) for MNIST and Fashion MNIST, respectively. From this, one can identify classes for which the classifier is both sensitive and robust to transformations, represented by small and large scale weights respectively. We can also examine which classes share the use of the same transformations. The images to the right of in Fig.~\ref{subfig:spread_mnist} show transport operators being applied to samples with high encoded scale weights (in a yellow box) and samples with low encoded scale weights (in a blue box). It is interesting to examine the characteristics of transformations that are better suited to some classes than others. For instance, operator 3 in Fig.~\ref{subfig:spread_mnist} increases the curve at the bottom of digits. This is a natural transformation for classes 3, 5, and 6 which all have higher coefficient scale weights for this operator, but when this is applied to a 1, that makes it look like an 8.

{
\subsection{Applying Augmentations for MNIST Classification}
To test how well the learned operators represent data variations and to examine the potential for using transport operators in downstream tasks, we enrich a limited MNIST dataset using the learned operators and compare against other common augmentation strategies. We train a LeNet-5 classifier \cite{lenet5} using 10 examples per class, sampled with replacement at each training step with a random augmentation applied. This experimental set up was intentionally selected to make MNIST more challenging to demonstrate the potential benefit of augmentations. We compare transport operator-based augmentations against two common image space augmentations, a modified version of RandAugment\footnote{Since RandAugment requires 3 image channels, we repeat grayscale MNIST images 3 times over the channel dimension, apply RandAugment with N=2 and M=6, and select a random channel as our training example. We refer to this as "Modified RandAugment".} \cite{randaug} and elastic distortion, as described in \cite{elastic_distortion}. To test the effectiveness of our learned coefficient encoder, we compare against transport operator augmentations sampled either from a fixed prior or a coefficient encoder trained following the methodology in Section~\ref{sec:coeffEnc}. To prevent overfitting due to limited training data, we apply a weight decay of $1\mathrm{e}{-4}$ and early stopping after $10,000$ iterations, measuring accuracy over the full test dataset over multiple trials.

Results can be found in Table~\ref{table:mnist_clf}, where it can be seen that augmentations generated using transport operators controlled by coefficients sampled from the learned coefficient encoder produce the highest classification accuracy. We observe higher variance in the performance of RandAugment and elastic distortion, which apply augmentations that can alter the identity of the image. We note that the coefficient encoder uses supervision to learn the extent with which the transport operators can be applied, as opposed to the fully unsupervised comparison methods. However, this experiment demonstrates that the learned transport operators effectively represent the data variations and the coefficient encoder enables effective augmentations. Future work can extend the coefficient encoder to  settings with limited (e.g., semi-supervised) or no labels.

\begin{table}[]
\caption{\label{table:mnist_clf}Test Accuracy training a LeNet-5 on MNIST with 10 examples per class with different augmentations applied at each training step. Among augmentation strategies, transport operator augmentations sampled from the coefficient encoder has the highest accuracy.}
\centering
\begin{tabular}{cc}
\hline
Model                                                         & \multicolumn{1}{l}{Test Accuracy} \\ \hline
No Augmentations                                              & $81.49 \pm 3.35\%$                      \\
Modified RandAugment                                                   & $87.94 \pm 3.34\%$                      \\
Elastic Distortion                                            & $80.40 \pm 4.02\%$                      \\
Transport Operator w/ Fixed Prior                             & $70.35 \pm 1.70\%$                      \\
Transport Operator w/ Coefficient Encoder & $93.57 \pm 0.67\%$                      \\ \hline
\end{tabular}
\end{table}

}

\section{Conclusion}

In this work, we develop methods to improve the effectiveness and utility of Lie group operator models for learning manifold representations of datasets with complex transformations that cannot be labeled. We do this by introducing a perceptual point pair selection strategy for training operators and by developing a method that uses a pretrained classifier to learn local regions where operators are likely to be used while preserving the identity of transformed samples. We demonstrate the efficacy of our approach in learning natural dataset variations with the MAE. While this is a powerful model, users should be mindful of the biases that are introduced through training with specific supervision methods when drawing conclusions about learned data transformations. This work presents a promising technique for learning representations of natural data variations that can improve model robustness. In future work, we can address some limitations by expanding this work to consider methods for localizing manifold structure with limited or no class labels and improving the reconstruction fidelity by applying the model within more complex generative model embeddings.

\clearpage

\bibliography{references.bib}
\bibliographystyle{tmlr}
\raggedbottom

\pagebreak
\appendix

\section{Coefficient Encoder Details}\label{app:coeffEnc}
\begin{figure}[ht]
\centering

\includegraphics[width=0.70\textwidth]{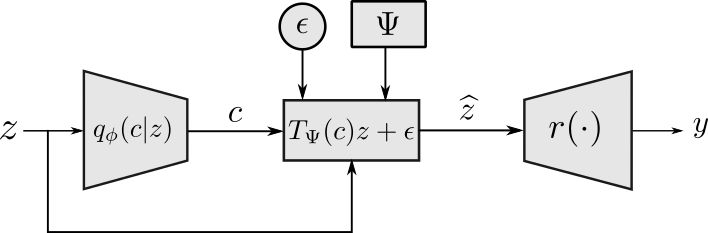}

	\caption{\label{fig:MAENet} System diagram for learning local coefficient statistics.}

\end{figure}

Our motivation in training an encoder to learn the coefficient statistics is to understand the regions of the manifold in which specific transport operators are applicable and the invariances of a pretrained classifier to  the learned transport operator-induced transformations. In this section, we will outline a derivation of our coefficient encoder training objective from a variational inference perspective by learning a variational posterior for the coefficients using a deep neural network \citep{kingma2013auto, rezende2014stochastic}. Given observations $(\vtr{z}_{i}, y_{i})$ for $i = 1, \dots, N$, and a pre-trained classifier $r(\cdot)$, we aim to learn a distribution $q_\phi (\vtr{c} | \vtr{z})$ from which we can sample to define the transformation $T_\Psi(\vtr{c})$ that can be applied to $\vtr{z}$ while maintaining class identity. In other words, we want the augmented $\vtr{\widehat{z}} = T_\Psi(\vtr{c})\vtr{z} + \epsilon$ to result in $r(\widehat{\vtr{z}}) = y$.

To encourage identity-preservation of vectors transformed with transport operators that are controlled by sampled coefficients, we maximize the likelihood of a class output for an observation under the system model diagrammed in Figure~\ref{fig:MAENet}. This system connects the observed latent vector $\vtr{z}$ to an output $y$ through a transformation defined by $\vtr{c}$. We follow the derivation used in Rezende \& Mohamed \citep{rezende2016variational} which introduces the variational posterior to estimate the log likelihood of the data. 

Consider a parameterized distribution for the conditional likelihood of our observations:
\begin{align}
\log p_{\theta}(y \mid \vtr{z}) & = \log \mathbb{E}_{p_{\zeta}(\vtr{c})} \left [p_{\theta}(y \mid \vtr{c}, \vtr{z}) \right]\label{eq:marginC} \\
& = \log \int_{\vtr{c}}p_{\zeta}(\vtr{c}) p_{\theta}(y \mid \vtr{c}, \vtr{z}) d\vtr{c} \label{eq:marginInt}\\
& = \log \int_{\vtr{c}}q_\phi(\vtr{c} \mid \vtr{z}) \frac{p_{\zeta}(\vtr{c})}{q_\phi(\vtr{c} \mid \vtr{z})} p_{\theta}(y \mid \vtr{c}, \vtr{z}) d\vtr{c} \label{eq:introVI}\\
& \geq \int_{\vtr{c}}q_\phi(\vtr{c} \mid \vtr{z}) \log \left [\frac{p_{\zeta}(\vtr{c})}{q_\phi(\vtr{c} \mid \vtr{z})} p_{\theta}(y \mid \vtr{c}, \vtr{z}) \right]d\vtr{c} \label{eq:lowerBound}\\
& = E_{q_\phi} \left [\log p_{\theta}(y \mid \vtr{c}, \vtr{z}) \right ] - D_{KL}\left( q_\phi(\vtr{c} \mid \vtr{z}) | p_\zeta(\vtr{c}) \right) \\
& = E_{q_\phi} \left [\log E_{\epsilon} p_{\theta}(y \mid \vtr{c}, \vtr{z}, \epsilon, \widehat{\vtr{z}}) \right ] - D_{KL}\left( q_\phi(\vtr{c} \mid \vtr{z}) | p_\zeta(\vtr{c}) \right).\label{eq:reparamExp}
\end{align}

In~\eqref{eq:marginC} and \eqref{eq:marginInt}, we marginalize over the coefficients $\vtr{c}$ as required under our system model in Figure~\ref{fig:MAENet}. In \eqref{eq:lowerBound}, we lower bound the conditional likelihood with a variational lower bound derived from Jensen's inequality. Finally, we use the reparameterization trick to define $\widehat{\vtr{z}}$ as a deterministic function of $\vtr{c}$, $\vtr{z}$, and the parameter-free random variable $\epsilon$ in \eqref{eq:reparamExp}. 

In order to sample from $q_\phi(\vtr{c}\mid \vtr{z})$ when computing the expectations, we first use the reparameterization trick \citep{connor2021variational, kingma2013auto, rezende2014stochastic} to define the sampled coefficients $\vtr{\widehat{c}}$ as a function of a uniform random variable $\vtr{u} \sim~\mathrm{Unif}\left(-\frac12,\frac12\right)^M$:
\begin{equation}
    \vtr{\widehat{c}} = l_{\phi}(\vtr{u}, \vtr{z}) = -h_\phi(\vtr{z}) \operatorname{sgn}(\vtr{u})\log(1-2\abs{\vtr{u}}),
\end{equation}
where $l_{\phi}$ is a defined mapping from a uniform distribution to a Laplace distribution and the Laplace scale parameters are defined by the output of the coefficient encoder $h_\phi$ that we aim to learn.

Using the reparameterization of $\vtr{\widehat{c}}$ and $\vtr{\widehat{z}}$, we can define the expectation:

\begin{align}
E_{q_\phi} \left [\log E_{\epsilon} p_{\theta}(y \mid \vtr{c}, \vtr{z}, \epsilon, \widehat{\vtr{z}}) \right ] & = E_{\vtr{u}} \left [\log E_{\epsilon} p_{\theta}(y \mid \widehat{\vtr{c}} = l_{\phi}(\vtr{u}, \vtr{z}), \epsilon, \widehat{\vtr{z}}) \right ] \\
& \approx \frac{1}{JK} \sum_{j=1}^J \log \sum_{k=1}^K p_{\theta}\left(y |\widehat{\vtr{c}}^{(j)} = l_{\phi}(\vtr{u}^{(j)}, \vtr{z}), \epsilon^{(k)}, \widehat{\vtr{z}}\right). 
\end{align}

Our pre-trained classifier defines $p_\theta$ which allows us to specify final objective using of $r(\cdot)$. Using a KL-divergence as the likelihood function between the ground truth labels $y_i$ and the classifier output for the augmented inputs $r(\widehat{\vtr{z}}^{(j)}_i)$ and simplifying the model by setting $\epsilon$ to $0$, we get:
\begin{equation}
\log p_{\theta}(y \mid \vtr{z}) \geq - \frac{1}{J}\sum_{j=1}^J D_{KL}\left( y_i \mid r(\widehat{\vtr{z}}^{(j)}_i) \right) - D_{KL}\left( q_\phi(\vtr{c} \mid \vtr{z}) | p_\zeta(\vtr{c}) \right),
\end{equation}
where in practice we use a single sample $J = 1$, resulting in the objective in~(\ref{eq:cEncObj}) which we want to minimize. The KL divergence between the $q_\phi(\vtr{c} \mid \vtr{z})$ and $p_\zeta(\vtr{c})$ has a closed form expression~\citep{gil2011renyi}:

\begin{equation}
    D_{KL}\left( q_\phi(\vtr{c}\mid \vtr{z}) \Vert p_\zeta(\vtr{c})\right) = \log(h_\phi(\vtr{z})) - \log(\zeta) + \frac{\zeta}{h_\phi(\vtr{z})} - 1
\end{equation}

\section{Coefficient Inference Details} \label{app:inference_details}

\begin{figure}[ht]
\centering

\includegraphics[width=0.4\textwidth]{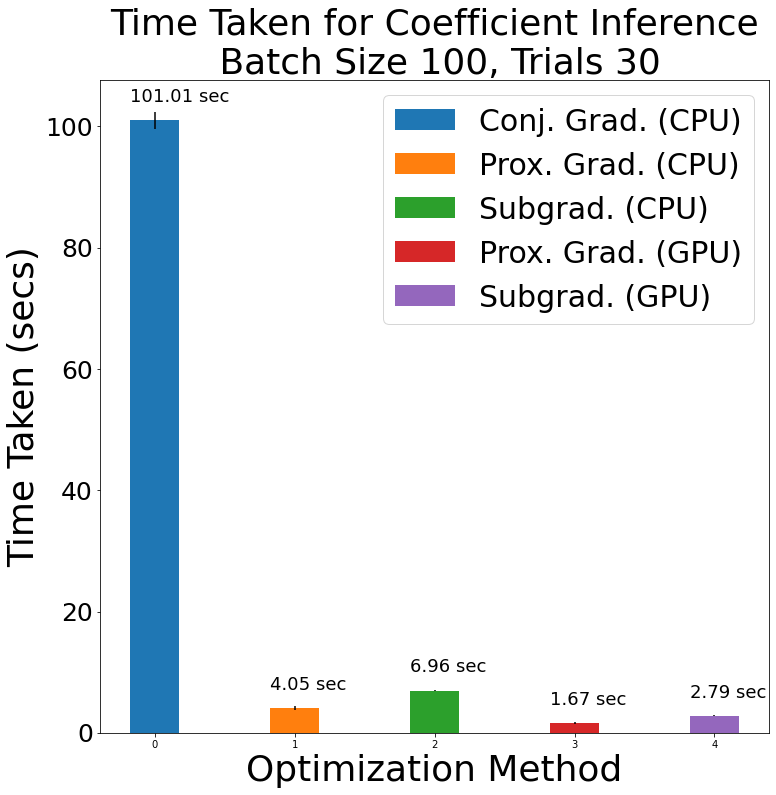}

	\caption{\label{fig:inf_comp} Comparison in time taken for various coefficient inference algorithms on fully trained transport operators. Each inference trial is performed over random batches of 100 point pairs drawn from the CelebA dataset (with a latent dimension of 32). 30 trials are run with the same $\vtr{c}_0$ and tolerance. We used the proximal gradient (GPU) method.}

\end{figure}

In order to perform coefficient inference, previous Lie group operator methods \citep{connor2020representing, connor2021variational} use the analytic gradient proposed in \citep{culpepper2009learning} with a conjugate gradient descent solver. This method requires an eigenvalue decomposition for computing the gradient which is not favorable for parallel computations. Alternatively, we use the PyTorch implementation of the matrix exponential that allows for automatic differentiation. Furthermore, to handle the non-smooth $\ell_1$ norm in objective~(\ref{eq:objFun}), we apply a proximal gradient step in a forward-backward splitting scheme \citep{beck_fast_2009}. For the $\ell_1$ norm, the proximal gradient has a known, closed-form solution in the soft-thresholding function \citep{10.1561/2400000003}:
\begin{equation}
    \mathcal{T}_\lambda(\vtr{c}) =  \mbox{sign}(\vtr{c})* \max \Bigl (|\vtr{c}| - \lambda, 0 \Bigr)
\end{equation}

Let $\nabla_{\vtr{c}} \frac{1}{2}\left\|\vtr{z}_1 - \mathrm{expm}\left(\sum_{m=1}^M{\mtx{\Psi}_mc_m} \right) \vtr{z}_0\right\|_2^2 \approx \nabla \widetilde{f}(\vtr{c})$ be a numerical approximation to the gradient of the $\ell_2$ term, found through automatic differentiation. Given initial coefficient values $\vtr{c}_0$ drawn from an isotropic Gaussian with variance $4 \times 10^{-4}$, our gradient descent step is:
\begin{equation}
    \vtr{c}_{k+1}=  \mathcal{T}_{\zeta \alpha_k}\left(\vtr{c}_k - \alpha_k \nabla \widetilde{f}(\vtr{c}_k) \right) 
\end{equation}
These steps are iterated upon until either a max iteration count is reached, or the change in coefficients, $\Vert \vtr{c}_{k+1} - \vtr{c}_{k} \Vert_2$, falls below some threshold. For all experiments we use a max iteration count of 800 and a tolerance of $1\times 10^{-5}$. For our step-size, we use $\alpha_k = (0.985)^k \alpha_0$ with $\alpha_0 = 1\times 10^{-2}$. We experimented with different acceleration methods \citep{kingma2014adam, beck_fast_2009}, and found that in many cases they resulted in worse performance.

The main parameter to select for inference is the sparsity-inducing parameter on the coefficient prior, $\zeta$. This value is sensitive to the dataset, latent dimension, and $\gamma$ hyper-parameter value. In practice, we fix a $\gamma$ value and pick $\zeta$ to be as high as possible, inducing sparsity in the inferred coefficients, while maintaining good reconstruction performance of $\widehat{\vtr{z}}_1$. If $\zeta$ is too large, that will result in all coefficients going to zero, preventing proper reconstructive performance. On the other hand, setting $\zeta$ to be too small will result in all operators being used to represent the transformation between each point pair, preventing any meaningful structure from being learned during training. Parameters used in specific training runs are included in their respective Appendix sections.

When compared against the coefficient inference implementation from~\citep{connor2020representing}, we perform inference for a batch of 100 samples in an average of 1.67 seconds over 100 trials whereas the inference from~\citep{connor2020representing} took an average of 101.01 seconds. We also compare the speed-up over a baseline that strictly uses automatic differentiation (subgradients) and compare the benefits from moving to GPU hardware in Figure~\ref{fig:inf_comp}. We use the proximal gradient (GPU) method from this figure. Experiments were run on a machine with an Intel i7-6700 CPU with 4.00 GHz and a Nvidia TITAN RTX.

\section{Training Strategy}

In all experiments, we follow the general training procedure put forth previously~\citep{connor2020representing}. We train the MAE with three training phases: the autoencoder training phase, the transport operator training phase, and the fine-tuning phase. During the autoencoder training phase, the network weights are updated using a reconstruction loss objective: $E_\text{AE} = \left\|\vtr{x} - \hat{\vtr{x}}\right\|_2^2$. 

During the transport operator training phase, the network weights are fixed and the transport operators are trained between pairs of points using the objective (\ref{eq:objFun}). Pairs of images $x_0, x_1$ are chosen using the perceptual point pair selection strategy described in Section~\ref{sec:pointPair}. The images are then encoded into the latent space $z_0, z_1$. For each batch, the first step is to infer the coefficients between all pairs of latent vectors. Coefficient inference is best performed when the entries of the latent vectors are close to the range $[-1, 1]$. Because of this, we define a scale factor that can be applied to encoded latent vectors to reduce the magnitude of their entries prior to performing coefficient inference. In practice, we inspect the latent vector magnitudes after the autoencoder training phase and choose a scale that will adjust the magnitudes of the latent vector entries to be in the range $[-1,1]$. This does not have to be a precise range for the latent vector magnitudes but instead is a practical guideline. Coefficient inference is performed on the scaled latent vectors as described in Sections~\ref{sec:coeffInfer} and Appendix~\ref{app:inference_details}. After the coefficients are inferred for a batch, the weights on the dictionary elements are updated. This phase of training is performed until the loss values reach a plateau and the dictionary magnitudes plateau.

The fine-tuning training phase begins after the transport operator training phase is complete. In this phase, both the network weights and the transport operator weights are updated using the joint objective: 
\begin{equation}\label{eq:fullObj}
E = \lambda\left(\left\|\vtr{x}_0 - \hat{\vtr{x}}_0\right\|_2^2 + \left\|\vtr{x}_1 - \hat{\vtr{x}}_1\right\|_2^2\right) +  \left(1-\lambda\right)E_{\Psi},
\end{equation}
where $E_{\Psi}$ is defined in (\ref{eq:objFun}). Using this objective, we alternate between taking steps on the transport operator weights while the network weights are fixed and taking steps on the network weights while the transport operator weights are fixed. Additionally, during the fine-tuning phase we incorporate occasional steps in which we update the network weights using only the reconstruction loss to ensure effective image reconstruction.

In most cases, it is necessary to reduce the $\gamma$ parameter in front of the Frobenius norm dictionary regularizer prior to fine-tuning or the dictionary magnitudes will reduce to zero. We report the $\gamma$ we use for transport operator training and fine-tuning in the experimental details sections below. It may also be necessary to decrease the network learning rate during fine-tuning.

The coefficient encoder training requires a network that is trained to classify data from our selected dataset. We train this classifier using training data from a given dataset. For datasets with worse autoencoder reconstruction quality, we train the classifier in the latent space. Otherwise, we train the classifier on images in the data space. With the MAE and classifier trained, we train the coefficient encoder network following the strategy described in Section~\ref{sec:coeffEnc} and Appendix~\ref{app:coeffEnc}.

In order to train the CAE, we use the same autoencoder architecture used with the MAE with the addition of a Frobenius norm regularizer on the encoder Jacobian, weighted by a selected $\lambda$ value (different from the $\lambda$ in \eqref{eq:fullObj}). The Jacobian is computed using PyTorch automatic differentiation. We find the Jacobian norm decreases to the same value irrespective of our choice of $\lambda$, leading us to choose $\lambda = 1$. For the $\beta$-VAE we use the same architectures outlined in \citep{higgins2017beta} with $\beta=10$ for CelebA and $\beta=5$ for MNIST and Fashion MNIST. We find that setting $\beta$ any higher results in poor reconstructive performance. Scripts for training both comparison methods are included in the code repository.

Hyper-parameter tuning for all experiments was performed on the Georgia Tech Partnership for Advanced Computing Environment (PACE) clusters \citep{PACE}. Experiments were performed using a Nvidia Quadro RTX 6000. Runs training the CAE, and $\beta$-VAE on CelebA were all run on a separate machine with a Nvidia TITAN RTX.

\section{Parameter Selection}

The MAE model has several hyperparameters that must be tuned and we will provide guidance to determining ideal parameter values for our experiments and future experiments. First, we will describe some signs to look out for to identify if a run is succeeding or failing. One indicator that we compute is the transport operator difference which is:

\begin{equation}
   E_{\Psi\text{diff}} =  \frac{1}{2}\left\|\vtr{z}_1 - \mathrm{expm}\left(\sum_{m=1}^M{\mtx{\Psi}_mc_m} \right) \vtr{z}_0\right\|_2^2 - \frac{1}{2}\left\|\vtr{z}_1 - \mathrm{expm}\left(\sum_{m=1}^M{\widehat{\mtx{\Psi}}_mc_m} \right) \vtr{z}_0\right\|_2^2,
\end{equation}
where $\widehat{\mtx{\Psi}}_m$ is the dictionary after the gradient step is taken. A gradient step should decrease the transport operator objective meaning the $E_{\Psi\text{diff}}$ value should be positive for an effective step. If there are many gradient steps that result in negative $E_{\Psi\text{diff}}$ values, that indicates that the parameters are not optimal or that the learning rate is too large. This is an important metric to observe during fine-tuning to determine whether to select a smaller $\gamma$ or smaller network learning rate.

Signs of failure of a training run with selected training parameters:
\begin{itemize}
    \item All operator magnitudes reduce towards zero.
    \item All inferred coefficients between point pairs are zero.
    \item Most of the operators generate transformation paths with latent values that increase quickly to infinity.
    \item Many steps have negative values for $E_{\Psi\text{diff}}$.
    \item The operator magnitudes increase which results in unstable training steps with NaN values in the computed objective.
\end{itemize}

\paragraph{Dictionary regularizer parameter} The dictionary regularizer parameter $\gamma$ is the weight on the Frobenius norm term in \eqref{eq:objFun}. This objective term serves two purposes. First, it balances the effect of the coefficient sparsity regularizer with the parameter $\zeta$. If $\zeta$ is large, the sparsity regularizer encourages small coefficient magnitudes and one way to achieve that while still effectively inferring paths is to increase the magnitude of the operators. The Frobenius norm term must have a large enough influence to counterbalance this force or the operators will increase in magnitude to the point of being unstable. If a run is becoming unstable, we recommend decreasing $\zeta$ or increasing $\gamma$.

The second purpose of the dictionary regularizer term is to identify which operators are necessary to represent the transformations on the data manifold. If an operator is not being used to represent a transformation between $\vtr{z}_0$ and $\vtr{z}_1$ then the dictionary regularizer reduces its magnitude to zero. Therefore, during training we are able to estimate the model order based on how many dictionary elements remain non-zero. In our tests we vary $\gamma$ between $2 \times 10^{-8}$ and $2 \times 10^{-4}$. If $\gamma$ is too small, it will not counterbalance the coefficient sparsity term and the dictionaries will grow to unstable magnitudes. If $\gamma$ is too large it will reduce the magnitude of all operators to zero. During the fine-tuning steps, we have often found it necessary to reduce the $\gamma$ because, with a larger $\gamma$, both the operator magnitudes and the latent vector magnitudes can decrease substantially, which leads to an ineffective manifold model.

\paragraph{Coefficient sparsity parameter} The coefficient sparsity parameter $\zeta$ in~\eqref{eq:objFun} controls the sparsity of the coefficients that are used to estimate paths between $\vtr{z}_0$ and $\vtr{z}_1$. We run tests with values of $\zeta$ between 0.005 and 2. From dataset to dataset the ideal value varies. If $\zeta$ values are too small then all the operators are used to represent all the paths between point pairs. The $\zeta$ should be increased so fewer than $M$ coefficients are used for each inferred path. When $\zeta$ is too large, all the coefficients go to zero during inference. This means there is no path inferred between $\vtr{z}_0$ and $\vtr{z}_1$ because of overweighting the sparsity constraint. 

\paragraph{Number of dictionary elements} As mentioned above, the dictionary regularizer acts as a model order selection tool so our strategy for selecting number of dictionary elements $M$ is to increase the number of dictionary elements until some of their magnitudes begin reducing to zero during training. This indicates that some of the operators are not necessary for representing transformations.

\paragraph{Latent dimension} The latent dimension of the autoencoder is selected to ensure quality reconstructed image outputs.

\paragraph{Relative weight of reconstruction and transport operator objectives} The parameter $\lambda$ determines the weight of the reconstruction term relative to the transport operator term. We observe good performance of the model for $\lambda$ between 0.5 and 0.75.

\section{Analysis on the Effect of Coefficient Sparsity}\label{app:sparsity_analysis}

\begin{figure}[h]
\centering
\begin{subfigure}{0.3\textwidth}
  \centering
	{\includegraphics[width=0.98\textwidth]{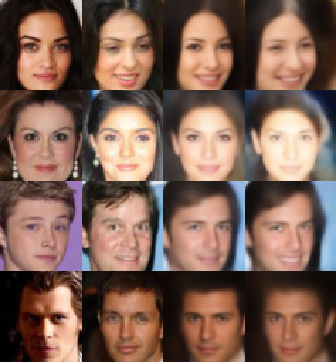}}
  \caption{Sparse Coefficients}
	\label{subfig:sparse_recon}
\end{subfigure}
\begin{subfigure}{0.3\textwidth}
  \centering
	{\includegraphics[width=0.98\textwidth]{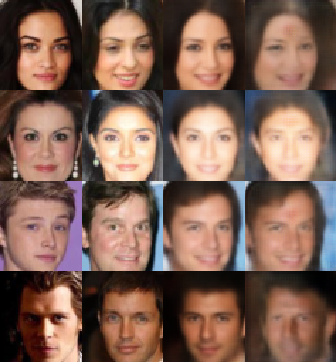}}%
  \caption{Dense Coefficients}
	\label{subfig:dense_recon}
\end{subfigure}
\caption{\label{fig:dense_sparse_recon} Qualitative comparison of reconstructing transport operator transformation paths when trained with (a) sparse coefficients and (b) dense coefficients. For each row, the first two columns denote a point pair ($\mathbf{x}_0$ and $\mathbf{x}_1$), the third column shows a reconstructed estimate of the second column $\mathbf{\widehat{x}}_1$ after applying transport operators with inferred coefficients $\mathbf{c}$, and the fourth column shows a reconstructed estimate after applying transport operators with a 2 times extrapolation of the inferred coefficients $2\mathbf{c}$.}
\end{figure}

In our work, we impose sparsity in coefficients $\mathbf{c}$ that describe the manifold transformation between a point pair in the latent space. Sparse coding has been used to encourage learned representations that are statistically independent \cite{olshausen1997sparse}. In our context, this means that we expect each transport operator to transform a single semantic attribute in an image. 

We investigate the effects of the sparsity penalty by comparing to transport operators trained using dense coefficients (i.e., all coefficients non-zero for every point pair). To train our dense coefficient model, we follow the exact methodology from the main text on CelebA, replacing our coefficient inference step with the forward pass of a learned DNN. We use a fully-connected network with two hidden layers that takes the features of the point pairs concatenated as input (e.g., $[\mathbf{z}_0, \mathbf{z}_1]$) and coefficients as output, similar to the method used in \cite{ibrahim2022}. To prevent coefficients from growing too large, we also apply an $\ell_2$ penalty on our coefficients with a weight of $1\mathrm{e}{-3}$. We train our auto-encoder and coefficient inference network together end-to-end using automatic differentiation.

We first compare the qualitative reconstruction capabilities of sparse and dense coefficients. For both methods, we randomly sample $5,000$ point pairs and perform coefficient inference to estimate transport operator paths. Averaged across all point pairs, the dense and sparse coefficients result in a transport operator error of $2.804\mathrm{e}{-4}$ and $1.023\mathrm{e}{-3}$, respectively. This is to be expected, since the sparse coefficients are constrained to use fewer operators for every point pair, mirroring the intuition of higher training MSE in LASSO over unconstrained least-squares methods. We show qualitative reconstructions in Figure~\ref{fig:dense_sparse_recon}, where the fourth column denotes the reconstruction of extrapolated paths. Extrapolated paths with sparse coefficients have higher reconstruction quality, attributed to  better operator stability (see Appendix~\ref{app:pp_supervision}). To measure this, we take the eigenvalue with the largest real component from each operator and normalize by the average coefficient magnitude across the $5,000$ evaluation point pairs. The median of the largest normalized eigenvalue over all operators was $3.08\mathrm{e}{-2}$ and $1.06\mathrm{e}{-2}$, for operators trained with dense and sparse coefficients, respectively.

\begin{figure}[h]
\centering
\begin{subfigure}{0.48\textwidth}
  \centering
	{\includegraphics[width=0.98\textwidth]{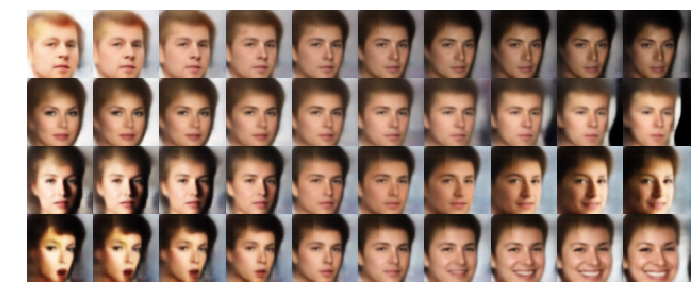}}
  \caption{Sparse Coefficients}
	\label{subfig:sparse_interp}
\end{subfigure}
\begin{subfigure}{0.48\textwidth}
  \centering
	{\includegraphics[width=0.98\textwidth]{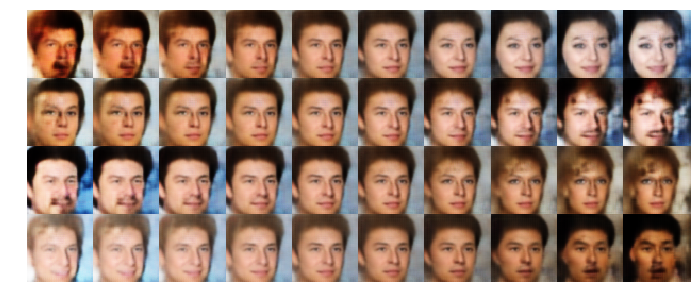}}%
  \caption{Dense Coefficients}
	\label{subfig:dense_interp}
\end{subfigure}
\caption{\label{fig:dense_sparse_interp} Transport operator augmentations applied by interpolating coefficients within a fixed range for (a) sparse coefficients and (b) dense coefficients, as in Figure~\ref{fig:qualTransform}. Each row denotes a different operator applied to an input point. From left to right, each column denotes a single operator applied to the same input point, with coefficients interpolated between $[-N_c, \dots, N_c]$.}
\end{figure}

Finally, we compare transformations for both methods by applying a single operator with coefficients within a fixed range (as in Figure~\ref{fig:qualTransform}). We select four operators from both methods and find the average coefficient magnitude across the $5,000$ evaluation point pairs. We then interpolate samples between -5 and 5 times this average magnitude, applying transformations from a single operator. The results are shown in Figure~\ref{fig:dense_sparse_interp}, where it can be seen that operators trained with sparse coefficients lead to more semantically meaningful transformations and higher reconstruction quality.

\section{Comparing Point Pair Selection Strategies}\label{app:pp_supervision}

\begin{figure}[ht]

  \centering
	{\includegraphics[width=0.65\textwidth]{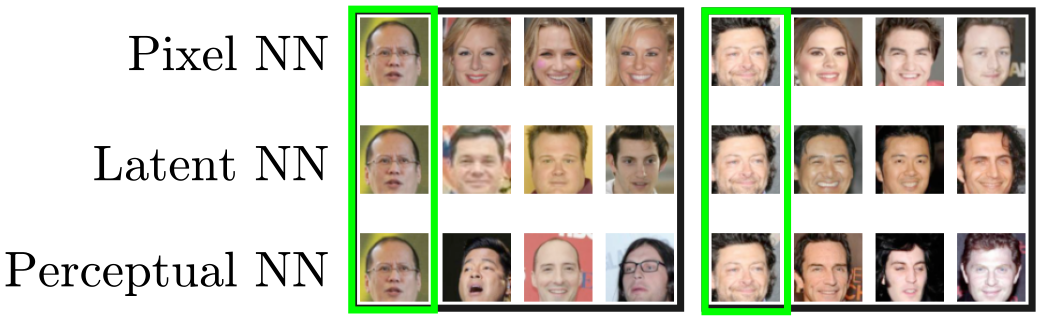}}

  \caption{\label{fig:NNComp} Examples of nearest neighbors identified through pixel similarity, latent similarity, and the perceptual loss metric. The images in the green boxes are the initial reference images and the images to the right of those show three selected nearest neighbors. The perceptual loss metric identifies neighbors that share similar characteristics like hair style without being exactly the same.}
	
\end{figure}

In the past, transport operators have been trained using point pairs that were selected randomly from the same class~\citep{connor2021variational} or with some knowledge of transformation labels~\citep{connor2020representing}. In this work, we establish a method that learns natural transformations without requiring transformation labels using a perceptual point pair selection strategy. The perceptual loss metric enables us to select point pairs that may share semantically meaningful transformations without being exactly the same. Fig.~\ref{fig:NNComp} shows examples of nearest neighbors selected using pixel similarity, similarity in the latent space, and the perceptual loss metric. This highlights how the perceptual loss metric can be useful for identifying inputs with similar qualitative characteristics. For instance, on the left, the perceptual loss metric identifies another bald man as a nearest neighbor which has the similar characteristics of the initial image even though the exact image looks quite different. 

\begin{figure}[t]

\centering
\begin{subfigure}[b]{0.4\textwidth}
  \centering
	{\includegraphics[width=0.98\textwidth]{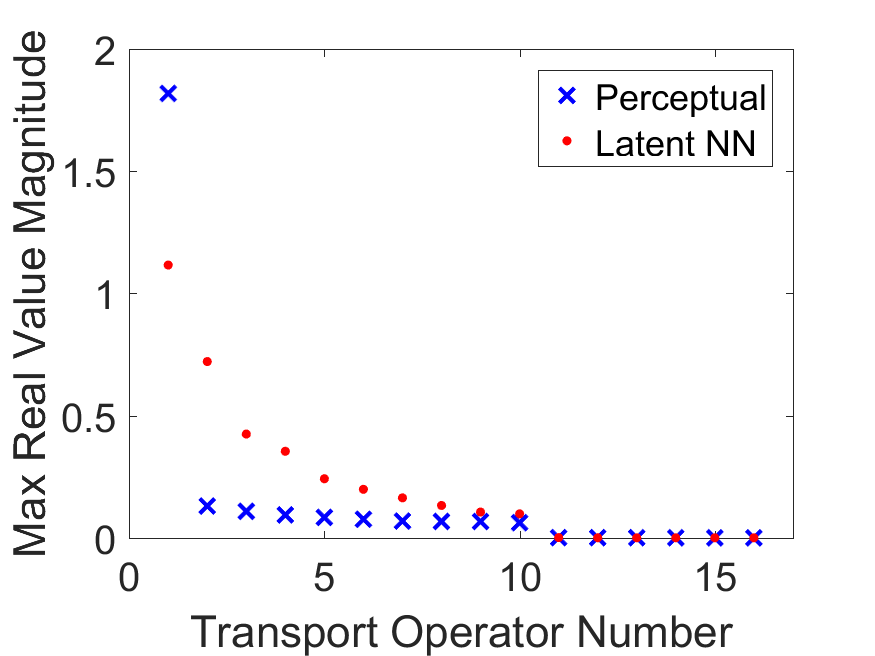}}
  \caption{}
	\label{subfig:NNeigMax}
\end{subfigure}
\begin{subfigure}[b]{0.4\textwidth}
 \centering
	{\includegraphics[width=0.98\textwidth]{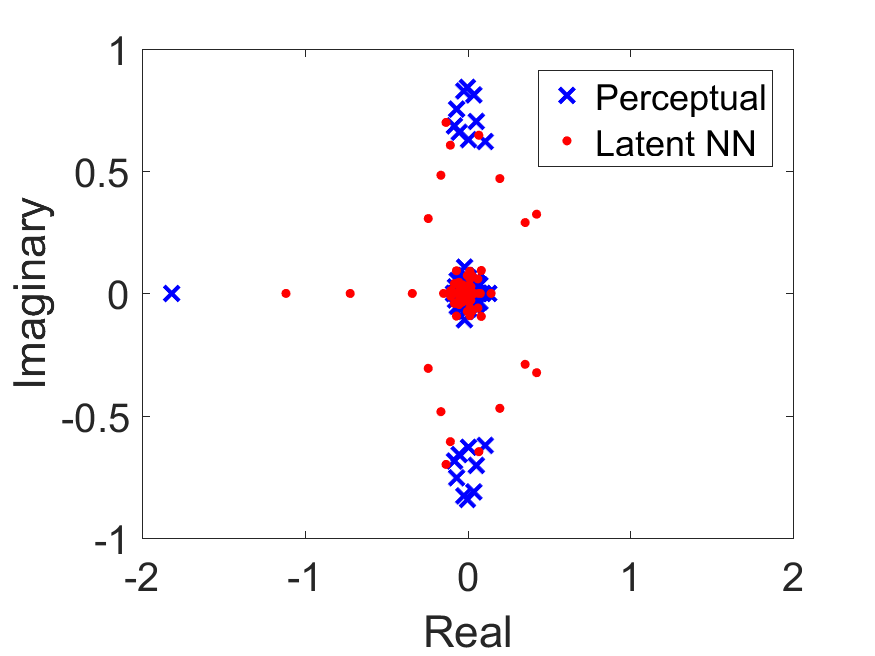}}
 \caption{}
	\label{subfig:NNeigPlot}
\end{subfigure}

  \caption{\label{fig:NNeig} Analysis of eigenvalues of operators learned in the MNIST experiment. (a) The maximum absolute value of the real parts of eigenvalues computed from each learned operator. (b) Plot of the real and imaginary parts of the eigenvalues for each operator.}
	
\end{figure}

When examining the success of learned operators, one characteristic we care about is how well the operators maintain stability of generated paths. We have defined this as a useful characteristic for identity-preservation of applied operators -- if operators expand latent vectors to infinity, that will very likely lead to a change in identity.  Each of the operators can be viewed as the dynamics matrix of a linear dynamical systems model and we can analyze their stability by observing their eigenvalues~\citep{strogatz:2000}. In our model, because the transport operators relate to natural data transformations without a temporal component, it does not matter if the transport operators are applied with positive or negative coefficients. Therefore, general stability or instability of the dynamical system associated with a given transport operator has the same impact of causing the latent vectors to increase to infinity because our coefficients can be either positive- or negative-valued. In this setting, we define the stability of our transport operator system in the context of marginal stability of a dynamical system. Marginally stable dynamical systems generate cyclic transformations that never increase or decrease the magnitude of the inputs. A marginal stable system has only imaginary eigenvalues with no real parts~\citep{strogatz:2000}. Therefore, we can identify transport operators approaching marginal stability by investigating the magnitude of the real parts of their eigenvalues and specifically the eigenvalue with the maximum magnitude of its real part. When the maximum magnitude of the real part for all eigenvectors associated with an operator is close to zero, that indicates that the operator is closer to marginal stability. Therefore, we quantify the stability of transport operator paths by looking at the maximum magnitude of real parts of eigenvalues associated with each transport operator. Fig~\ref{subfig:NNeigMax}. shows the sorted maximum magnitude of the real parts of eigenvalues in each of the 16 operators learned in the MNIST experiment when using the perceptual point pair selection strategy and when using a simple strategy of selecting point pairs as nearest neighbors in the latent space. Except for one operator, all the operators trained using latent space similarity for supervision have larger maximum magnitudes of real eigenvalue components than the operators trained using the perceptual loss metric. This indicates that the operators trained using the latent space similarity to select training point pairs are farther from marginal stability and can be seen as less stable by our definition of transport operator stability.

\begin{figure}[ht]

\centering
\begin{subfigure}[b]{0.8\textwidth}
  \centering
	{\includegraphics[width=0.98\textwidth]{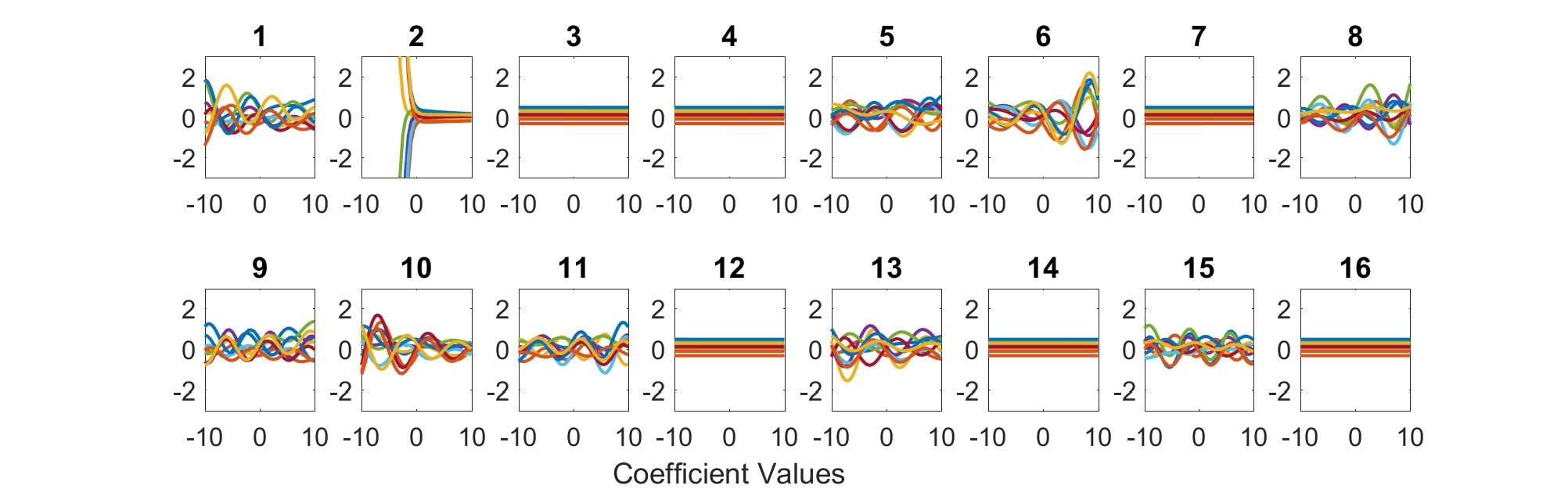}}
  \caption{}
	\label{subfig:NNpathVGG}
\end{subfigure}

\begin{subfigure}[b]{0.8\textwidth}
 \centering
	{\includegraphics[width=0.98\textwidth]{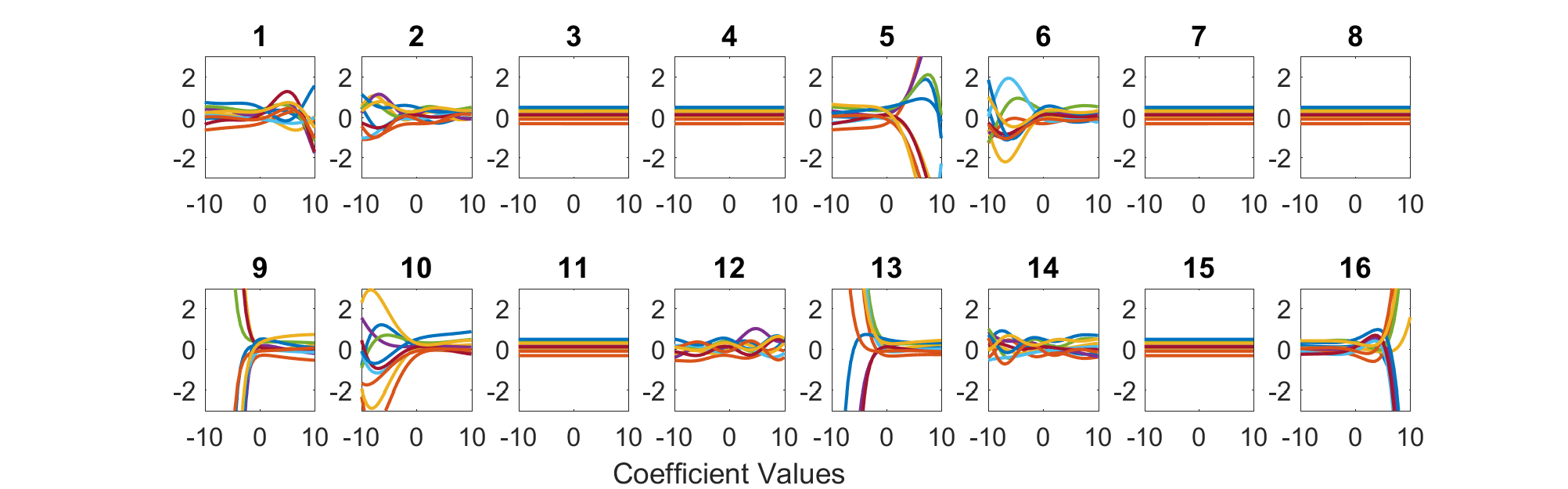}}
 \caption{}
	\label{subfig:NNpathNN}
\end{subfigure}

  \caption{\label{fig:NNPath} Visualizations of the effect of each learned transport operator on each dimension of an encoded latent vector. In each plot, each line represents a single latent dimension and the coefficient magnitude of the transformation varies on the x-axis. (a) Paths from operators learned with perceptual point pair supervision. (b) Paths from operators learned with point pairs selected as nearest neighbors in the latent space. }
	
\end{figure}

To view the effect of these operators more intuitively, we plot the values of the latent dimensions as individual operators are applied in Fig.~\ref{fig:NNPath}. The paths of the lines in each of these plots show a lot about how the operators influences each latent dimension. Fig~\ref{subfig:NNpathVGG} shows the paths generated by transport operators trained with the perceptual point pair selection strategy. The plots with straight lines show the effect of operators whose magnitudes are reduced to 0 during training. Most of the operators learned using the perceptual point pair selection strategy are close to cyclic except for operator 2 (the operator with the large real component magnitude in Fig.~\ref{subfig:NNeigMax}). In contrast, several of the operators trained with point pairs selected as nearest neighbors in the latent space extend to infinity (Fig~\ref{subfig:NNpathNN}). This can explain the larger maximum real value magnitudes in Fig.~\ref{subfig:NNeigMax}. This analysis leads us to conclude that the perceptual point pair selection strategy is more likely to yield identity-preserving transport operators.

\section{MNIST Experiment Details}

The MNIST dataset is made available under the terms of the Creative Commons Attribution-Share Alike 3.0 license. We split the MNIST dataset into training, validation, and testing sets. The training set contains 50,000 images from the traditional MNIST training set. The validation set is made up of the remaining 10,000 images. The traditional MNIST testing set is used for our testing set. The input images are normalized so their pixel values are between 0 and 1. The network architecture used for the autoencoder is shown in Table~\ref{tab:MNISTNet}. The training parameters for the transport operator training phase and the fine-tuning phase are shown in Tables~\ref{tab:MNIST_TOparams} and \ref{tab:MNIST_FTparams}.

Prior to training the coefficient encoder for the MNIST dataset, we train a classifier on the labeled MNIST image data which we use to encourage identity-preservation during coefficient encoder training. The training parameters for the coefficient encoder are shown in Table~\ref{tab:MNIST_CEparams}. The image classifier we use is based on the simple LeNet architecture with two convolutional layers and three fully connected layers~\citep{lecun1998gradient}.

\begin{table}[!htb]
\centering
\caption{Network Architecture for MNIST and Fashion MNIST Experiments}
\label{tab:MNISTNet}
\begin{tabular}{||l l||} 
 \hline
 Encoder Network & Decoder Network  \\ 
 \hline
 Input $\in \mathbb{R}^{28 \times 28}$ & Input $\in \mathbb{R}^2$  \\ 
 conv: chan: 64 , kern: 4, stride: 2, pad: 1  & Linear: 3136 Units  \\
 BatchNorm: feat: 64 & ReLU\\ 
 ReLU &  convTranpose: chan: 64, kern: 4, stride: 1, pad: 1\\
 conv: chan: 64, kern: 4, stride: 2, pad: 1  & BatchNorm: feat: 64 \\ 
 BatchNorm: feat: 64 & ReLU\\
 ReLU &  convTranpose: chann: 64, kern: 4, stride: 2, pad: 2 \\
 conv: chan: 64, kern: 4, stride: 1, pad: 0 &  BatchNorm: feat: 64\\ 
  BatchNorm: feat: 64 & ReLU \\
 ReLU & convTranpose: chan: 1, kernel: 4, stride: 2, pad: 1\\
 Linear: 2 Units & Sigmoid \\ 

 \hline
\end{tabular}

\end{table}

\begin{table}[!htb]
\centering
\caption{Training parameters for the transport operator training phase of the MNIST experiment}
\label{tab:MNIST_TOparams}
\begin{tabular}{||l||} 
 \hline
 MNIST Transport Operator Training Parameters \\ 
 \hline
 batch size: 250   \\ 
 autoencoder training epochs: 300   \\
 transport operator training epochs: 50   \\
 latent space dimension ($z_{dim}$): 10 \\
 $M:$ 16 \\
 $lr_{\mathrm{net}}: 10^{-4}$  \\
$lr_{\Psi}: 10^{-3}$ \\
 $\zeta:$ 0.1  \\
 $\gamma:$ $2 \times 10^{-6}$   \\
  initialization variance for $\Psi$: 0.05 \\
   number of restarts for coefficient inference: 1 \\
   nearest neighbor count: 5\\
   latent scale: 30\\
 \hline
\end{tabular}

\end{table}

\begin{table}[!htb]
\parbox{.50\linewidth}{
\centering
\caption{Training parameters for the fine-tuning phase of the MNIST experiment}
\label{tab:MNIST_FTparams}
\begin{tabular}{||l||} 
 \hline
 MNIST Fine-tuning Parameters \\ 
 \hline
 batch size: 250   \\ 
 transport operator training epochs: 100   \\
$lr_{\mathrm{net}}: 10^{-4}$  \\
$lr_{\Psi}: 10^{-3}$ \\
 $\zeta:$ 0.1  \\
 $\gamma:$ $2 \times 10^{-6}$   \\
 $\lambda$: 0.75 \\
  number of network update steps: 50  \\
  number of $\Psi$ update steps: 50 \\
 \hline
\end{tabular}
}
\parbox{.42\linewidth}{
\centering
\caption{Training parameters for the MNIST Coefficient Encoder}
\label{tab:MNIST_CEparams}
\begin{tabular}{||l||} 
 \hline
 MNIST Coefficient Encoder Parameters \\ 
 \hline
 batch size: 250   \\ 
 training epochs: 300   \\
lr: $10^{-3}$  \\
 $\zeta_\text{prior}:$ 0.1  \\
 $\lambda_\text{kl}$: 0.5 \\
  coefficient spread scale: 0.1  \\
  classifier domain: image \\
 \hline
\end{tabular}
}

\end{table}

\section{MNIST Experiment Additional Results}

\begin{figure}[!htb]
\centering

	{\includegraphics[width=0.4\textwidth]{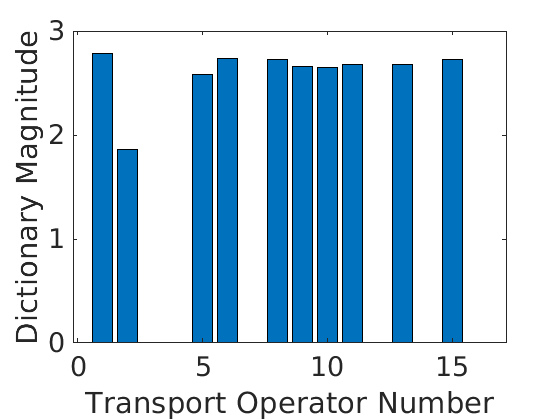}}

  \caption{\label{fig:mag_mnist} The magnitudes of the learned operators after fine-tuning.}
\end{figure}

\begin{figure}[!htb]

\centering
\begin{subfigure}[b]{0.24\textwidth}
  \centering
	{\includegraphics[width=0.98\textwidth]{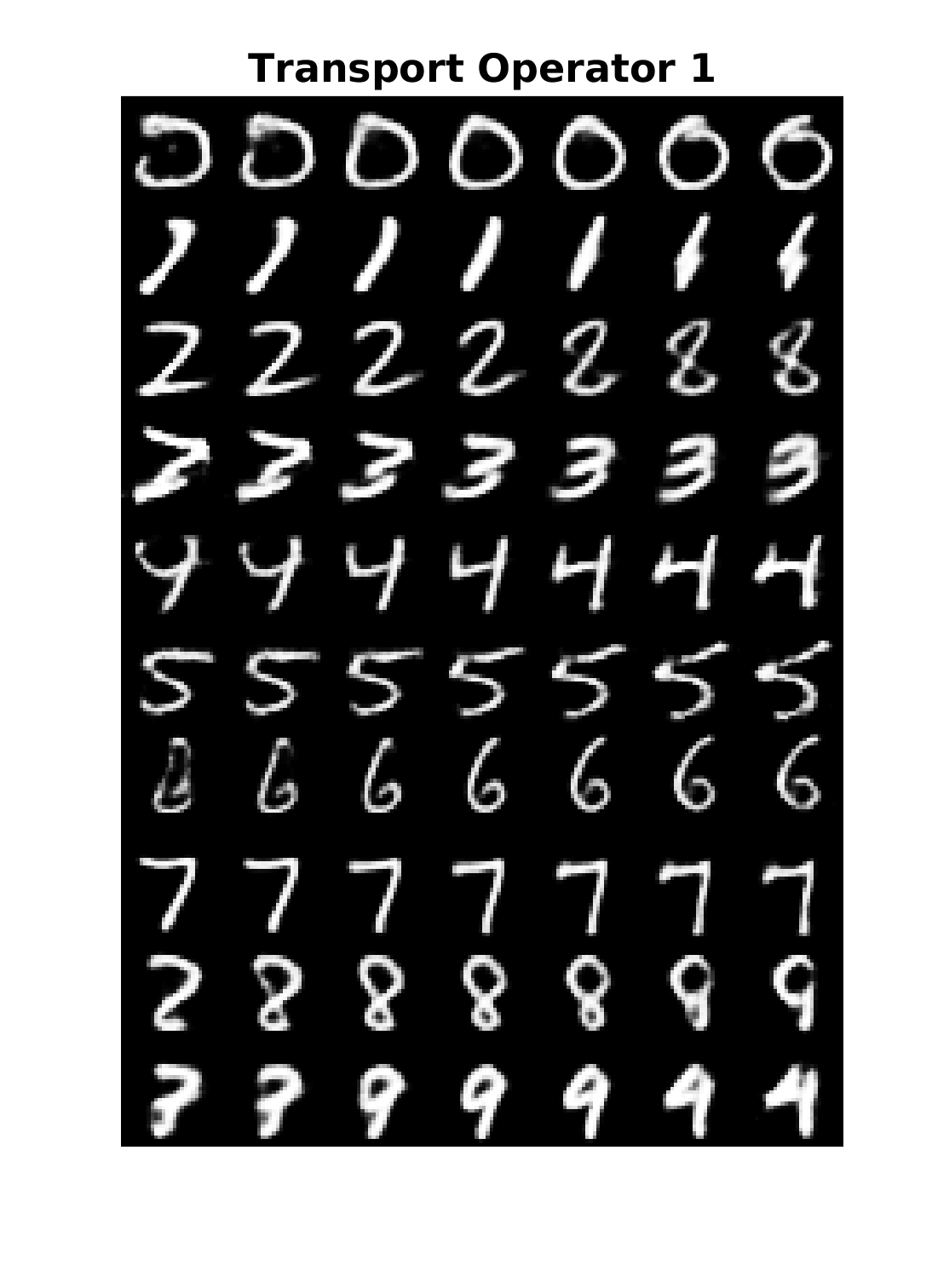}}
  \caption{}
\end{subfigure}
\begin{subfigure}[b]{0.24\textwidth}
 \centering
	{\includegraphics[width=0.98\textwidth]{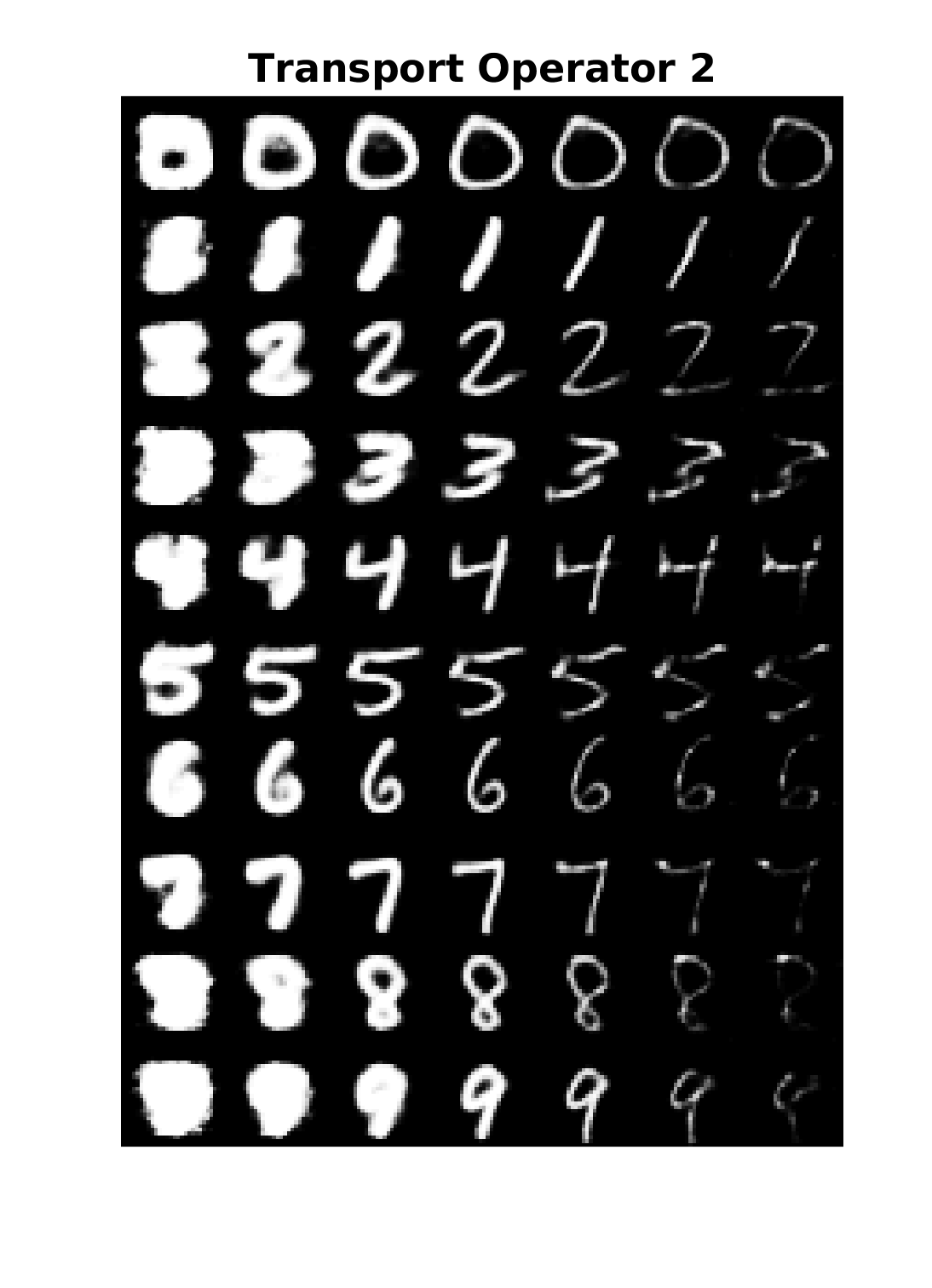}}
 \caption{}
\end{subfigure}
\begin{subfigure}[b]{0.24\textwidth}
  \centering
	{\includegraphics[width=0.98\textwidth]{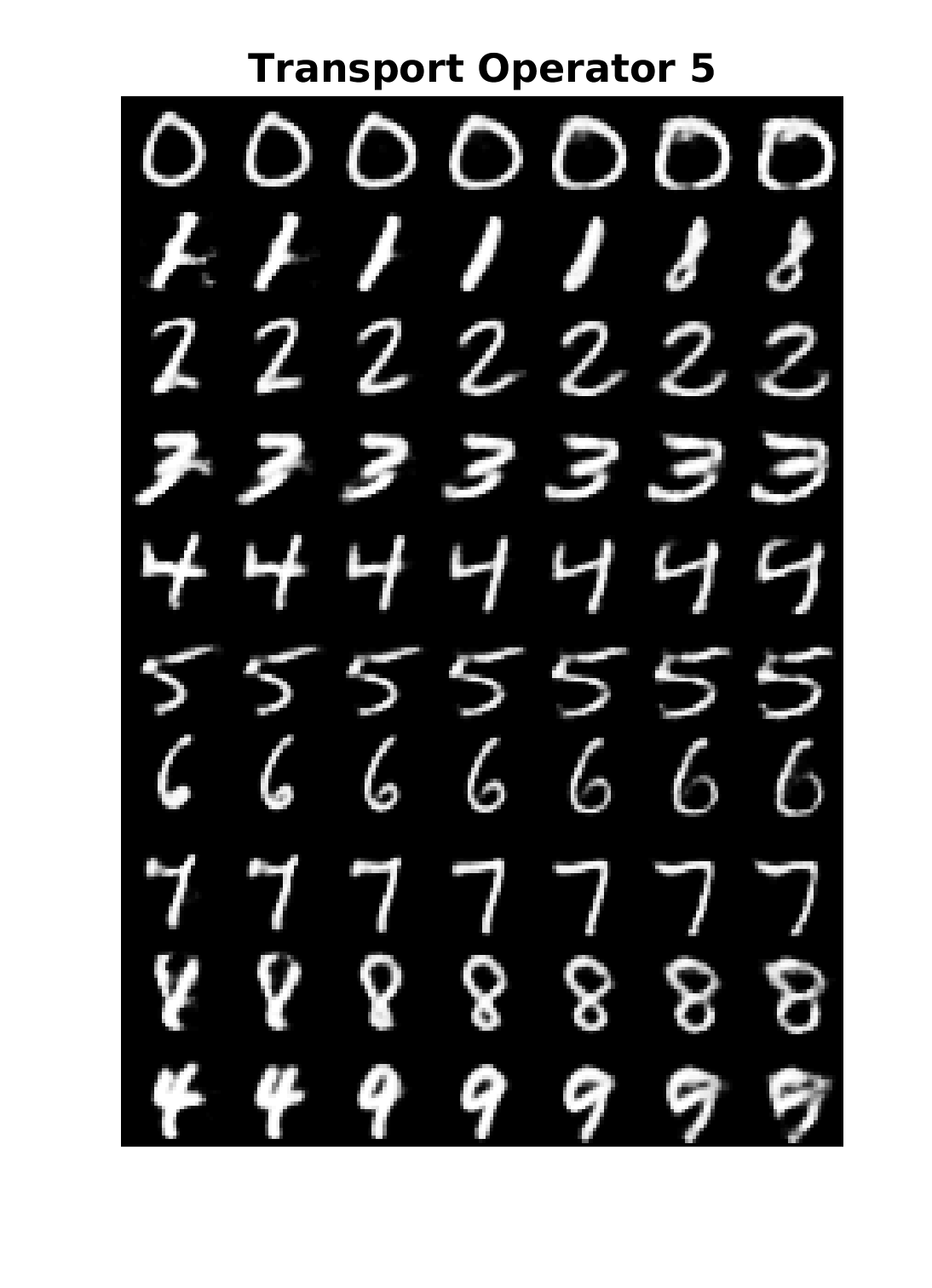}}
  \caption{}
\end{subfigure}
\begin{subfigure}[b]{0.24\textwidth}
 \centering
	{\includegraphics[width=0.98\textwidth]{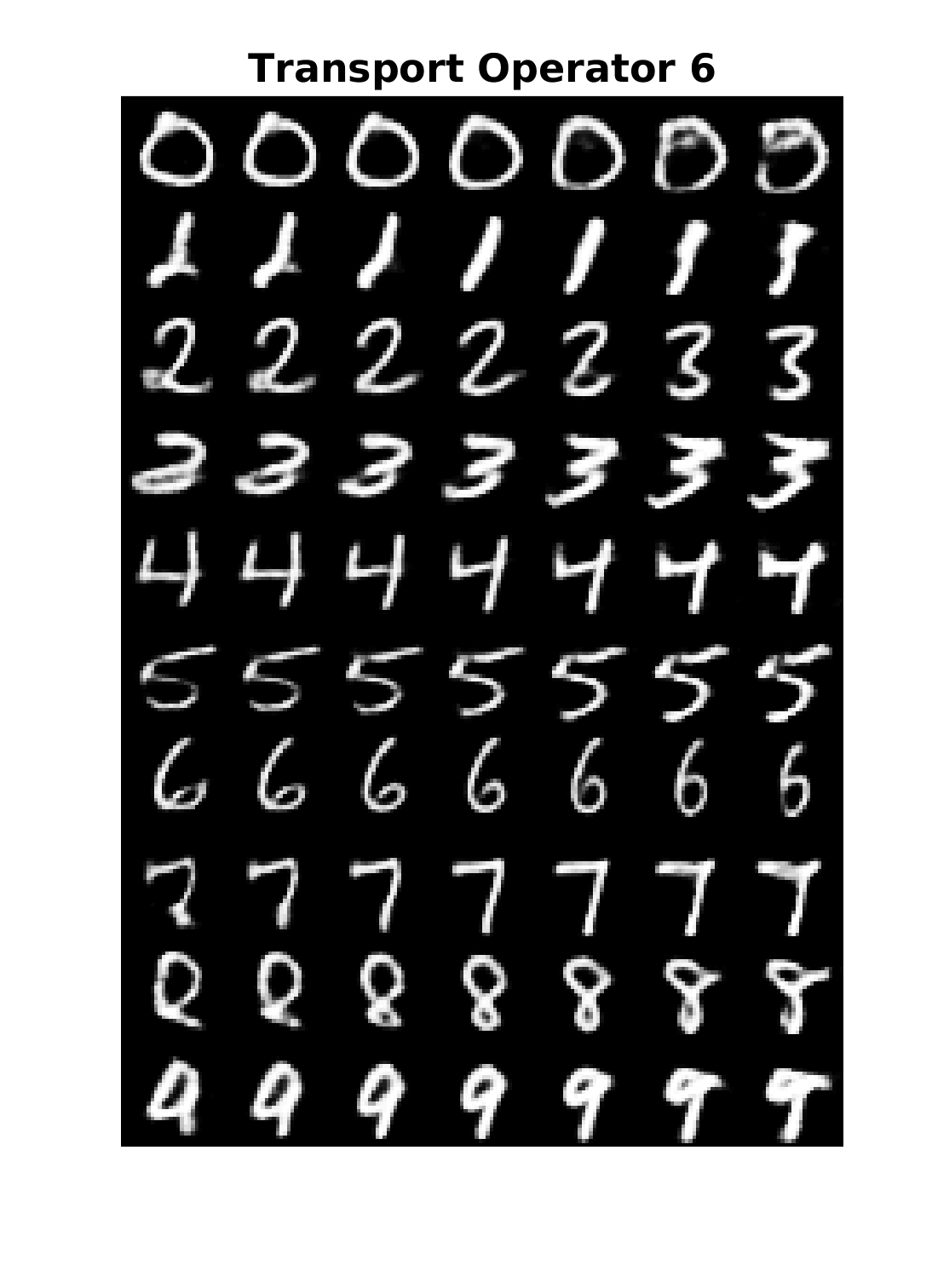}}
 \caption{}
\end{subfigure}

\begin{subfigure}[b]{0.24\textwidth}
  \centering
	{\includegraphics[width=0.98\textwidth]{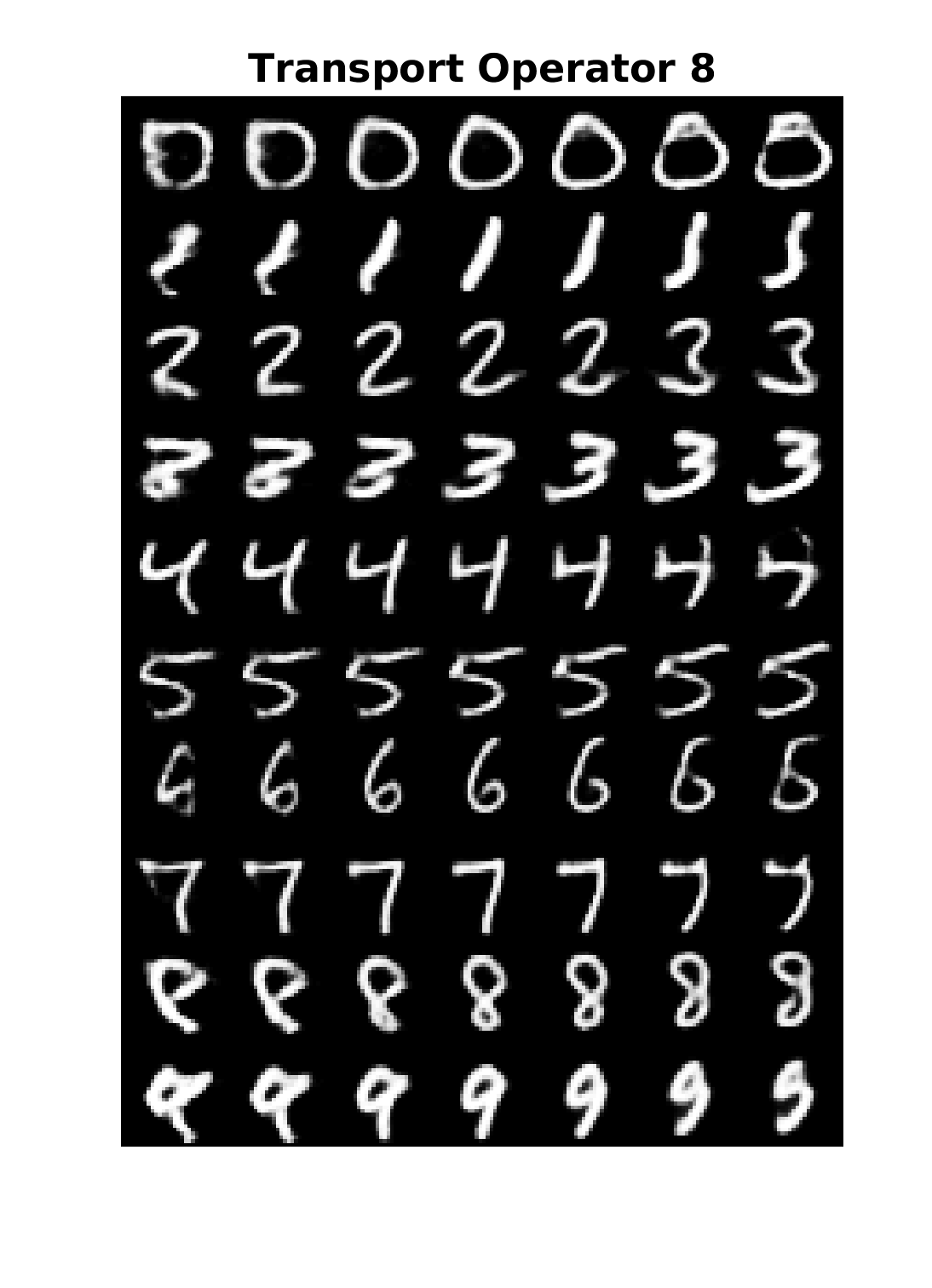}}
  \caption{}
\end{subfigure}
\begin{subfigure}[b]{0.24\textwidth}
 \centering
	{\includegraphics[width=0.98\textwidth]{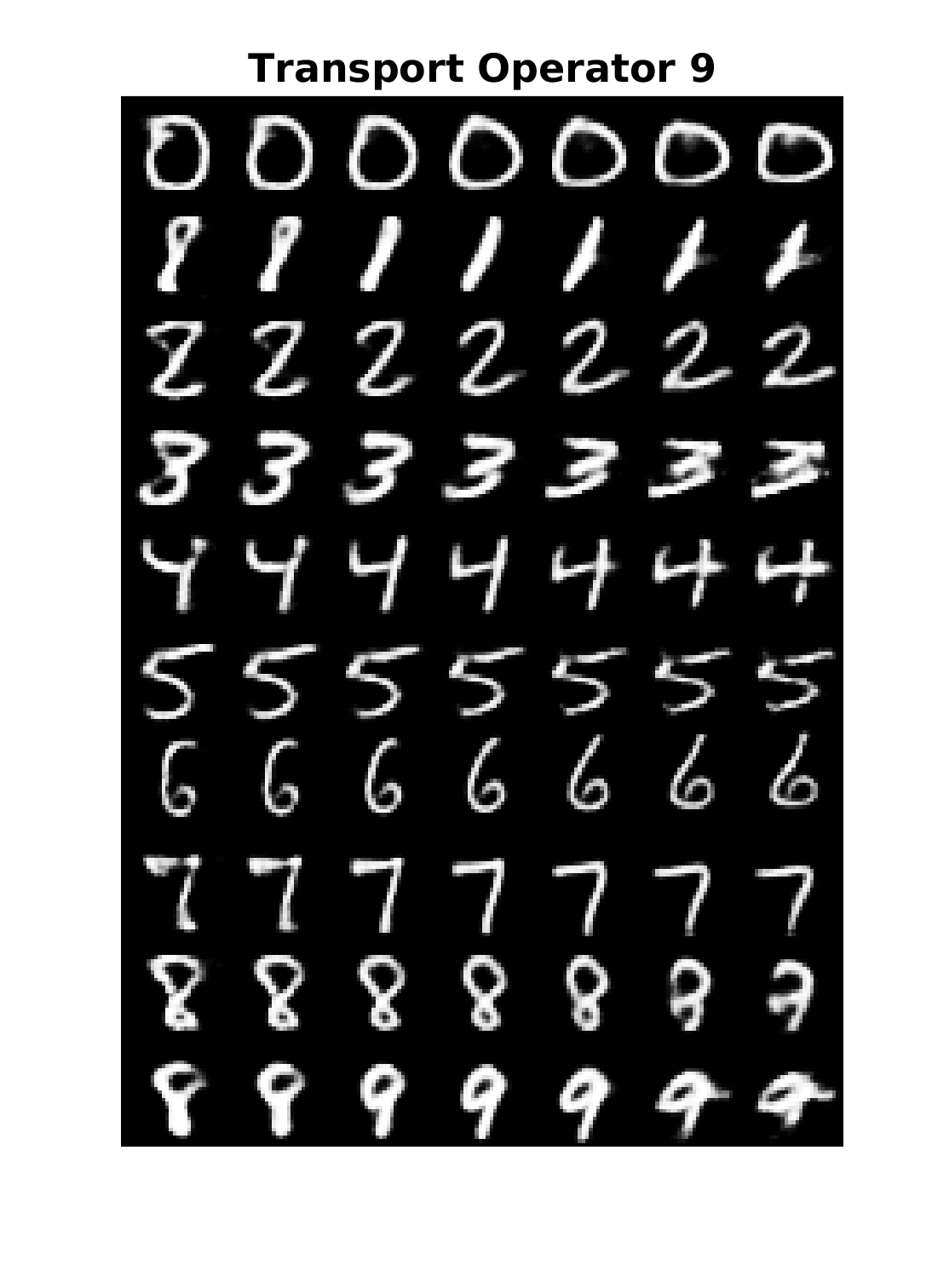}}
 \caption{}
\end{subfigure}
\begin{subfigure}[b]{0.24\textwidth}
  \centering
	{\includegraphics[width=0.98\textwidth]{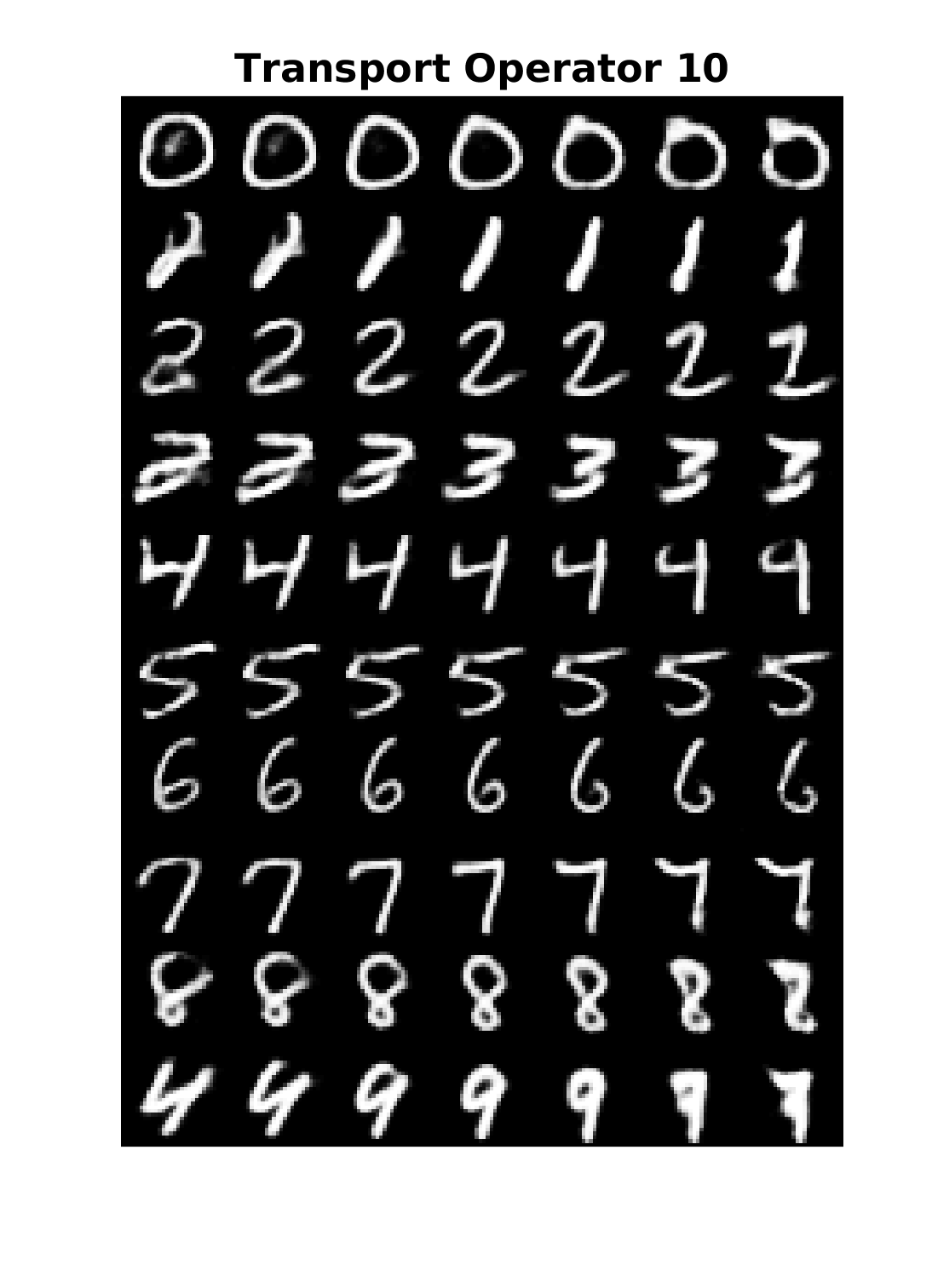}}
  \caption{}
\end{subfigure}
\begin{subfigure}[b]{0.24\textwidth}
 \centering
	{\includegraphics[width=0.98\textwidth]{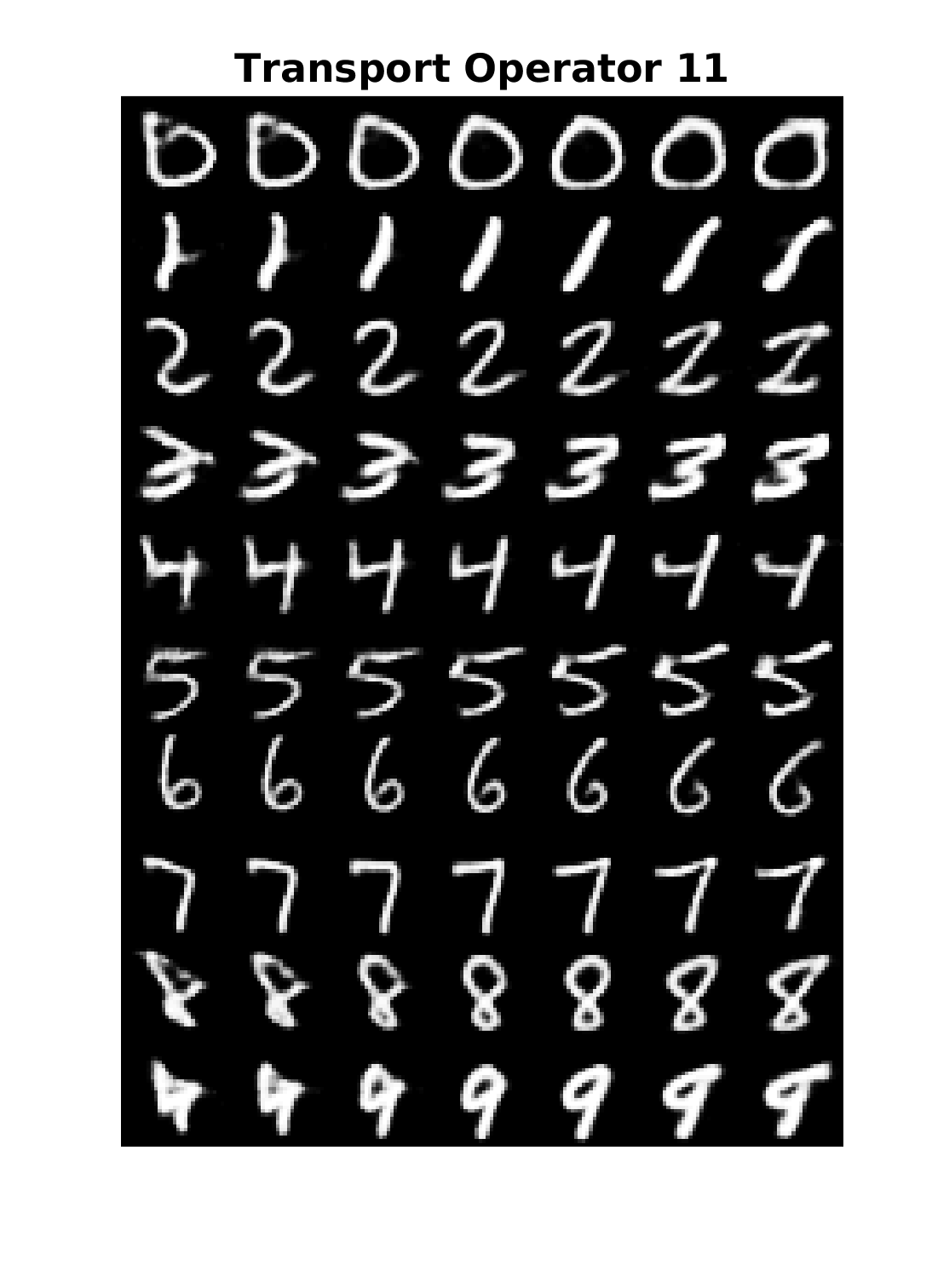}}
 \caption{}
\end{subfigure}

\begin{subfigure}[b]{0.24\textwidth}
  \centering
	{\includegraphics[width=0.98\textwidth]{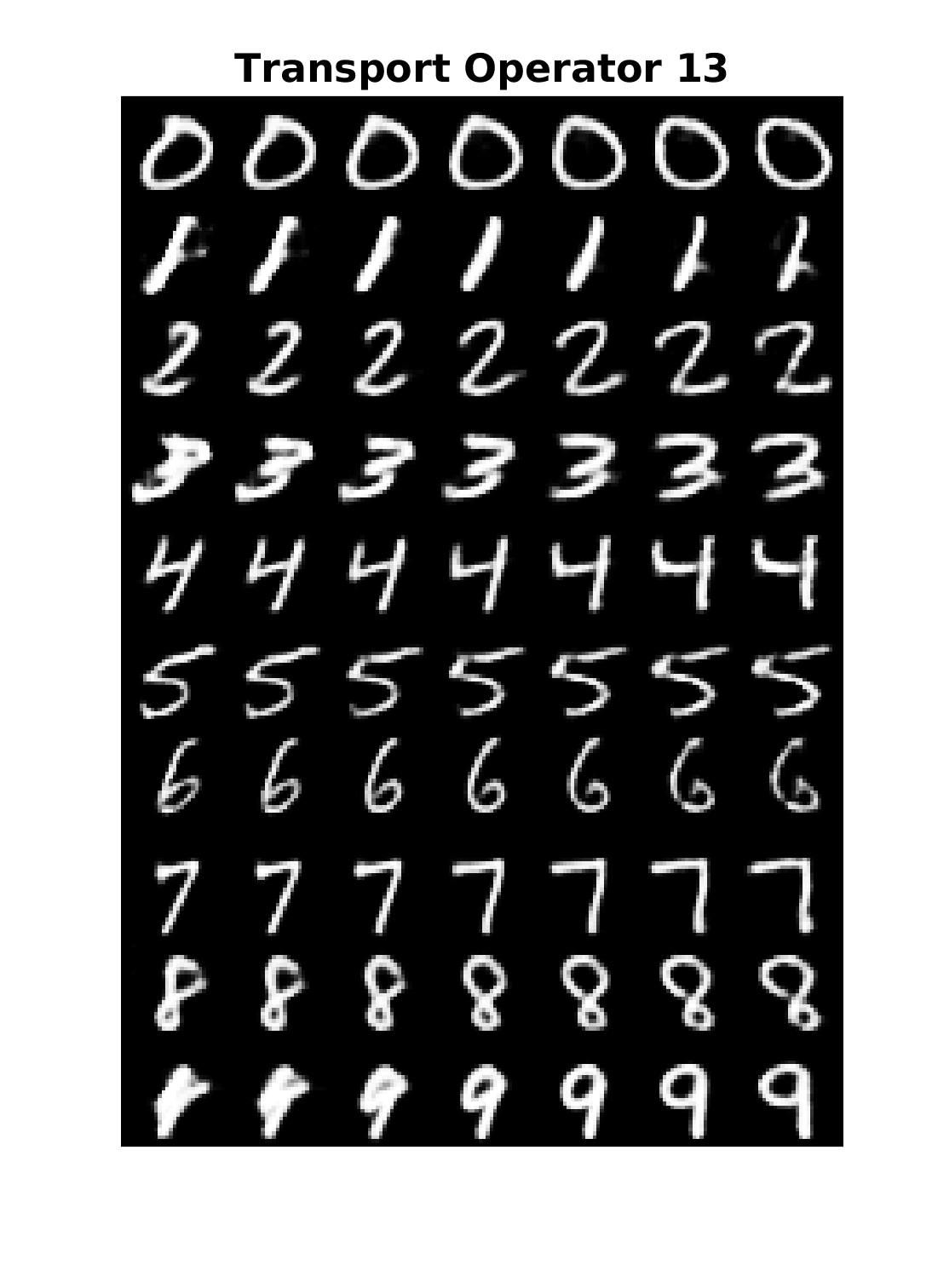}}
  \caption{}
\end{subfigure}
\begin{subfigure}[b]{0.24\textwidth}
 \centering
	{\includegraphics[width=0.98\textwidth]{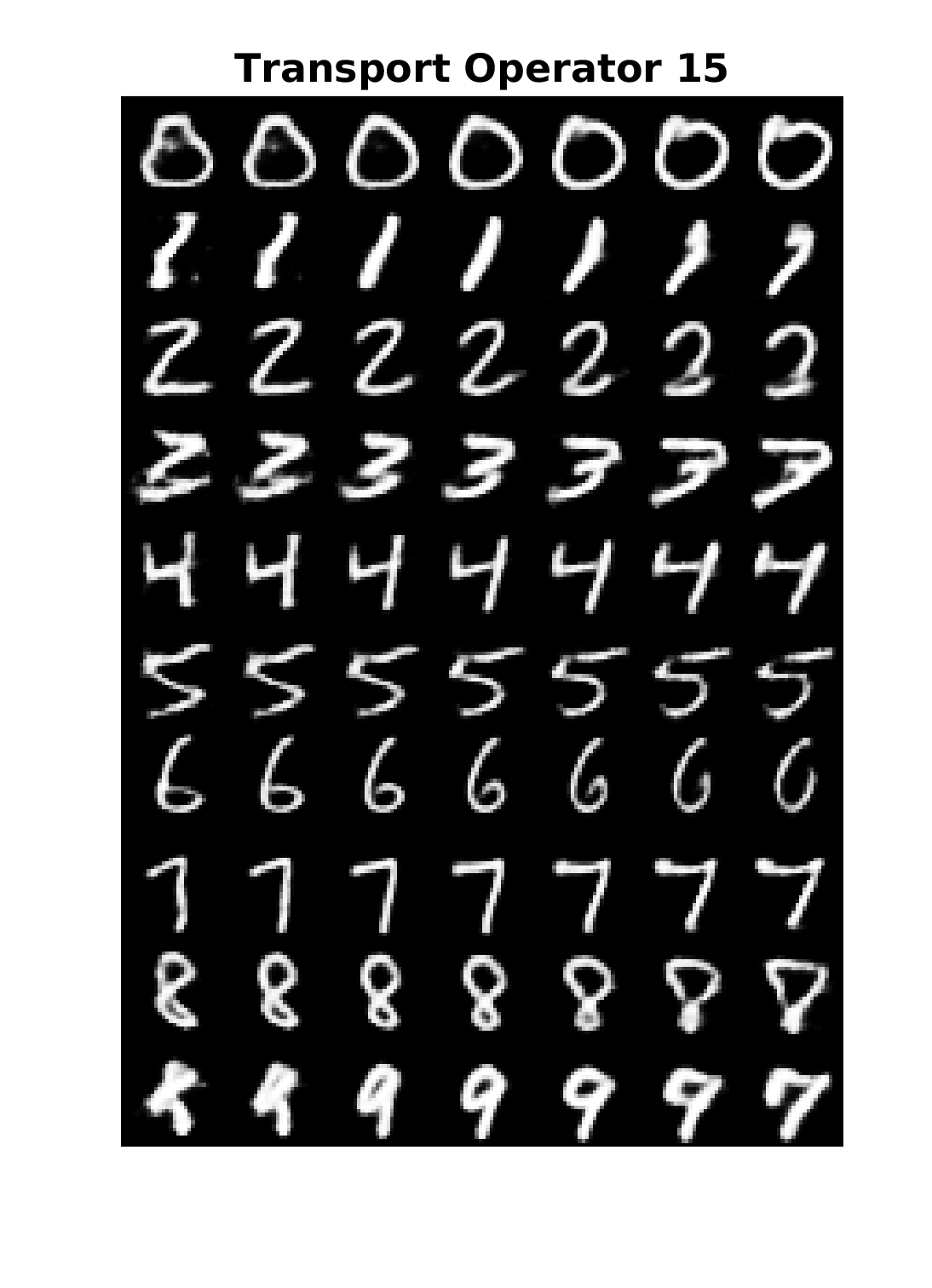}}
 \caption{}
\end{subfigure}

  \caption{\label{fig:mnistPathGen1} Paths generated by all non-zero transport operators trained on the MNIST dataset. Images in the middle column of the image block are the reconstructed inputs and images to the right and left are images decoded from transformed latent vectors in positive and negative directions, respectively}
	
\end{figure}

Here we show additional experimental details and results for the MNIST experiment. Fig.~\ref{fig:mag_mnist} shows the magnitude of all 16 operators after the fine-tuning phase. Six of the operators have their magnitudes reduced to zero. Fig.~\ref{fig:mnistPathGen1} shows the paths generated by transport operators trained on MNIST data. 

\section{Fashion MNIST Experiment Details}

The Fashion MNIST dataset is made available under the terms of the MIT license. We split the Fashion MNIST dataset into training, validation, and testing sets. The training set contains 50,000 images from the Fashion MNIST training set. The validation set is made up of the remaining 10,000 images. The traditional Fashion MNIST testing set is used for our testing set. The input images are normalized so their pixel values are between 0 and 1. The network architecture used for the autoencoder is the same as in the MNIST experiment and it is shown in Table~\ref{tab:MNISTNet}. The training parameters for the transport operator training phase and the fine-tuning phase are shown in Tables~\ref{tab:fmnist_TOparams} and \ref{tab:fMNIST_FTparams}.

Prior to training the coefficient encoder for the Fashion MNIST dataset, we train a classifier the latent vectors associated with labeled Fashion MNIST data which we use to encourage identity-preservation during coefficient encoder training. The training parameters for the coefficient encoder are shown in Table~\ref{tab:fMNIST_CEparams}. The latent vector classifier has a simple architecture of Linear(512), ReLU, Linear(10), Softmax.

\begin{table}[!htb]
\centering
\caption{Training parameters for the transport operator training phase of the Fashion MNIST experiment}
\label{tab:fmnist_TOparams}
\begin{tabular}{||l||} 
 \hline
 Fashion MNIST Transport Operator Training Parameters \\ 
 \hline
 batch size: 200   \\ 
 autoencoder training epochs: 300   \\
 transport operator training epochs: 50   \\
 latent space dimension ($z_{dim}$): 10 \\
 $M:$ 16 \\
 $lr_{\mathrm{net}}: 10^{-4}$  \\
$lr_{\Psi}: 10^{-3}$ \\
 $\zeta:$ 0.5  \\
 $\gamma:$ $2 \times 10^{-5}$   \\
  initialization variance for $\Psi$: 0.05 \\
   number of restarts for coefficient inference: 1 \\
   nearest neighbor count: 5\\
   latent scale: 30\\
 \hline
\end{tabular}

\end{table}

\begin{table}[!htb]
\parbox{.50\linewidth}{
\centering
\caption{Training parameters for the fine-tuning phase of the Fashion MNIST experiment}
\label{tab:fMNIST_FTparams}
\begin{tabular}{||l||} 
 \hline
 Fashion MNIST Fine-tuning Parameters \\ 
 \hline
 batch size: 200   \\ 
 transport operator training epochs: 150   \\
$lr_{\mathrm{net}}: 10^{-4}$  \\
$lr_{\Psi}: 10^{-3}$ \\
 $\zeta:$ 0.5  \\
 $\gamma:$ $2 \times 10^{-6}$   \\
 $\lambda$: 0.75 \\
  number of network update steps: 50  \\
  number of $\Psi$ update steps: 50 \\
 \hline
\end{tabular}
}
\parbox{.48\linewidth}{
\centering
\caption{Training parameters for the Fashion MNIST Coefficient Encoder}
\label{tab:fMNIST_CEparams}
\begin{tabular}{||l||} 
 \hline
 Fashion MNIST Coefficient Encoder Parameters \\ 
 \hline
 batch size: 200   \\ 
 training epochs: 300   \\
lr: $10^{-3}$  \\
 $\zeta_\text{prior}:$ 0.5  \\
 $\lambda_\text{kl}$: 0.5 \\
  coefficient spread scale: 0.1  \\
  classifier domain: latent \\
 \hline
\end{tabular}
}

\end{table}

\section{Fashion MNIST Experiment Additional Results}

\begin{figure}[!htb]

 \centering
	{\includegraphics[width=0.4\textwidth]{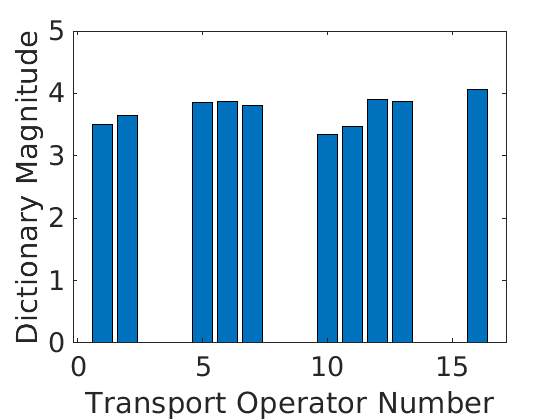}}

  \caption{\label{fig:mag_fmnist} The resulting magnitude of the learned operators after fine-tuning the MAE.}
	
\end{figure}

Here we show additional experimental details and results for the Fashion MNIST experiment.  Fig.~\ref{fig:mag_fmnist} shows the magnitude of all 16 operators after the fine-tuning phase. Six of the operators had their magnitudes reduced to zero.

\begin{figure}[!htb]

\centering
\begin{subfigure}[b]{0.24\textwidth}
  \centering
	{\includegraphics[width=0.98\textwidth]{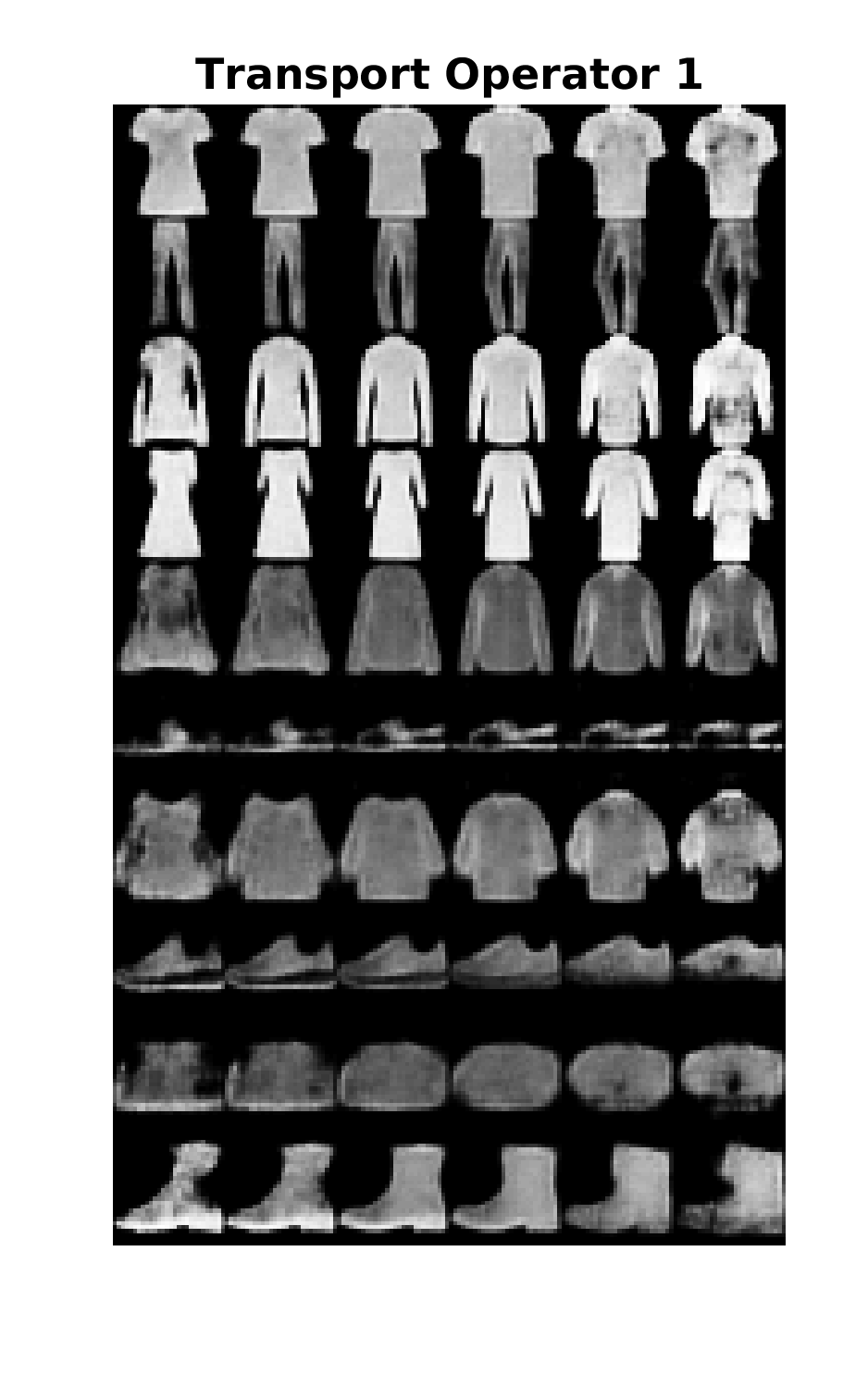}}
  \caption{}
\end{subfigure}
\begin{subfigure}[b]{0.24\textwidth}
 \centering
	{\includegraphics[width=0.98\textwidth]{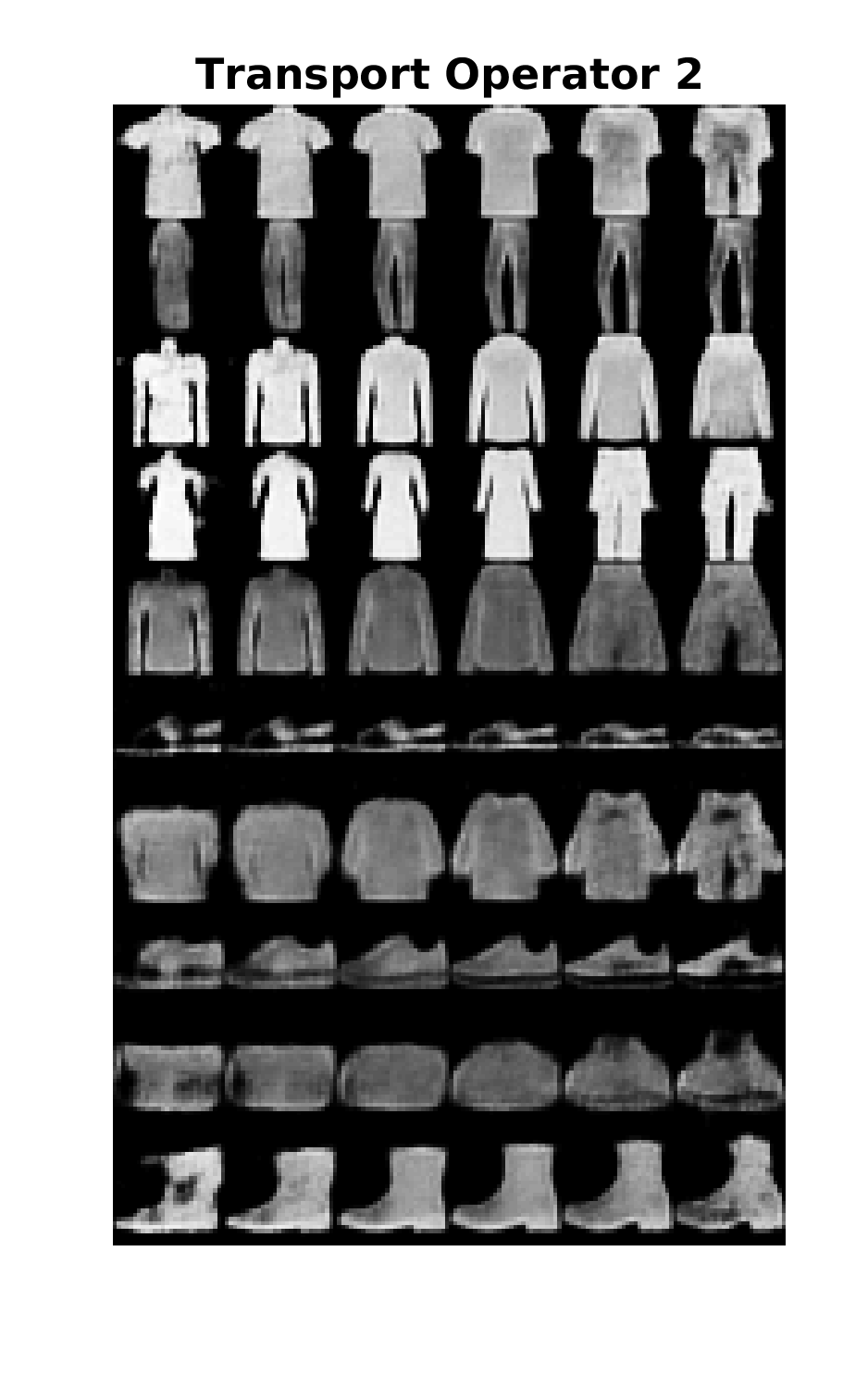}}
 \caption{}
\end{subfigure}
\begin{subfigure}[b]{0.24\textwidth}
  \centering
	{\includegraphics[width=0.98\textwidth]{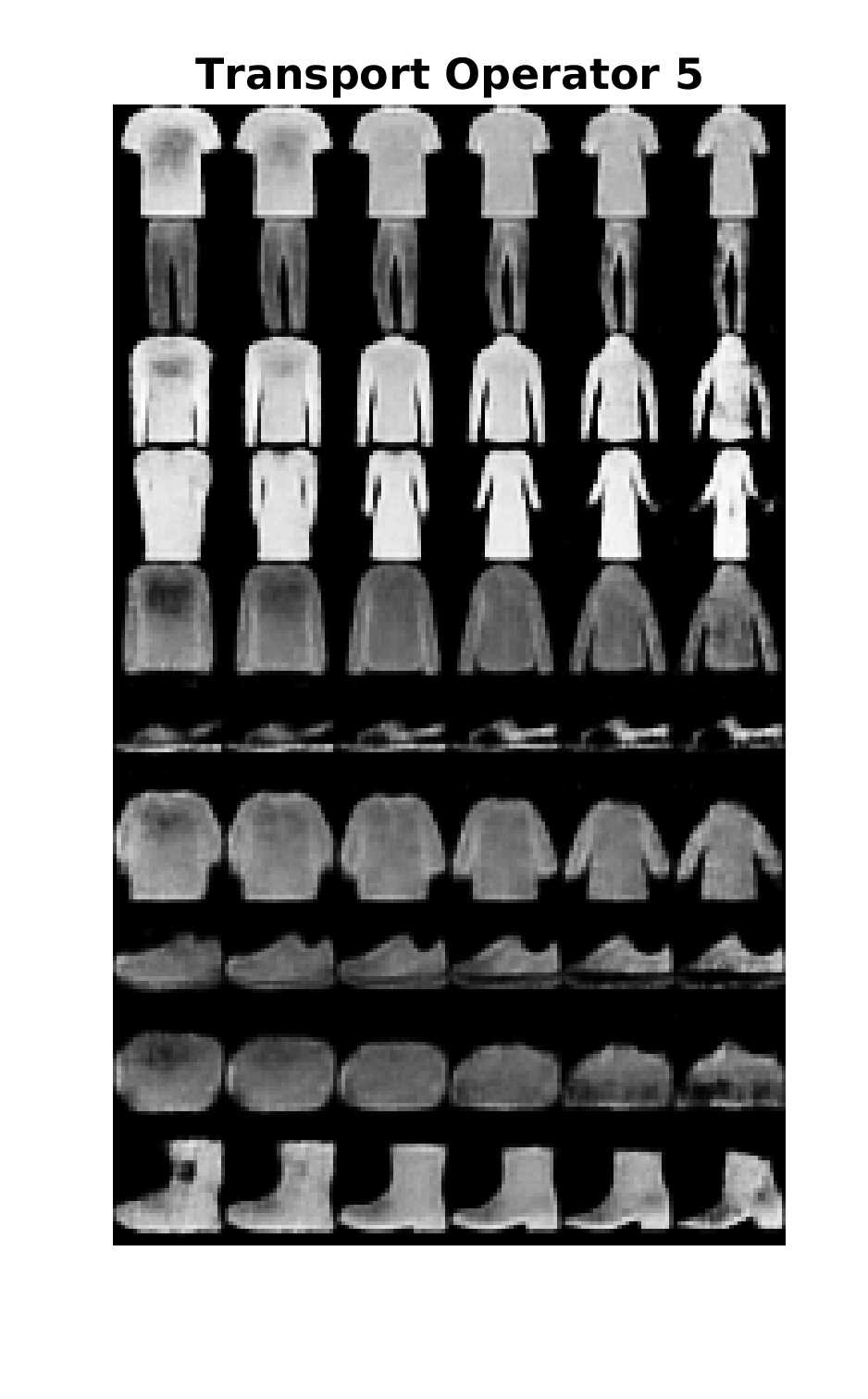}}
  \caption{}
\end{subfigure}
\begin{subfigure}[b]{0.24\textwidth}
 \centering
	{\includegraphics[width=0.98\textwidth]{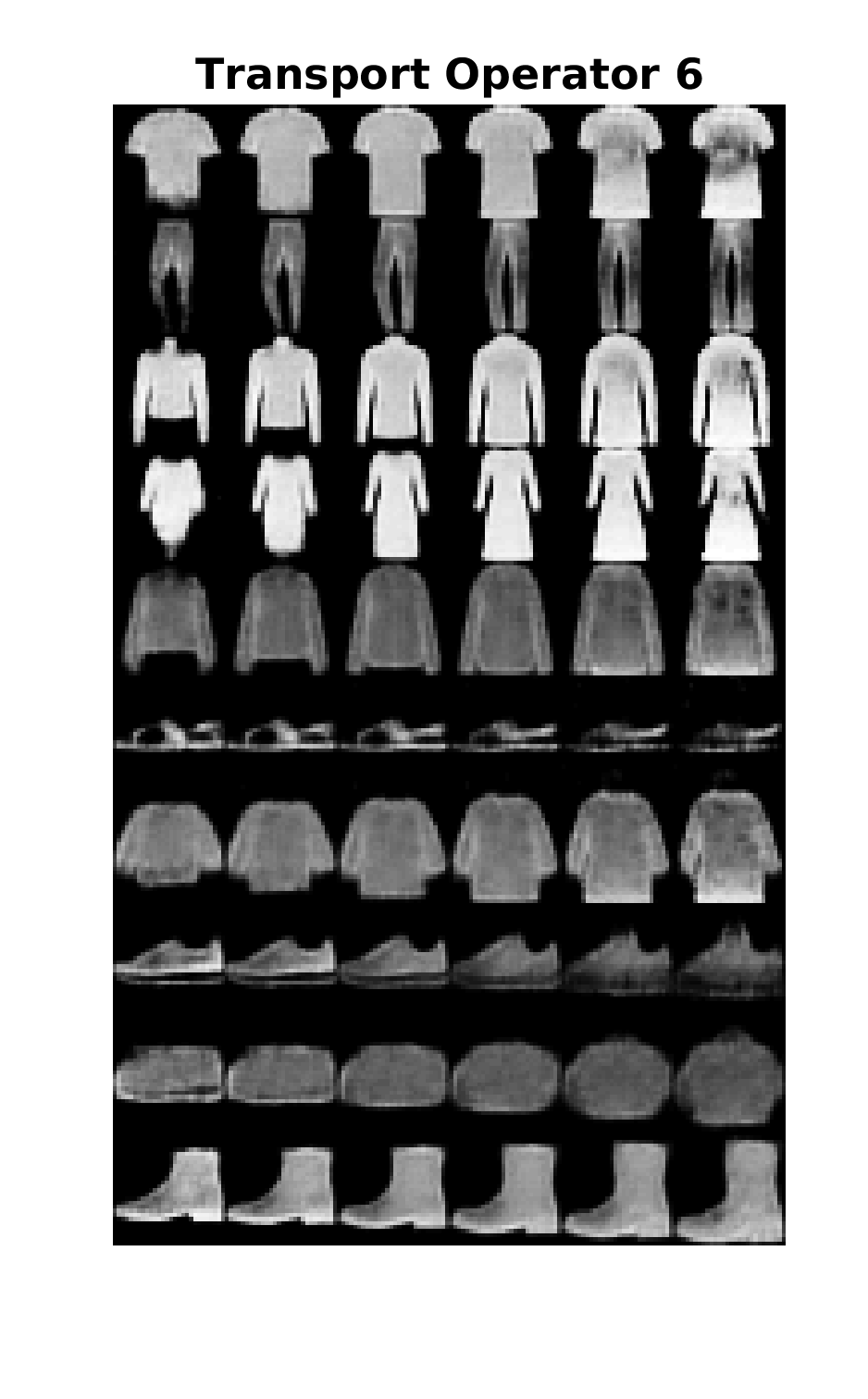}}
 \caption{}
\end{subfigure}

\begin{subfigure}[b]{0.24\textwidth}
  \centering
	{\includegraphics[width=0.98\textwidth]{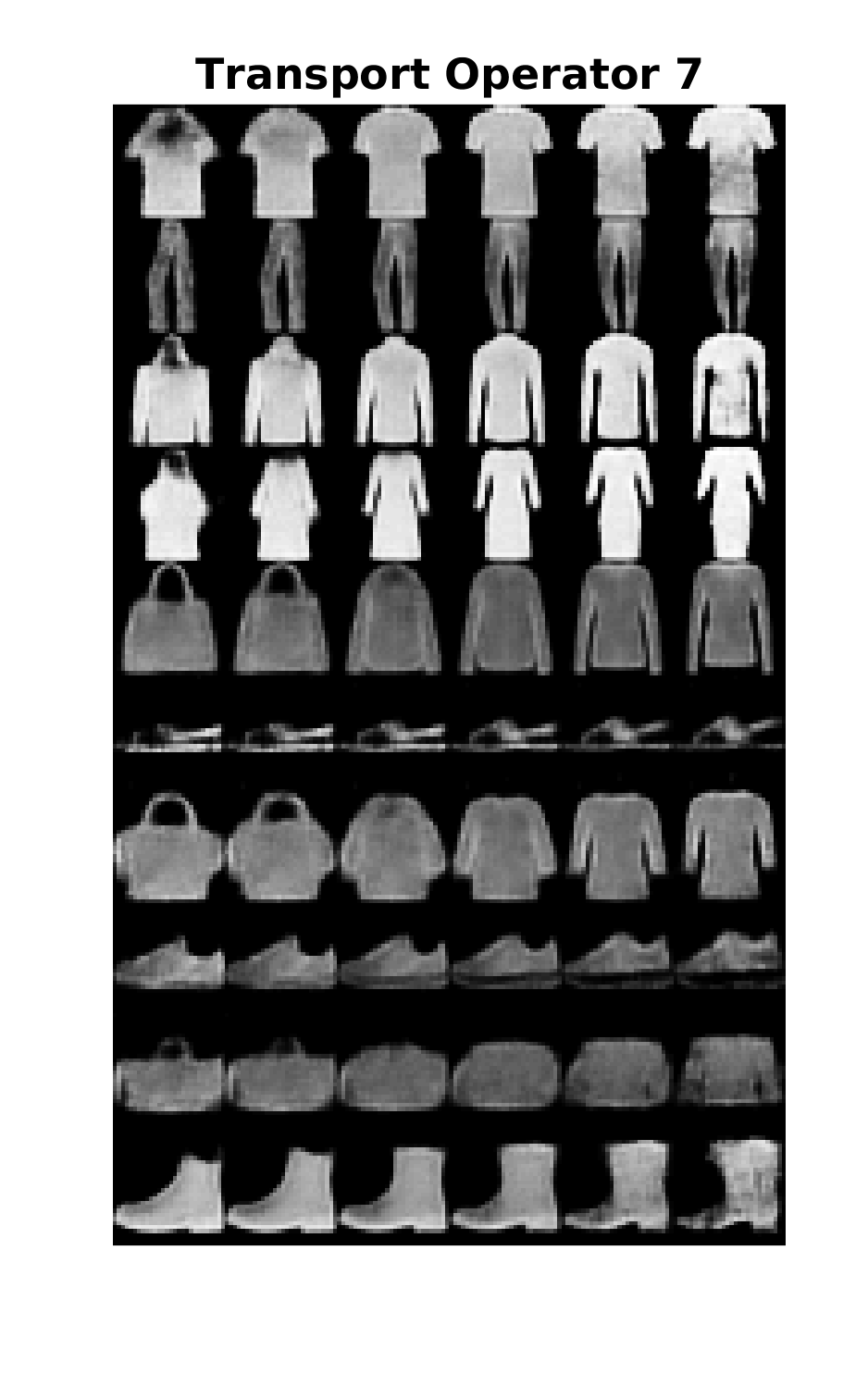}}
  \caption{}
\end{subfigure}
\begin{subfigure}[b]{0.24\textwidth}
 \centering
	{\includegraphics[width=0.98\textwidth]{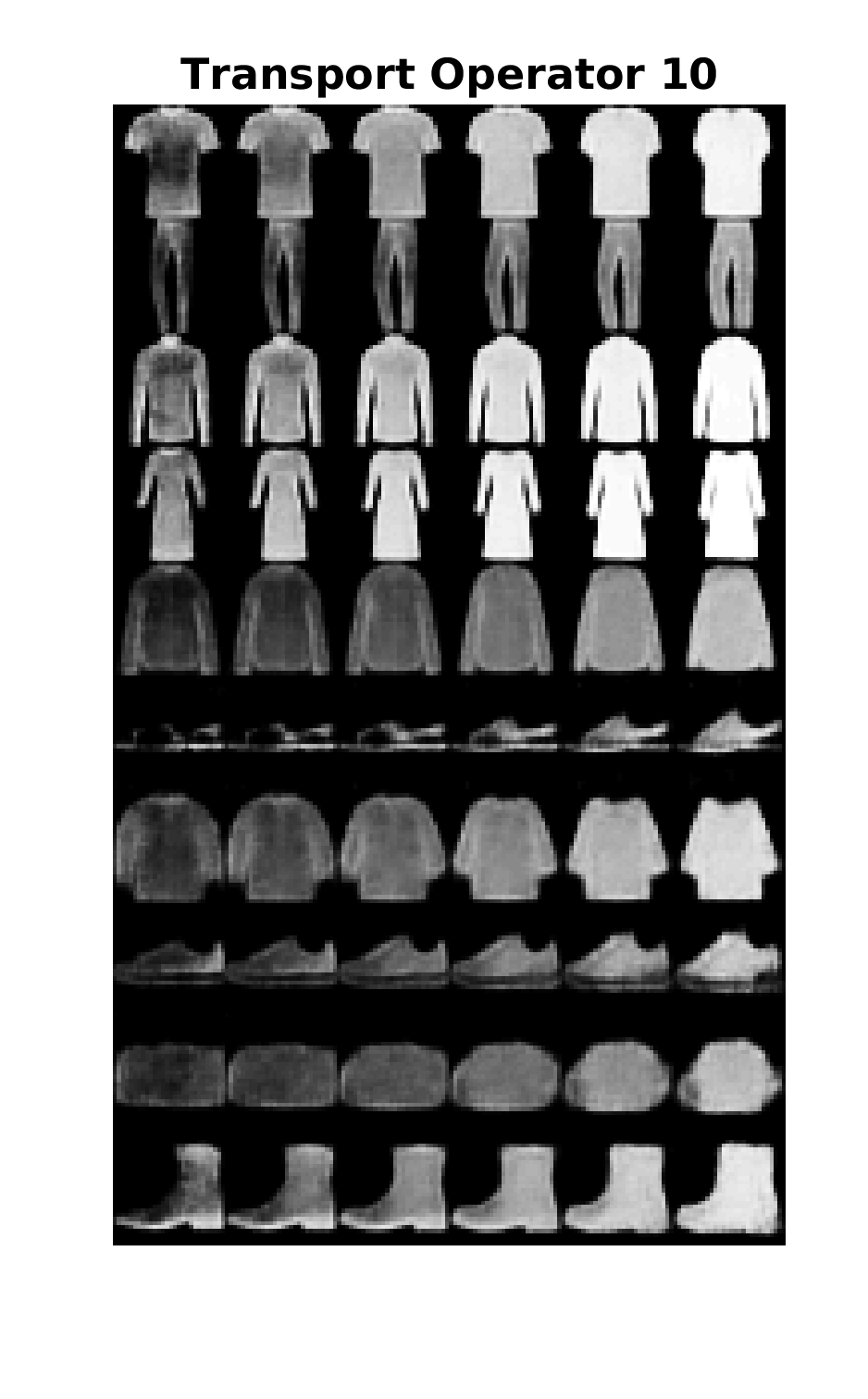}}
 \caption{}
\end{subfigure}
\begin{subfigure}[b]{0.24\textwidth}
  \centering
	{\includegraphics[width=0.98\textwidth]{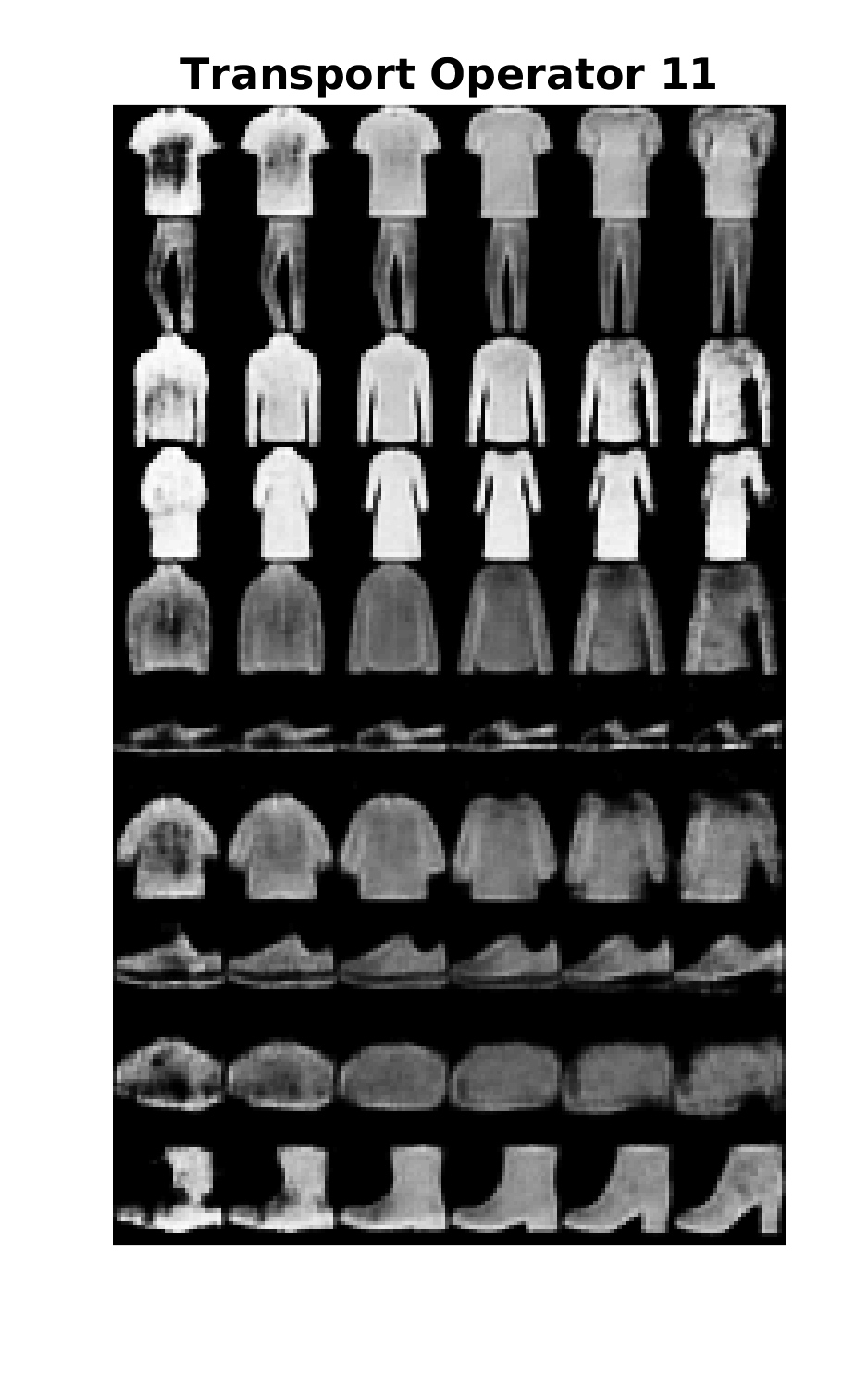}}
  \caption{}
\end{subfigure}
\begin{subfigure}[b]{0.24\textwidth}
 \centering
	{\includegraphics[width=0.98\textwidth]{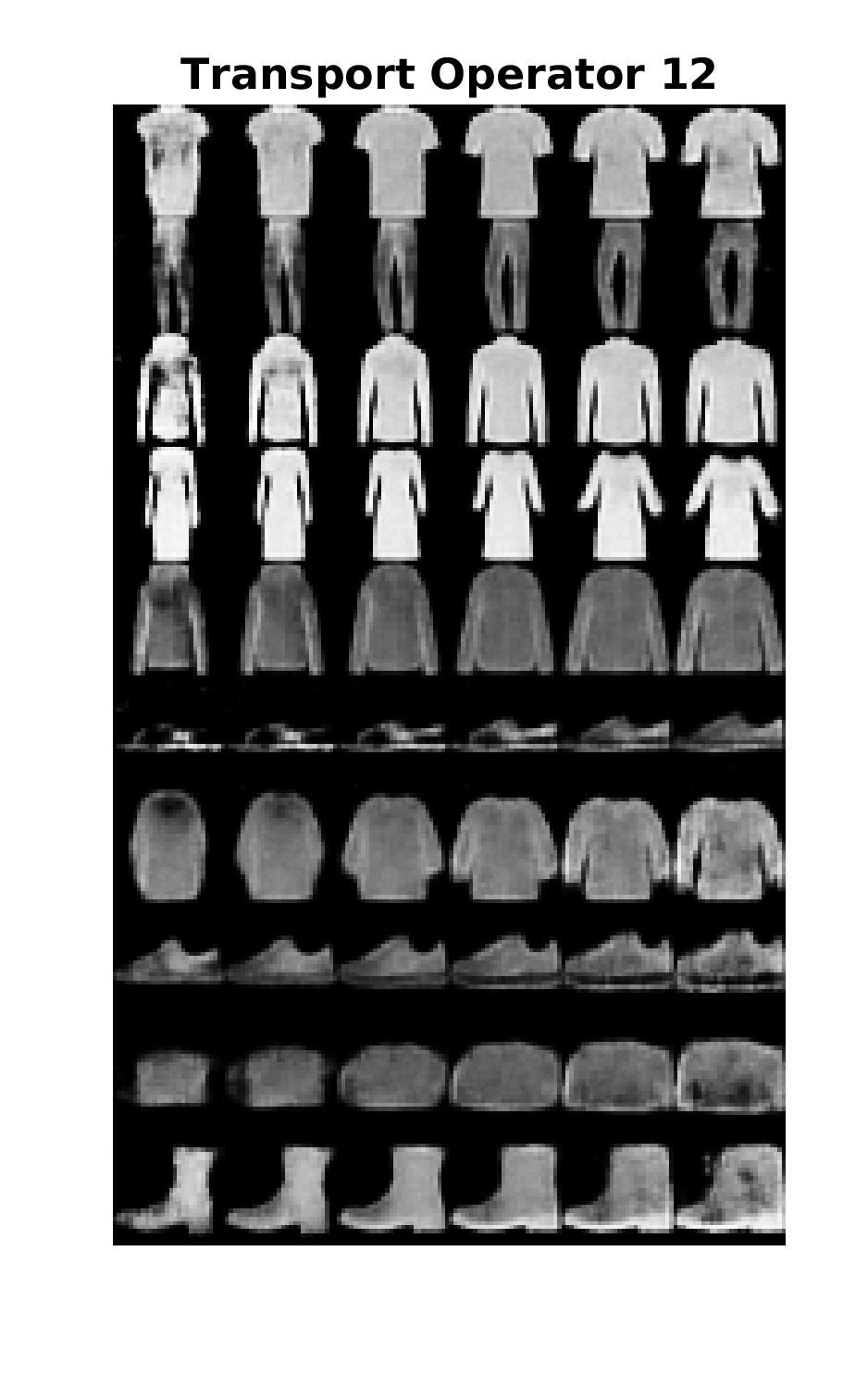}}
 \caption{}
\end{subfigure}

\begin{subfigure}[b]{0.24\textwidth}
  \centering
	{\includegraphics[width=0.98\textwidth]{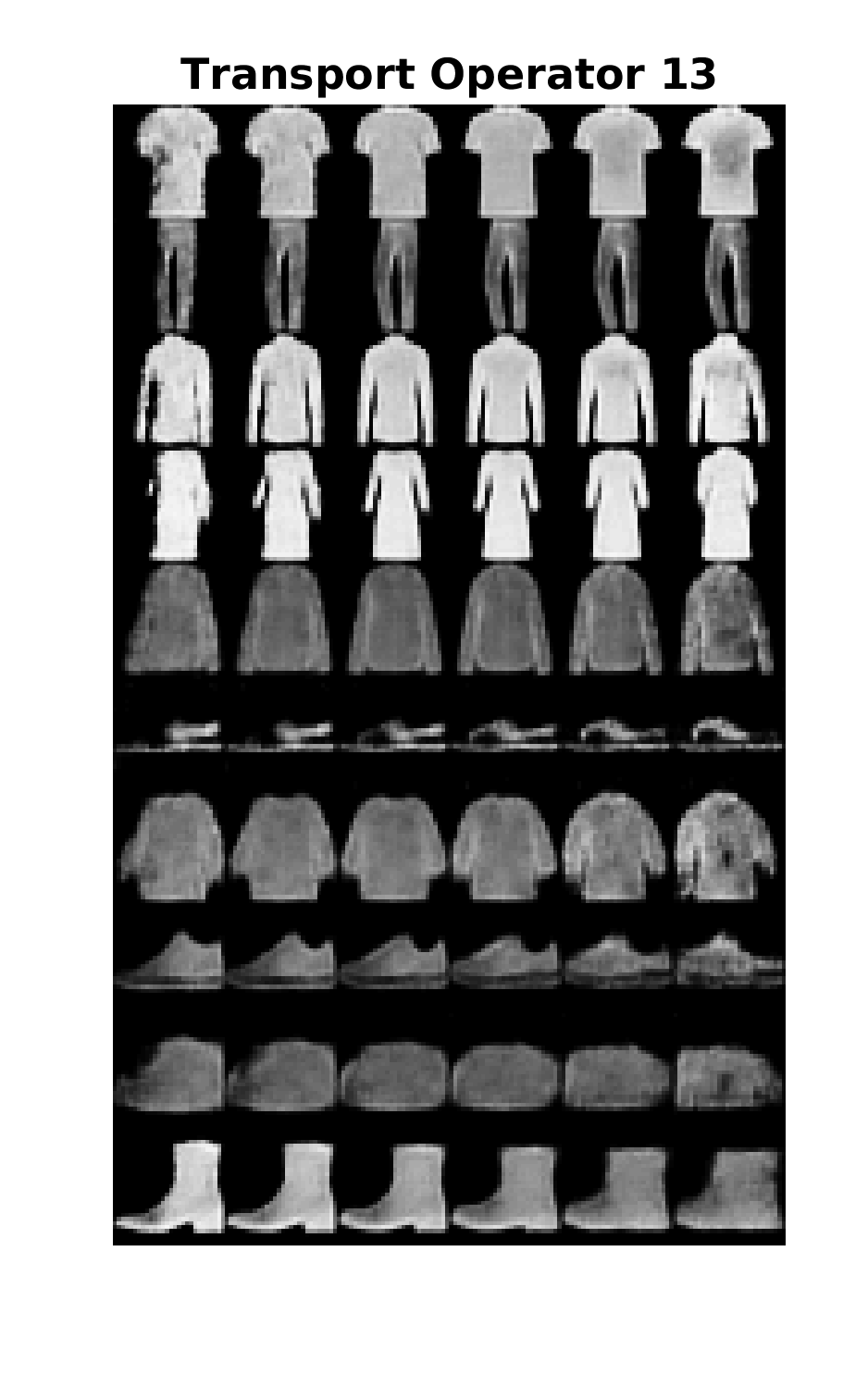}}
  \caption{}
\end{subfigure}
\begin{subfigure}[b]{0.24\textwidth}
 \centering
	{\includegraphics[width=0.98\textwidth]{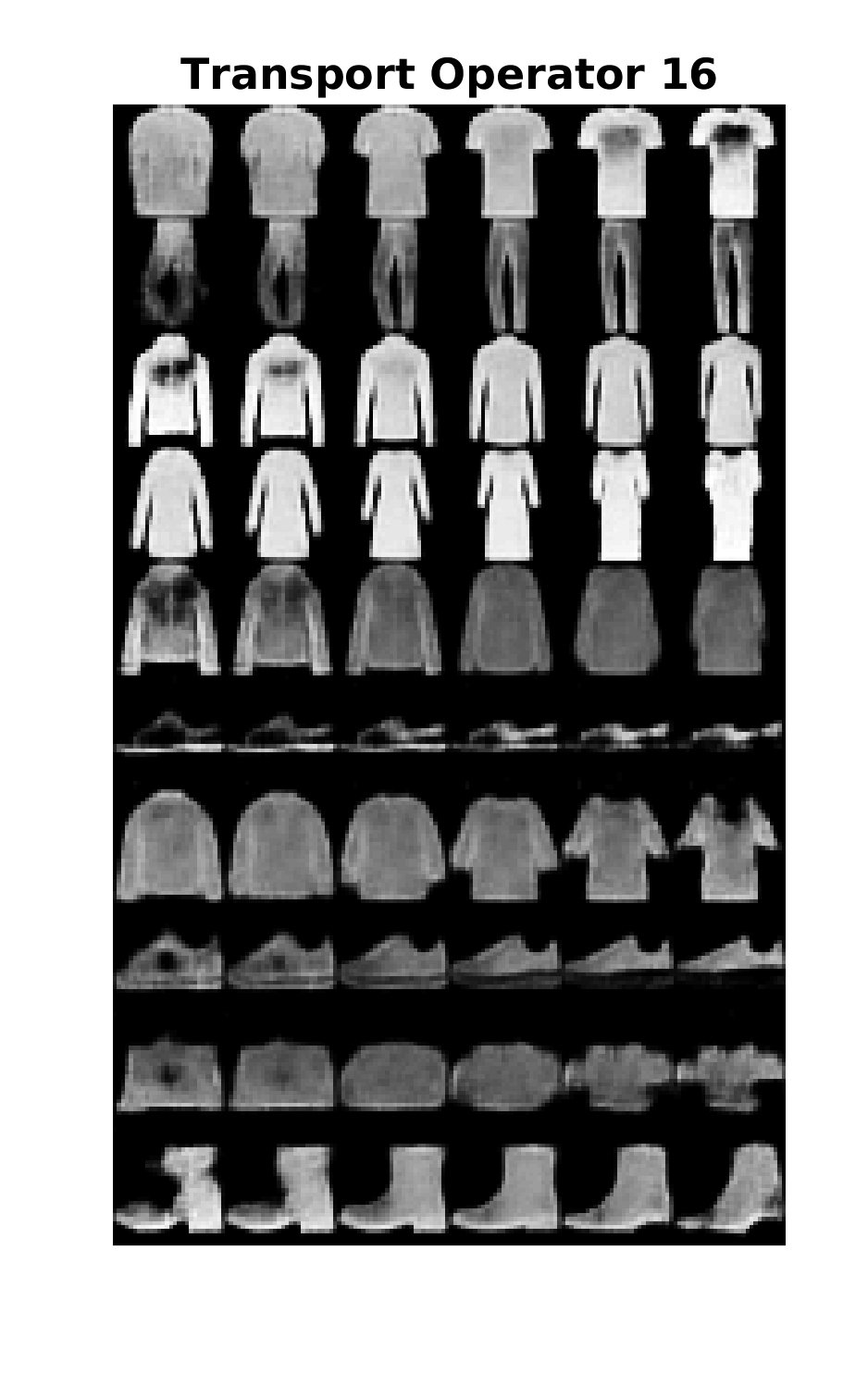}}
 \caption{}
\end{subfigure}

  \caption{\label{fig:fmnistPathGen} Paths generated by all non-zero transport operators trained on the Fashion MNIST dataset. Images in the middle column of the image block are the reconstructed inputs and images to the right and left are images decoded from transformed latent vectors in positive and negative directions, respectively}
	
\end{figure}

To visualize how the use of the transport operators varies over the latent space, we generate an Isomap embedding~\citep{tenenbaum2000global} of latent vectors and color each point by the encoded scale parameter for coefficients associated with each of the transport operators. Fig.~\ref{subfig:embed_fmnist} shows these embeddings for Fashion MNIST data. Each operator has regions of the latent space where their use is concentrated. Fig.~\ref{subfig:spread_fmnist} shows the average coefficient scale weights for each class (rows) and each transport operator (columns) for Fashion MNIST. There are some classes like trouser and sandal (classes 1 and 5) which have large encoded coefficient scale weights for most of the transport operators. This means they are robust to many natural transformations. Other classes like coat and shirt (classes 4 and 6) have smaller encoded coefficient scale weights which means they are sensitive to many transformations. The images to the right in Fig.~\ref{subfig:spread_fmnist} show transport operators being applied to samples with high encoded scale weights (in a yellow box) and samples with low encoded scale weights (in a blue box).  Fig.~\ref{fig:fmnistPathGen} shows the paths generated by non-zero transport operators trained on Fashion MNIST data.

\section{CelebA Experiment Details}

The CelebA dataset is publicly available for non-commercial research purposes. We split the CelebA dataset into training and testing sets. The training set contains the first 150,000 images accompanied with the entire test set. The input images are normalized so their pixel values are between 0 and 1. The network architecture used for the autoencoder is shown in Table~\ref{tab:celebANet}. The training parameters for the transport operator training phase and the fine-tuning phase are shown in Table~\ref{tab:celebA_TOparams}.

The attribute classifier is a ResNet-18 model adapted from \texttt{\href{https://github.com/d-li14/face-attribute-prediction}{https://github.com/d-li14/face-attribute-prediction}} with 16 classifier heads after the layer with 512 hidden units. Each classifier head corresponds to a single attribute and is modeled as a fully-connected linear layer with 256 hidden units, followed by batch normalization, dropout, and ReLU layers. Afterwards, two logits are output for a 0/1 prediction for each classifier output. The training procedure, including the dynamic loss weighting, follows \citep{mao2020deep}.

\begin{table}[!htb]
\centering
\caption{Network Architecture for CelebA Experiments}
\label{tab:celebANet}
\begin{tabular}{||l l||} 
 \hline
 Encoder Network & Decoder Network  \\ 
 \hline
 Input $\in \mathbb{R}^{64 \times 64}$ & Input $\in \mathbb{R}^{32}$  \\ 
 conv: chan: 32 , kern: 4, stride: 2, pad: 1  & Linear: 80,000 Units  \\
 BatchNorm: feat: 32 & ReLU\\ 
 ReLU &  convTranpose: chan: 256, kern: 3, stride: 1, pad: 0\\
 conv: chan: 64, kern: 4, stride: 2, pad: 1  & BatchNorm: feat: 256 \\ 
 BatchNorm: feat: 64 & ReLU\\
 ReLU &  convTranpose: chann: 256, kern: 3, stride: 1, pad: 0 \\
 conv: chan: 128, kern: 3, stride: 2, pad: 1 &  BatchNorm: feat: 256\\ 
  BatchNorm: feat: 128 & ReLU \\
 ReLU & convTranpose: chan: 256, kernel: 3, stride: 1, pad: 1\\
  conv: chan: 256, kern: 3, stride: 1, pad: 1 &  BatchNorm: feat: 256\\ 
  BatchNorm: feat: 128 & ReLU \\
 ReLU & convTranpose: chan: 128, kernel: 3, stride: 1, pad: 1\\
  conv: chan: 256, kern: 4, stride: 2, pad: 1 &  BatchNorm: feat: 128\\ 
  BatchNorm: feat: 256 & ReLU \\
 ReLU & convTranpose: chan: 128, kernel: 3, stride: 1, pad: 0\\
  conv: chan: 128, kern: 4, stride: 2, pad: 1 &  BatchNorm: feat: 128\\ 
  BatchNorm: feat: 128 & ReLU \\
 ReLU & convTranpose: chan: 3, kernel: 4, stride: 2, pad: 0\\
 Linear: 32 Units & Sigmoid \\ 

 \hline
\end{tabular}
\end{table}

\begin{table}[!htb]
\caption{Training parameters for the CelebA experiment}
\parbox{0.6\linewidth}{
\centering
\label{tab:celebA_TOparams}
\begin{tabular}{||l||} 
 \hline
 CelebA Transport Operator Training Parameters \\ 
 \hline
 batch size: 500   \\ 
 autoencoder training epochs: 300   \\
 transport operator training epochs: 50   \\
 latent space dimension ($z_{dim}$): 32 \\
 $M:$ 40 \\
 $lr_{\mathrm{net}}: 10^{-4}$  \\
$lr_{\Psi}: 10^{-3}$ \\
 $\zeta:$ 1.5  \\
 $\gamma:$ $1 \times 10^{-5}$   \\
  initialization variance for $\Psi$: 0.05 \\
   number of restarts for coefficient inference: 1 \\
   nearest neighbor count: 5\\
   latent scale: 2\\
 \hline
\end{tabular}
}
\parbox{0.35\linewidth}{
\label{tab:celebA_FTparams}
\begin{tabular}{||l||} 
 \hline
 CelebA Fine-tuning Parameters \\ 
 \hline
 batch size: 500   \\ 
 fine-tuning epochs: 10   \\
$lr_{\mathrm{net}}: 10^{-4}$  \\
$lr_{\Psi}: 10^{-3}$ \\
 $\zeta:$ 0.8  \\
 $\gamma:$ $5 \times 10^{-7}$   \\
 $\lambda$: 0.75 \\
  number of network update steps: 50  \\
  number of $\Psi$ update steps: 50 \\
 \hline
\end{tabular}
}
\end{table}

\section{CelebA Experiment Additional Results}
\label{app:celeba_extra}

\begin{figure}[!htb]

\centering
\begin{subfigure}[b]{0.48\textwidth}
  \centering
	{\includegraphics[width=0.98\textwidth]{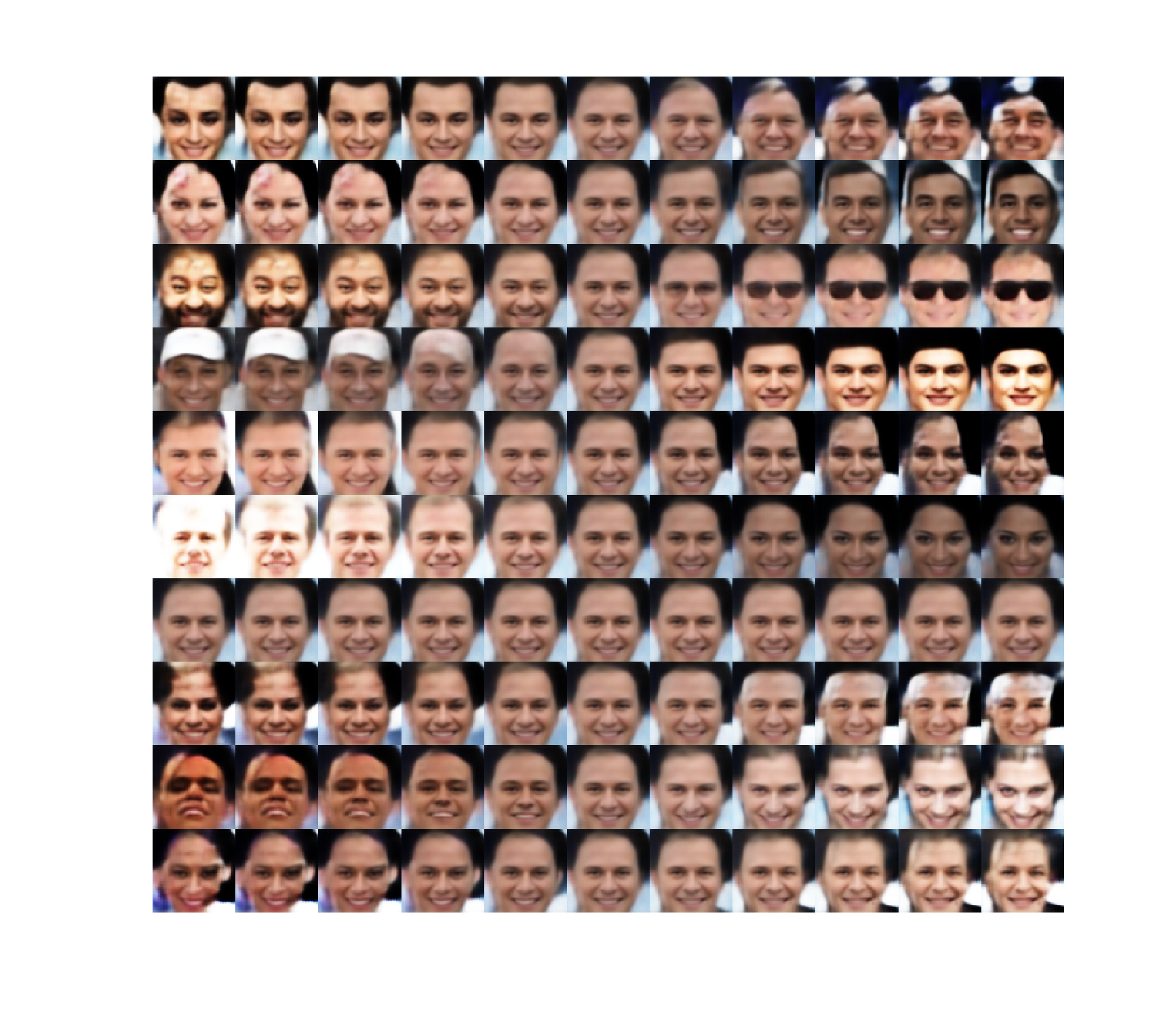}}
  \caption{}
\end{subfigure}
\begin{subfigure}[b]{0.48\textwidth}
 \centering
	{\includegraphics[width=0.98\textwidth]{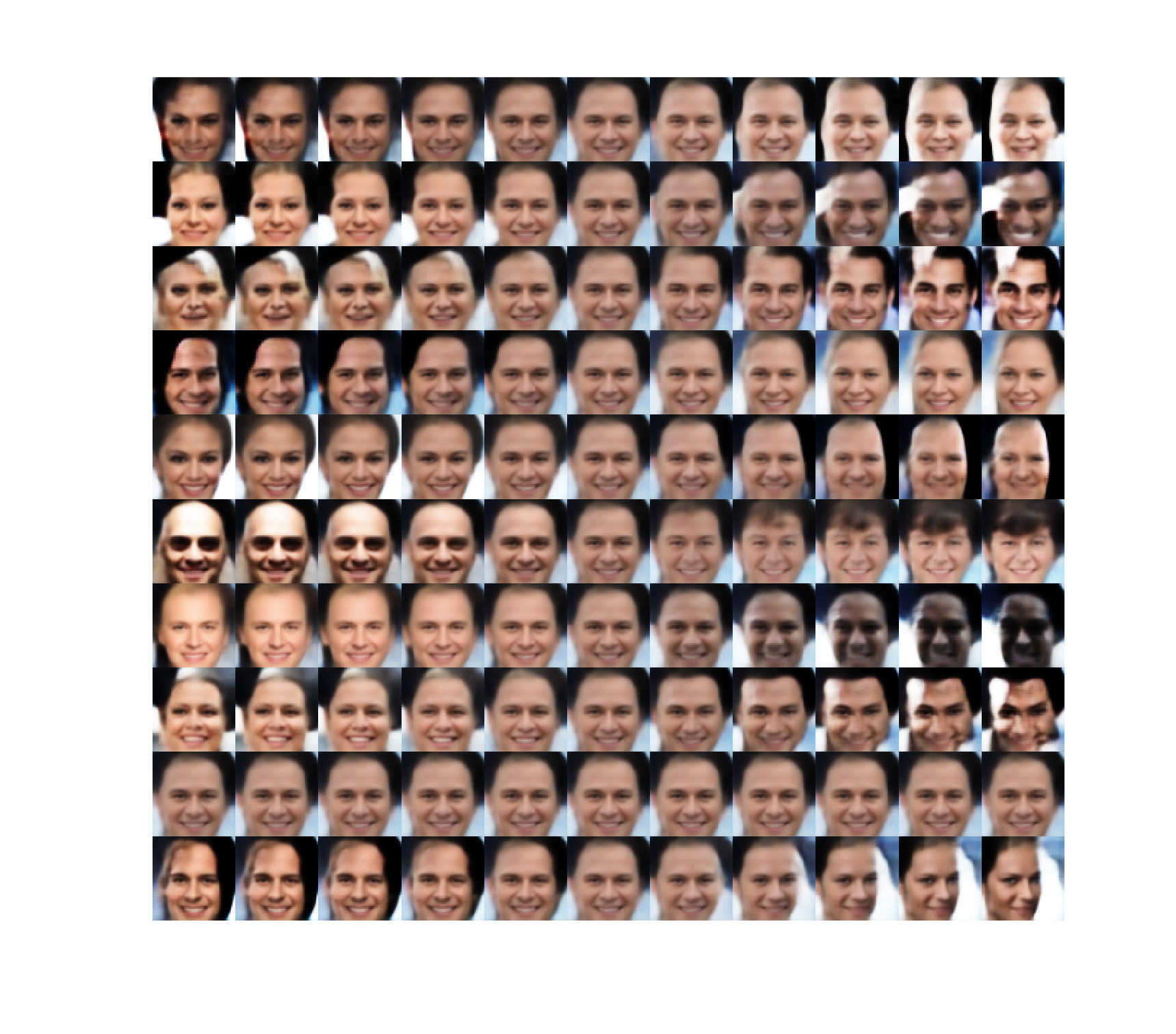}}
 \caption{}
\end{subfigure}

\begin{subfigure}[b]{0.48\textwidth}
  \centering
	{\includegraphics[width=0.98\textwidth]{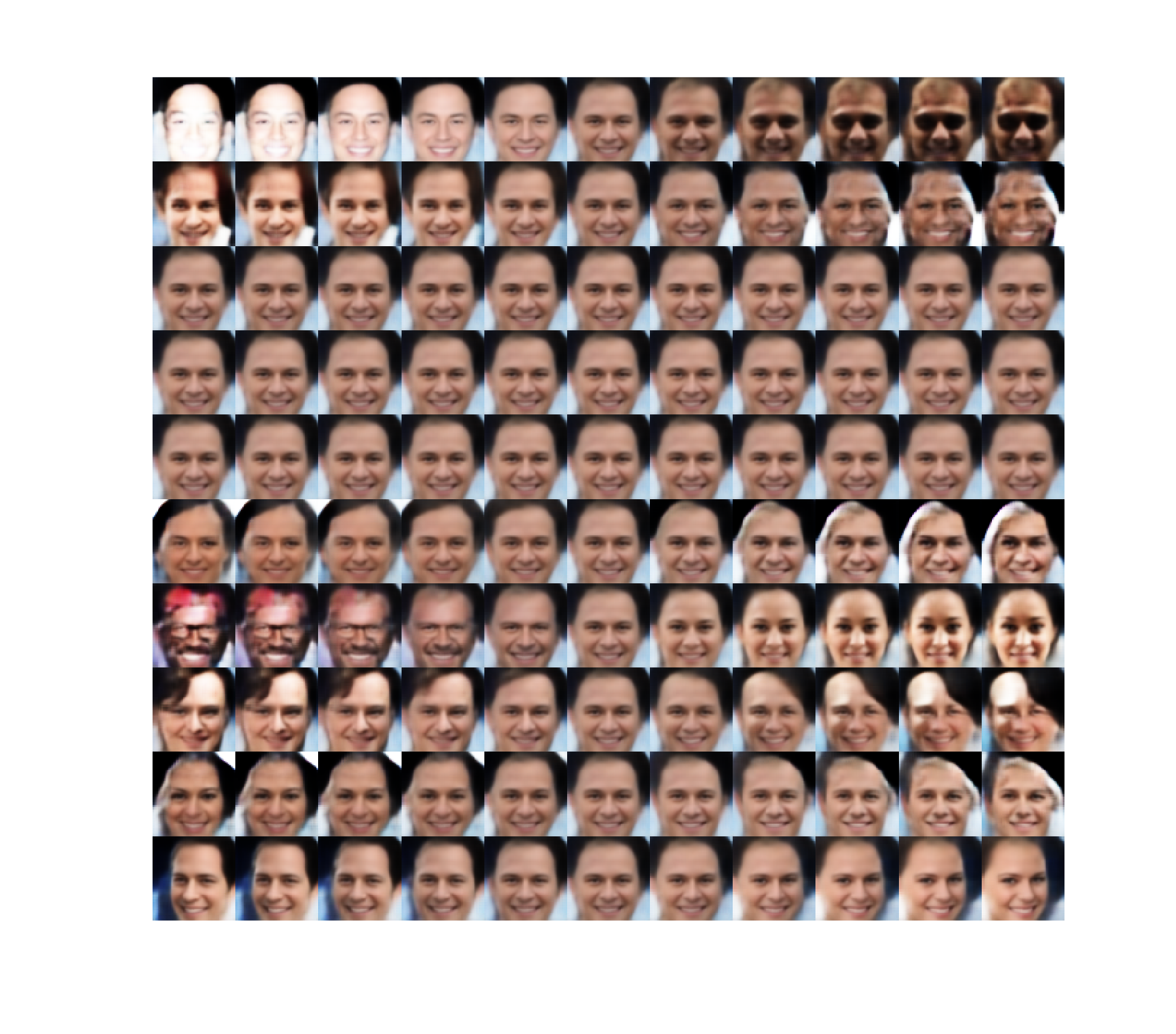}}
  \caption{}
\end{subfigure}
\begin{subfigure}[b]{0.48\textwidth}
 \centering
	{\includegraphics[width=0.98\textwidth]{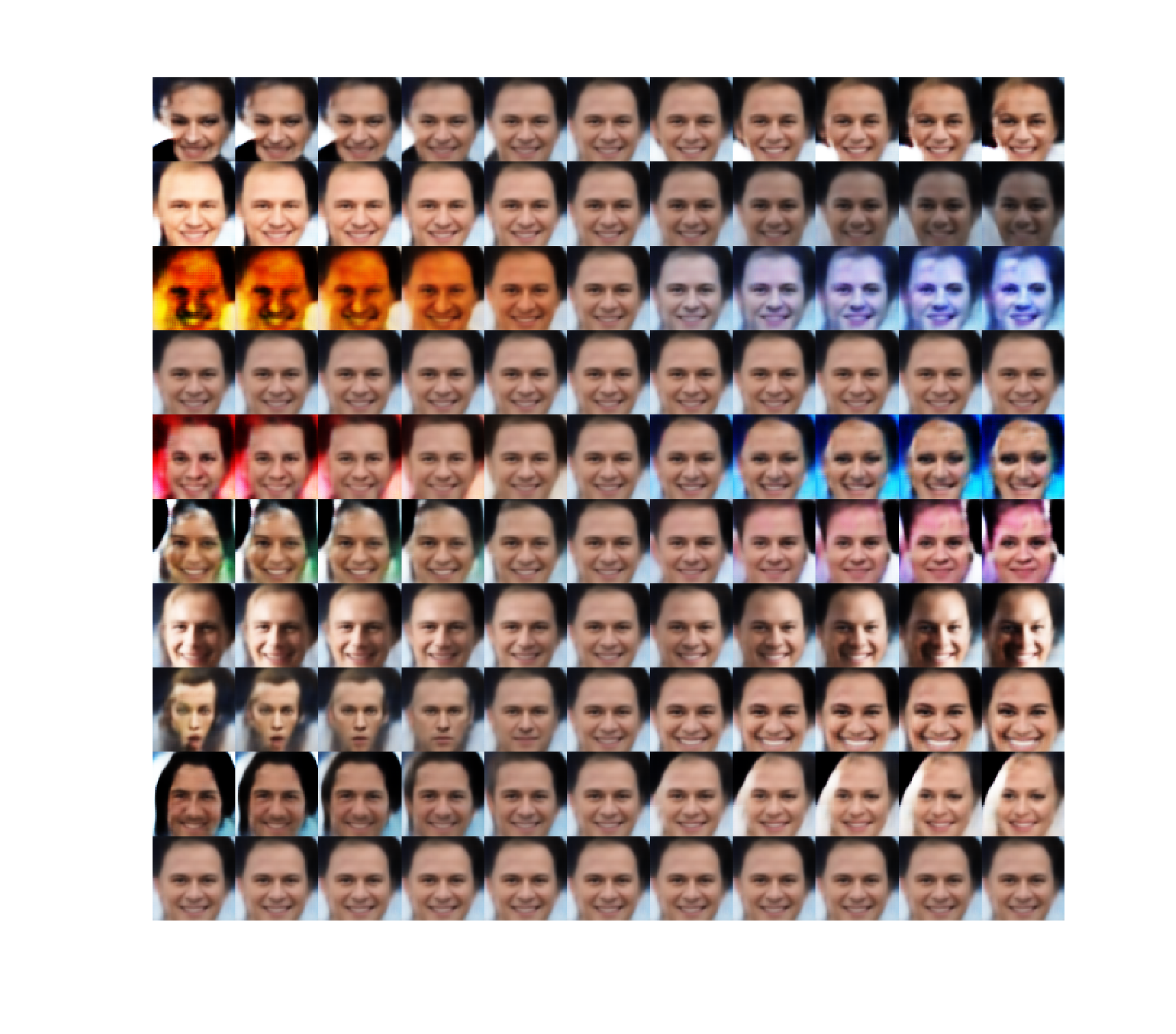}}
 \caption{}
\end{subfigure}

  \caption{\label{fig:celebAPathGen} Paths generated by all 40 transport operators trained on the celebA dataset. Images in the middle column of the image block are the reconstructed inputs and images to the right and left are images decoded from transformed latent vectors in positive and negative directions, respectively}
	
\end{figure}

Here we show additional experimental results for the CelebA experiment. Fig.~\ref{fig:celebAPathGen} shows the paths generated by the 40 transport operators trained on celebA data. Each row represents a different operator acting on the same input image. Images in the middle column of the image block are the reconstructed inputs and images to the right and left are images decoded from transformed latent vectors in positive and negative directions, respectively.

\begin{figure}[!htb]

\centering
\begin{subfigure}[b]{0.85\textwidth}
  \centering
	{\includegraphics[width=0.98\textwidth]{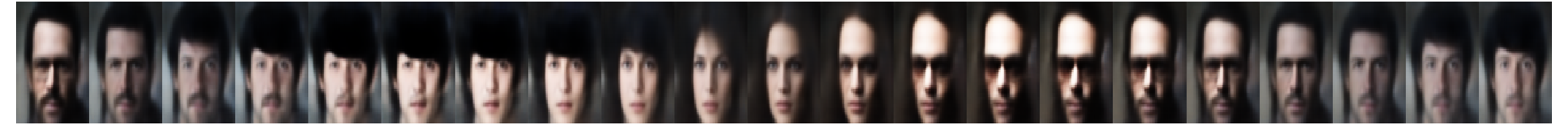}}
  \caption{}
	\label{subfig:cyclicTrans}
\end{subfigure}

\begin{subfigure}[b]{0.40\textwidth}
	{\includegraphics[width=0.98\textwidth]{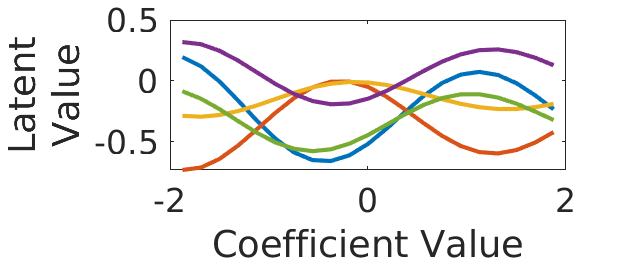}}%
  \caption{}
	\label{subfig:cyclicLatent}
\end{subfigure}
  \caption{\label{fig:cyclic} An example of an operator that generates a nearly cyclic path in the latent space (a) Image outputs along a path generated by a single learned operator. The images on the far left and far right look similar which indicates that this operator generates a nearly closed path that begins and ends at the same point. (b) Paths of five of the 32 latent dimensions as the learned operator is applied to them. This again highlights the cyclic nature of the transport operator generated paths.}
\end{figure}

Fig.~\ref{fig:cyclic} shows an interesting feature in our learned operators -- many of them generate cyclic paths that begin and end at nearly the same point. Also, in these cyclic sequences, the transformations seem to lead to a change in gender. The image sequence in Fig.~\ref{subfig:cyclicTrans} shows the path generated by a single operator. The image in the middle is the reconstructed input image and the images to the left and right are the paths generated by negative and positive coefficients respectively. This operator changes the hairline and quantity of bangs. As we apply the transport operator with a negative coefficient (to the left of center), the woman gains bangs and then becomes a man with a moustache on the far left of the image. As we apply the transport operator with a positive coefficient (to the right of center), the woman's forehead gets higher and then she becomes a man with a high forehead and eventually, on the far right the woman transforms into a man with bangs, similar to the man on the far left. The similarity between the generated images on the far left and far right is notable because this indicates the transformation path is nearly closed. Fig.~\ref{subfig:cyclicLatent} shows the change in five of the 32 latent dimensions over the generated path. These paths have a nearly cyclic structure. 

This is particularly interesting because in \citep{connor2020representing} they learn closed transformation paths by selecting point pairs on known cyclic transformation paths and highight the benefit of the transport operator model for representing these types of paths. In this case, we learn this cyclic path with only perceptual point pair supervision. Additionally, this identifies a benefit of the transport operator approach over other models of the manifold in a neural network latent space. We can learn nonlinear paths that keep the generated points in the same neighborhood in the latent space without extending to infinity which is inevitable when linear paths that are used to represent natural transformations.

\end{document}